# On the logical skills of large language models: evaluations using arbitrarily complex first-order logic problems

Shokhrukh Ibragimov[1], Arnulf Jentzen[2,3], and Benno Kuckuck[4]

[1] Applied Mathematics: Institute for Analysis and Numerics,
University of Münster, Germany; e-mail: `sibragim@uni-muenster.de`

[2] School of Data Science and Shenzhen Research Institute of Big Data,
The Chinese University of Hong Kong, Shenzhen (CUHK-Shenzhen),
China, e-mail: `ajentzen@cuhk.edu.cn`

[3] Applied Mathematics: Institute for Analysis and Numerics,
University of Münster, Germany; e-mail: `ajentzen@uni-muenster.de`

[4] Applied Mathematics: Institute for Analysis and Numerics,
University of Münster, Germany; e-mail: `bkuckuck@uni-muenster.de`

February 20, 2025

**Abstract**

We present a method of generating first-order logic statements whose complexity can be controlled along multiple dimensions. We use this method to automatically create several datasets consisting of questions asking for the truth or falsity of first-order logic statements in Zermelo–Fraenkel set theory. While the resolution of these questions does not require any knowledge beyond basic notation of first-order logic and set theory, it does require a degree of planning and logical reasoning, which can be controlled up to arbitrarily high difficulty by the complexity of the generated statements. Furthermore, we do extensive evaluations of the performance of various large language models, including recent models such as DeepSeek-R1 and OpenAI's o3-mini, on these datasets. All of the datasets along with the code used for generating them, as well as all data from the evaluations is publicly available at https://github.com/bkuckuck/logical-skills-of-llms.

# Contents

# 1 Introduction

Large language models (LLMs) have made significant progress in recent months on a wide variety of tasks (cf., for example, [4, 6, 10, 11, 19, 23, 52, 54, 55]). Mathematical reasoning is one area where LLMs have shown great promise and mathematical benchmarks remain a staple in the evaluation of the capabilities of these models (cf., for instance, [9, 15, 17, 24, 25, 33, 43, 63, 64]). However, many commonly used benchmarks are becoming saturated as LLM performance continues to improve, with models increasingly conquering even high-level competition or "PhD level" problems (cf., for example, [7, 11, 51, 62, 67]), thus encouraging the compilation of ever harder sets of problems (cf., for instance, [18, 20, 28, 59, 70]). At the same time there remain doubts about how representative these benchmark successes are of actual reasoning ability as opposed to mere memorization (cf., for example, [1, 8, 14, 47, 56, 58, 77]), with a number of benchmarks showing surprisingly weak performance of LLMs at seemingly not-so-challenging problems or on variations of more commonly employed problems (cf., for instance, [3, 24, 26, 31, 49, 68, 69]).

There is also considerable interest in employing language models for automated (or semi-automated) theorem proving (cf., for example, [2, 5, 12, 16, 22, 34, 35, 38–42, 44, 57, 60, 61, 71, 75]). In this context, there is a need for datasets of problems with known solutions that can be used in the training of specialized LLMs (cf., for instance, [21, 36, 37, 66, 72, 73, 76, 78, 79]). In particular, in the more recent development of "reasoning models" that are trained using reinforcement learning to produce long reasoning traces before arriving at an answer and in the distillation of the reasoning capabilities of these models into smaller LLMs, large (often synthetic) datasets of problems with known solutions play a crucial role (cf., for example, [11, 30, 45, 46, 48, 50, 65]).

Methods for programmatically generating arbitrary amounts of problems with known solutions are therefore of interest both for the purpose of benchmarking (mitigating issues with memorization) and for the training of specialized LLMs and reasoning models. A number of such methods have appeared in the literature (cf., for instance, [13, 24, 26, 27, 49, 80]). The major purpose of this work is to present a method for generating first-order logic statements that can be used to create datasets of problems with known solutions. Solving these problems does not require any knowledge beyond basic notation of first-order logic and set theory, but it does require a degree of planning and logical reasoning. The complexity of these statements and with it the amount of reasoning necessary to solve them can be controlled along several dimensions, making it possible to generate problems of varying or arbitrarily high difficulty. Our approach is based on the following observation in Theorem 1.2, which uses the notion introduced in Definition 1.1. Here, like in the rest of this article, whenever we say that a statement like (1.1) holds (or is true) we mean that it is provable in *Zermelo–Fraenkel set theory* (ZF).[1]

**Definition 1.1.** *Let $A$ and $B$ be sets and let $k \in \{0, 1, 2\}$. Then we say that $A$ is in $k$-relation to $B$ if and only if we have that*

*(i) it holds that $k = 0$ and $A \subseteq B$,*

*(ii) it holds that $k = 1$ and $A \subsetneq B$, or*

---
[1] For all the statements considered in this article, provability is equivalent to satisfiability in ZF. Similarly, whenever we say that a statement is false, we mean that its negation is provable or (equivalently in all cases considered here) satisfiable in ZF.

*(iii) it holds that $k = 2$ and $A \in B$.*

**Theorem 1.2.** *Let $n, m \in \mathbb{N}$, $i_1, i_2, \ldots, i_m, j_1, j_2, \ldots, j_m \in \{1, 2, \ldots, n\}$, $\ell_1, \ell_2, \ldots, \ell_m \in \{0, 1, 2\}$. Then the following two statements are equivalent:*

*(i) It holds that*

$$\exists\, X_1, X_2, \ldots, X_n \colon \left[ \bigwedge\nolimits_{k \in \{1,2,\ldots,m\}} (X_{i_k} \text{ is in } \ell_k\text{-relation to } X_{j_k}) \right] \quad (1.1)$$

*(cf. Definition 1.1).*

*(ii) For all $l \in \mathbb{N}$, $e_1, e_2, \ldots, e_l \in \{1, 2, \ldots, m\}$ with $j_{e_l} = i_{e_1}$ and $\forall\, k \in \mathbb{N} \cap (0, l)\colon j_{e_k} = i_{e_{k+1}}$ it holds that $\sum_{k=1}^{l} \ell_{e_k} = 0$.*

Theorem 1.2 is Corollary 2.19, which a direct consequence of Theorem 2.18 in Section 2.2 below. Note that, roughly speaking, Theorem 1.2 says that a statement of the form (1.1) asserting the existence of sets satisfying certain element, subset, and strict subset relations is true (in Zermelo–Fraenkel set theory) if and only if no cycle of relations involving an element or strict subset relation is being asserted. Varying, for example, the number of variables involved, the number of conjuncts, the types of relations used, or the minimum number of relations making up a cycle (in a false statement), naturally allows to adjust the difficulty of deciding the truth or falsity of such a statement. We use this method to create a number of datasets consisting of questions asking for the truth or falsity of first-order logic statements in Zermelo–Fraenkel set theory. We then perform extensive evaluations of the performance of various LLMs on these datasets. All of the datasets along with the code used for generating them, as well as all data from the evaluations is publicly available at https://github.com/bkuckuck/logical-skills-of-llms.

The remainder of this article is structured as follows: In Section 2 we present the theoretical background underlying our method of generating first-order logic statements, in Section 3 we describe in detail how we generate the datasets used in our evaluations, and in Section 4 we present the results of our evaluations.

## 2 A family of arbitrarily complex first-order logic statements

In this section, we present the theoretical background underlying our method of generating first-order logic statements. In Section 2.1 below we introduce some basic definitions and statements from graph theory. All of the material contained in this section is standard and included here only for completeness and to fix notations. In Section 2.2 we present our main theoretical result, Theorem 2.18, which gives an equivalence between the truth of certain first-order logic statements and the absence of certain kinds of cycles in an associated graph. This result is the basis for the generation of the datasets in Section 3.

### 2.1 Preliminaries from graph theory

**Definition 2.1.** *Let $V, E$ be sets. Then we denote by $\Gamma_{V,E}$ the set which satisfies*

$$\Gamma_{V,E} = \{G \colon (G \text{ is a function from } E \times \{0,1\} \text{ to } V)\}. \quad (2.1)$$

**Definition 2.2** (Graph with vertices and edges). *We say that $G$ is a graph with vertices $V$ and edges $E$ if and only if $G \in \Gamma_{V,E}$.*

**Definition 2.3** (Graph). *We say that $G$ is a graph if and only if there exist sets $V$ and $E$ such that $G$ is a graph with vertices $V$ and edges $E$ (cf. Definition 2.2).*

**Definition 2.4.** *Let $G$ be a graph with vertices $V$ and edges $E$ and let $u, v \in V$ (cf. Definition 2.2). Then we denote by $P_G(u,v) \subseteq \bigcup_{n \in \mathbb{N}} E^n$ the set which satisfies*

$$P_G(u,v) = \bigcup_{n \in \mathbb{N}} \left\{ (e_1, \ldots, e_n) \in E^n \colon \begin{bmatrix} G(e_1, 0) = u,\ G(e_n, 1) = v, \\ \forall\, j \in \mathbb{N} \cap (0, n) \colon G(e_j, 1) = G(e_{j+1}, 0) \end{bmatrix} \right\}. \quad (2.2)$$

**Lemma 2.5.** *Let $G$ be a graph with vertices $V$ and edges $E$, let $u, v, w \in V$, and let $n, m \in \mathbb{N}$, $e_1, e_2 \ldots, e_n, f_1, f_2, \ldots, f_m \in E$ satisfy $(e_1, e_2, \ldots, e_n) \in P_G(u,v)$ and $(f_1, f_2, \ldots, f_m) \in P_G(v,w)$ (cf. Definitions 2.2 and 2.4). Then it holds that*

$$(e_1, e_2, \ldots, e_n, f_1, f_2, \ldots, f_m) \in P_G(u,w). \quad (2.3)$$

*Proof of Lemma 2.5.* Observe that the assumption that $(e_1, e_2, \ldots, e_n) \in P_G(u,v)$ ensures that

$$G(e_1, 0) = u, \quad G(e_n, 1) = v, \quad \text{and} \quad \forall\, j \in \mathbb{N} \cap (0, n) \colon G(e_j, 1) = G(e_{j+1}, 0). \quad (2.4)$$

Moreover, note that the assumption that $(f_1, f_2, \ldots, f_m) \in P_G(v,w)$ implies that

$$G(f_1, 0) = v, \quad G(f_m, 1) = w, \quad \text{and} \quad \forall\, j \in \mathbb{N} \cap (0, m) \colon G(f_j, 1) = G(f_{j+1}, 0). \quad (2.5)$$

Combining this and (2.4) shows that

$$G(e_n, 1) = G(f_1, 0). \quad (2.6)$$

This, (2.4), and (2.5) prove that

$$(e_1, e_2, \ldots, e_n, f_1, f_2, \ldots, f_m) \in P_G(u,w). \quad (2.7)$$

The proof of Lemma 2.5 is thus complete. $\square$

**Definition 2.6** (Topological ordering). *Let $G$ be a graph with vertices $V$ and edges $E$ (cf. Definition 2.2). Then we say that $\preceq$ is a topological ordering of $G$ if and only if*

(i) *it holds that $\preceq$ is a total order on $V$ and*

(ii) *it holds for all $e \in E$ that $G(e, 0) \preceq G(e, 1)$ and $G(e, 0) \neq G(e, 1)$.*

**Definition 2.7** (Acyclic graph). *Let $G$ be a graph with vertices $V$ and edges $E$ (cf. Definition 2.2). Then we say that $G$ is acyclic if and only if it holds for all $v \in V$ that $P_G(v,v) = \emptyset$ (cf. Definition 2.4).*

**Lemma 2.8.** *Let $G$ be a graph and assume that $G$ is acyclic (cf. Definitions 2.3 and 2.7). Then there exists a topological ordering of $G$ (cf. Definition 2.6).*

*Proof of Lemma 2.8.* Throughout this proof let $V, E$ be sets which satisfy that $G$ is a graph with vertices $V$ and edges $E$ and let $\leqq \subseteq V \times V$ satisfy for all $u, v \in V$ that

$$u \leqq v \leftrightarrow (u = v \vee P_G(u, v) \neq \emptyset) \tag{2.8}$$

(cf. Definitions 2.2 and 2.4). Observe that (2.8) demonstrates that $\leqq$ is reflexive. Furthermore, note that Lemma 2.5 establishes that $\leqq$ is transitive. In the next step we combine (2.8) and Lemma 2.5 to obtain that for all $u, v \in V$ with $P_G(u, v) \neq \emptyset$ and $P_G(v, u) \neq \emptyset$ it holds that

$$P_G(u, u) \neq \emptyset. \tag{2.9}$$

Combining this and the assumption that $G$ is acyclic ensures that for all $u, v \in V$ with $P_G(u, v) \neq \emptyset$ it holds that $P_G(v, u) = \emptyset$. This implies that $\leqq$ is antisymmetric. Combining this and the fact that $\leqq$ is reflexive and transitive shows that $\leqq$ is a partial order on $V$. This and the order extension principle[2] (see, e.g., [29, Subsection 2.3.2]) prove that there exists a total order $\preceq \subseteq V \times V$ on $V$ which satisfies $\leqq \subseteq \preceq$. Next, observe that (2.8) and the assumption that $G$ is acyclic demonstrate that for all $e \in E$ it holds that

$$G(e, 0) \leqq G(e, 1) \qquad \text{and} \qquad G(e, 0) \neq G(e, 1). \tag{2.10}$$

Combining the fact that $\leqq \subseteq \preceq$ and the fact that $\preceq$ is a total order on $V$ therefore establishes that $\preceq$ is a topological ordering of $G$ (cf. Definition 2.6). The proof of Lemma 2.8 is thus complete. $\square$

## 2.2 Deciding the truth of certain first-order logic statements in Zermelo–Fraenkel set theory

**Lemma 2.9.** *Let $k \in \mathbb{N}$, let $X_0, X_1, \ldots, X_k$ be sets, and assume $X_k = X_0$. Then it does not hold that*

$$\forall\, j \in \{1, 2, \ldots, k\} \colon X_{j-1} \in X_j. \tag{2.11}$$

*Proof of Lemma 2.9.* Throughout this proof let $A$ be the set which satisfies $A = \{X_0, X_1, \ldots, X_k\}$. Note that the fact that $A \neq \emptyset$ and the axiom of regularity ensure that there exists $y \in A$ which satisfies

$$y \cap A = \emptyset. \tag{2.12}$$

This, the fact that $A = \{X_0, X_1, \ldots, X_k\}$, and the fact that $X_0 = X_k$ show that there exists $\jmath \in \{1, 2, \ldots, k\}$ which satisfies $y = X_\jmath$. Combining this with (2.12) demonstrates that for all $j \in \{0, 1, \ldots, k\}$ it holds that $X_j \notin X_\jmath$. This implies that it does not hold that $\forall\, j \in \{1, 2, \ldots, k\} \colon X_{j-1} \in X_j$. The proof of Lemma 2.9 is thus complete. $\square$

**Lemma 2.10.** *Let $G$ be a graph with vertices $V$ and edges $E$, let $\preceq \subseteq V \times V$ be a topological ordering of $G$, and assume that $\preceq$ is a well-ordering on $V$ (cf. Definition 2.2). Then there exists a set $S$ and a function $\varphi \colon V \to S$ such that for all $\ell \colon E \to \{0, 1, 2\}$ it holds that*

$$\forall\, e \in E \colon [\varphi(G(e, 0)) \text{ is in } \ell(e)\text{-relation to } \varphi(G(e, 1))] \tag{2.13}$$

*(cf. Definition 1.1).*

---

[2] ORDER EXTENSION PRINCIPLE: For every set $P$ and every partial order $\leqq \subseteq P \times P$ there exists a total order $\preceq \subseteq P \times P$ such that $\leqq \subseteq \preceq$.

*Proof of Lemma 2.10.* Observe that the assumption that $\preceq$ is a well-ordering on $V$ and transfinite recursion ensure that there exists a set $S$ and a function $\varphi \colon V \to S$ which satisfy for all $v \in V$ that

$$\varphi(v) = \bigcup\nolimits_{u \in V \colon u \preceq v, u \neq v} (\varphi(u) \cup \{\varphi(u)\}). \tag{2.14}$$

Note that (2.14) and the axiom of regularity imply that for all $u, v \in V$ with $u \preceq v$ and $u \neq v$ it holds that

$$\varphi(u) \subsetneq \varphi(v) \qquad \text{and} \qquad \varphi(u) \in \varphi(v). \tag{2.15}$$

This and the assumption that $\preceq$ is a topological ordering of $G$ show that for all $e \in E$ it holds that $\varphi(G(e,0)) \subsetneq \varphi(G(e,1))$ and $\varphi(G(e,0)) \in \varphi(G(e,1))$. Hence, we obtain that for all $e \in E$, $k \in \{0, 1, 2\}$ it holds that

$$\varphi(G(e,0)) \text{ is in } k\text{-relation to } \varphi(G(e,1)). \tag{2.16}$$

This proves (2.13). The proof of Lemma 2.10 is thus complete. $\square$

**Corollary 2.11.** *Let $G$ be a graph with vertices $V$ and edges $E$, assume that $V$ is finite, and assume that $G$ is acyclic (cf. Definitions 2.2 and 2.7). Then there exists a set $S$ and a function $\varphi \colon V \to S$ such that for all $\ell \colon E \to \{0, 1, 2\}$ it holds that*

$$\forall\, e \in E \colon [\varphi(G(e,0)) \text{ is in } \ell(e)\text{-relation to } \varphi(G(e,1))] \tag{2.17}$$

*(cf. Definition 1.1).*

*Proof of Corollary 2.11.* Observe that the assumption that $G$ is acyclic and Lemma 2.8 demonstrate that there exists $\preceq\, \subseteq V \times V$ which satisfies that $\preceq$ is a topological ordering of $G$ (cf. Definition 2.6). Note that the fact that $\preceq$ is a total ordering on $V$ and the fact that $V$ is finite establish that $\preceq$ is a well-ordering. Lemma 2.10 hence ensures (2.17). The proof of Corollary 2.11 is thus complete. $\square$

**Lemma 2.12.** *Let $G$ be a graph with vertices $V$ and edges $E$, let $\ell \colon E \to \{0, 1\}$ be a function, let $X_v$, $v \in V$, satisfy for all $e \in E$ that $X_{G(e,0)}$ is in $\ell(e)$-relation to $X_{G(e,1)}$, and let $u, v \in V$ satisfy $P_G(u, v) \neq \emptyset$. (cf. Definition 2.2). Then it holds that $X_u \subseteq X_v$.*

*Proof of Lemma 2.12.* Observe that the assumption that $P_G(u, v) \neq \emptyset$ implies there exist $l \in \mathbb{N}$, $e_1, e_2, \ldots, e_l \in E$ which satisfy $(e_1, e_2, \ldots, e_l) \in P_G(u, v)$. Note that the assumption that for all $k \in E$ it holds that $X_{G(k,0)}$ is in $\ell(k)$-relation to $X_{G(k,1)}$ and the fact that for all $k \in E$ it holds that $\ell(e_k) \in \{0, 1\}$ show that for all $j \in \{1, 2, \ldots, l\}$ it holds that $X_{G(e_j,0)} \subseteq X_{G(e_j,1)}$. Combining this and the fact that for all $j \in \mathbb{N} \cap (0, l)$ it holds that $G(e_j, 1) = G(e_{j+1}, 0)$ proves that $X_{G(e_1,0)} \subseteq X_{G(e_l,1)}$. The proof of Lemma 2.12 is thus complete. $\square$

**Lemma 2.13.** *Let $G$ be a graph with vertices $V$ and edges $E$, let $\ell \colon E \to \{0, 1, 2\}$ be a function, let $X_v$, $v \in V$, satisfy for all $e \in E$ that $X_{G(e,0)}$ is in $\ell(e)$-relation to $X_{G(e,1)}$, let $u, v \in V$, $l \in \mathbb{N}$, $e = (e_1, e_2, \ldots, e_l) \in P_G(u, v)$, and let $r = |\{k \in \{1, 2, \ldots, l\} \colon \ell(e_k) = 2\}|$ (cf. Definitions 2.2 and 2.4). Then there exist sets $Y_0, Y_1, \ldots, Y_r$ such that*

$$Y_0 = X_u, \qquad Y_r \subseteq X_v, \qquad \text{and} \qquad \forall\, k \in \mathbb{N} \cap (-\infty, r] \colon Y_{k-1} \in Y_k. \tag{2.18}$$

*Proof of Lemma 2.13.* Throughout this proof assume without loss of generality that $r > 0$ (cf. Lemma 2.12), let $i_1, i_2, \ldots, i_r \in \{1, 2, \ldots, l\}$ satisfy for all $j \in \{1, 2, \ldots, r\}$ that $i_1 < i_2 < \cdots < i_r$ and $\ell(e_{i_j}) = 2$, and let $Y_0, Y_1, \ldots, Y_r$ satisfy for all $k \in \{1, 2, \ldots, r\}$ that

$$Y_0 = X_{G(e_1,0)} \quad \text{and} \quad Y_k = X_{G(e_{i_k},1)}. \tag{2.19}$$

Observe that the assumption that for all $j \in \{1, 2, \ldots, r\}$ it holds that $i_1 < i_2 < \cdots < i_r$ and $\ell(e_{i_j}) = 2$ and the assumption that $r = |\{k \in \{1, 2, \ldots, l\} : \ell(e_k) = 2\}|$ demonstrate that

$$\{k \in \{1, 2, \ldots, l\} : \ell(e_k) = 2\} = \{i_1, i_2, \ldots, i_r\}. \tag{2.20}$$

In addition, note that the fact that $e \in P_G(u, v)$ establishes that for all $j, k \in \mathbb{N} \cap \{1, 2, \ldots, r\}$ with $j \leq k$ it holds that

$$(e_j, e_{j+1}, \ldots, e_k) \in P_G(G(e_j, 0), G(e_k, 1)). \tag{2.21}$$

This and Lemma 2.12 ensure that for all $j, k \in \mathbb{N} \cap \{1, 2, \ldots, r\}$ with $j \leq k$ and $\ell(e_j), \ell(e_{j+1}), \ldots, \ell(e_k) \in \{0, 1\}$ it holds that

$$X_{G(e_j,0)} \subseteq X_{G(e_k,1)}. \tag{2.22}$$

Combining this and the fact that for all $j \in \mathbb{N} \cap (0, l)$ it holds that $G(e_k, 1) = G(e_{k+1}, 0)$ implies that for all $j, k \in \mathbb{N} \cap \{1, 2, \ldots, r\}$ with $j \leq k$ and $\ell(e_j), \ell(e_{j+1}), \ldots, \ell(e_k) \in \{0, 1\}$ it holds that

$$X_{G(e_j,0)} \subseteq X_{G(e_k,0)} \quad \text{and} \quad X_{G(e_j,1)} \subseteq X_{G(e_k,1)}. \tag{2.23}$$

This, (2.19), (2.20), and the assumption that for all $e \in E$ it holds that $X_{G(e,0)}$ is in $\ell(e)$-relation to $X_{G(e,1)}$ show that

$$Y_0 = X_{G(e_1,0)} \subseteq X_{G(e_{i_1},0)} \in X_{G(e_{i_1},1)} = Y_1. \tag{2.24}$$

Next we combine (2.19), (2.20), (2.23), and the assumption that for all $e \in E$ it holds that $X_{G(e,0)}$ is in $\ell(e)$-relation to $X_{G(e,1)}$ to obtain that for all $j \in \mathbb{N} \cap (0, r)$ it holds that

$$Y_j = X_{G(e_{i_j},1)} \subseteq X_{G(e_{i_{j+1}},0)} \in X_{G(e_{i_{j+1}},1)} = Y_{j+1}. \tag{2.25}$$

Moreover, observe that (2.19), (2.23), and (2.20) prove that

$$Y_r = G(e_{i_r}, 1) \subseteq G(e_l, 1). \tag{2.26}$$

Combining (2.24) and (2.25) therefore demonstrates (2.18). The proof of Lemma 2.13 is thus complete. $\square$

**Lemma 2.14.** *Let $G$ be a graph with vertices $V$ and edges $E$, let $\ell \colon E \to \{0, 1, 2\}$ be a function, let $X_v$, $v \in V$, satisfy for all $e \in E$ that $X_{G(e,0)}$ is in $\ell(e)$-relation to $X_{G(e,1)}$, and let $v \in V$, $l \in \mathbb{N}$, $e = (e_1, e_2, \ldots, e_l) \in P_G(v, v)$ (cf. Definitions 2.2 and 2.4). Then it holds for all $k \in \{1, 2, \ldots, l\}$ that $\ell(e_k) < 2$.*

*Proof of Lemma 2.14.* Throughout this proof let $r = |\{k \in \{1, 2, \ldots, l\}\colon \ell(e_k) = 2\}|$. Note that Lemma 2.13 establishes that there exist sets $Y_0, Y_1, \ldots, Y_r$ such that

$$Y_0 = X_v, \quad Y_r \subseteq X_v, \quad \text{and} \quad \forall\, k \in \mathbb{N} \cap (-\infty, r]\colon Y_{k-1} \in Y_k. \qquad (2.27)$$

Lemma 2.9 hence ensures that $r = 0$. The proof of Lemma 2.14 is thus complete. $\square$

**Lemma 2.15.** *Let $G$ be a graph with vertices $V$ and edges $E$, let $\ell\colon E \to \{0, 1, 2\}$ be a function, let $X_v$, $v \in V$, satisfy for all $e \in E$ that $X_{G(e,0)}$ is in $\ell(e)$-relation to $X_{G(e,1)}$, and let $l \in \mathbb{N}$, $v \in V$, $e = (e_1, e_2, \ldots, e_l) \in P_G(v, v)$ (cf. Definitions 2.2 and 2.4). Then it holds for all $k \in \{1, 2, \ldots, l\}$ that $\ell(e_k) = 0$.*

*Proof of Lemma 2.15.* Observe that Lemma 2.14 implies that for all $k \in \{1, 2, \ldots, l\}$ it holds that

$$\ell(e_k) < 2. \qquad (2.28)$$

This and the assumption that for all $k \in \{1, 2, \ldots, l\}$ it holds that $X_{G(e_k,0)}$ is in $\ell(e_k)$-relation to $X_{G(e_k,1)}$ show that for all $k \in \{1, 2, \ldots, l\}$ it holds that

$$X_{G(e_k,0)} \subseteq X_{G(e_k,1)}. \qquad (2.29)$$

Furthermore, note that the fact that $e \in P_G(v, v)$ proves that

$$X_{G(e_l,1)} = X_v = X_{G(e_1,0)} \quad \text{and} \quad \forall\, k \in \mathbb{N} \cap (0, l)\colon X_{G(e_k,1)} = X_{G(e_{k+1},0)}. \qquad (2.30)$$

Combining this and (2.29) demonstrates that for all $k \in \{1, 2, \ldots, l\}$ it holds that

$$X_{G(e_k,0)} = X_{G(e_1,0)} = X_{G(e_k,1)}. \qquad (2.31)$$

This establishes that for all $k \in \{1, 2, \ldots, l\}$ it holds that $X_{G(e_k,0)}$ is not in 1-relation to $X_{G(e_k,1)}$. Combining the assumption that for all $k \in \{1, 2, \ldots, l\}$ it holds that $X_{G(e_k,0)}$ is in $\ell(e_k)$-relation to $X_{G(e_k,1)}$ and (2.28) therefore ensures that for all $k \in \{1, 2, \ldots, l\}$ it holds that $\ell(e_k) = 0$. The proof of Lemma 2.15 is thus complete. $\square$

**Lemma 2.16.** *Let $G$ be a graph with vertices $V$ and edges $E$, let $\ell\colon E \to \{0, 1, 2\}$ be a function, let $\rightsquigarrow\, \subseteq V \times V$ satisfy for all $u, v \in V$ that*

$$u \rightsquigarrow v \iff (u = v \vee P_{G|_{\ell^{-1}(\{0\})}}(u, v) \neq \emptyset), \qquad (2.32)$$

*for all $v \in V$ let $[v] = \{u \in V\colon u \rightsquigarrow v \wedge v \rightsquigarrow u\}$, let*

$$\mathscr{V} = \{[v]\colon v \in V\}, \quad \mathscr{E} = \{e \in E\colon \ell(e) > 0\} \cup \{e \in E\colon [G(e,0)] \neq [G(e,1)]\}, \qquad (2.33)$$

*let $\mathscr{G}\colon \mathscr{E} \times \{0, 1\} \to \mathscr{V}$ satisfy for all $e \in \mathscr{E}$, $j \in \{0, 1\}$ that $\mathscr{G}(e, j) = [G(e, j)]$, and let $v, w \in V$, $\mathfrak{l} \in \mathbb{N}$, $\mathfrak{e} = (\mathfrak{e}_1, \mathfrak{e}_2, \ldots, \mathfrak{e}_{\mathfrak{l}}) \in P_{\mathscr{G}}([v], [w])$ (cf. Definitions 2.2 and 2.4). Then there exist $l \in \mathbb{N}$, $e = (e_1, e_2, \ldots, e_l) \in P_G(v, w)$ such that $\{e_1, e_2, \ldots, e_l\} = \{\mathfrak{e}_1, \mathfrak{e}_2, \ldots, \mathfrak{e}_{\mathfrak{l}}\} \cup \{e_i\colon i \in \{1, 2, \ldots, l\}, \ell(e_i) = 0\}$.*

*Proof of Lemma 2.16.* Observe that (2.32) and Lemma 2.5 imply that $\rightsquigarrow$ is transitive and reflexive. This shows that $\{(u, v) \in V \times V \colon u \rightsquigarrow v \wedge v \rightsquigarrow u\}$ is an equivalence relation. Combining this and the fact that $v \in [v] = \mathscr{G}(\mathfrak{e}_1, 0) \ni G(\mathfrak{e}_1, 0)$ proves that

$$v \rightsquigarrow G(\mathfrak{e}_1, 0). \tag{2.34}$$

In the next step we note that the fact that $\{(u, v) \in V \times V \colon u \rightsquigarrow v \wedge v \rightsquigarrow u\}$ is an equivalence relation and the fact that for all $j \in \mathbb{N} \cap (0, \mathfrak{l})$ it holds that

$$G(\mathfrak{e}_j, 1) \in \mathscr{G}(\mathfrak{e}_j, 1) = \mathscr{G}(\mathfrak{e}_{j+1}, 0) \ni G(\mathfrak{e}_{j+1}, 0) \tag{2.35}$$

demonstrate that it holds for all $j \in \mathbb{N} \cap (0, \mathfrak{l})$ that

$$G(\mathfrak{e}_j, 1) \rightsquigarrow G(\mathfrak{e}_{j+1}, 0). \tag{2.36}$$

Next, observe that the fact that $\{(u, v) \in V \times V \colon u \rightsquigarrow v \wedge v \rightsquigarrow u\}$ is an equivalence relation and the fact that $w \in [w] = \mathscr{G}(\mathfrak{e}_\mathfrak{l}, 1) \ni G(\mathfrak{e}_\mathfrak{l}, 1)$ establish that

$$G(\mathfrak{e}_\mathfrak{l}, 1) \rightsquigarrow w. \tag{2.37}$$

This, (2.34), (2.36), and Lemma 2.5 ensure that there exist $l \in \mathbb{N}$, $e = (e_1, e_2, \ldots, e_l) \in P_G(v, w)$ such that $\{e_1, e_2, \ldots, e_l\} = \{\mathfrak{e}_1, \mathfrak{e}_2, \ldots, \mathfrak{e}_\mathfrak{l}\} \cup \{e_i \colon i \in \{1, 2, \ldots, l\}, \ell(e_i) = 0\}$. The proof of Lemma 2.16 is thus complete. $\square$

**Lemma 2.17.** *Let $G$ be a graph with vertices $V$ and edges $E$, assume that $V$ is finite, let $\ell \colon E \to \{0, 1, 2\}$ be a function, let $\rightsquigarrow \subseteq V \times V$ satisfy for all $u, v \in V$ that*

$$u \rightsquigarrow v \iff (u = v \vee P_{G|_{\ell^{-1}(\{0\})}}(u, v) \neq \emptyset), \tag{2.38}$$

*for all $v \in V$ let $[v] = \{u \in V \colon u \rightsquigarrow v \wedge v \rightsquigarrow u\}$, let*

$$\mathscr{V} = \{[v] \colon v \in V\}, \quad \mathscr{E} = \{e \in E \colon \ell(e) > 0\} \cup \{e \in E \colon [G(e, 0)] \neq [G(e, 1)]\}, \tag{2.39}$$

*let $\mathscr{G} \colon \mathscr{E} \times \{0, 1\} \to \mathscr{V}$ satisfy for all $e \in \mathscr{E}$, $j \in \{0, 1\}$ that $\mathscr{G}(e, j) = [G(e, j)]$, and assume that for all $v \in V$, $l \in \mathbb{N}$, $e = (e_1, e_2, \ldots, e_l) \in P_G(v, v)$ it holds that $\sum_{j=1}^{l} \ell(e_j) = 0$ (cf. Definitions 2.2 and 2.4). Then*

(i) *it holds that $\mathscr{G}$ is acyclic,*

(ii) *there exist sets $X_v$, $v \in \mathscr{V}$, such that for all $e \in \mathscr{E}$ it holds that $X_{\mathscr{G}(e,0)}$ is in $\ell(e)$-relation to $X_{\mathscr{G}(e,1)}$,*

(iii) *there exist sets $Y_v$, $v \in V$, such that for all $e \in E$ it holds that $Y_{G(e,0)}$ is in $\ell(e)$-relation to $Y_{G(e,1)}$*

*(cf. Definition 2.7).*

*Proof of Lemma 2.17.* Note that Lemma 2.16 implies that for all $v \in V$, $\ell \in \mathbb{N}$, $\mathbf{e} = (\mathbf{e}_1, \mathbf{e}_2, \ldots, \mathbf{e}_\ell) \in P_{\mathscr{G}}([v], [v])$ there exist $l \in \mathbb{N}$, $e = (e_1, e_2, \ldots, e_l) \in P_G(v, v)$ such that

$$\{e_1, e_2, \ldots, e_l\} \supseteq \{\mathbf{e}_1, \mathbf{e}_2, \ldots, \mathbf{e}_\ell\}. \tag{2.40}$$

Combining this and the assumption that for all $v \in V$, $l \in \mathbb{N}$, $e = (e_1, e_2, \ldots, e_l) \in P_G(v, v)$ it holds that $\sum_{j=1}^{l} \ell(e_j) = 0$ shows that for all $v \in V$, $\ell \in \mathbb{N}$, $\mathbf{e} = (\mathbf{e}_1, \mathbf{e}_2, \ldots, \mathbf{e}_\ell) \in P_{\mathscr{G}}([v], [v])$ there exist $l \in \mathbb{N}$, $e = (e_1, e_2, \ldots, e_l) \in P_G(v, v)$ such that

$$\{e_1, e_2, \ldots, e_l\} \supseteq \{\mathbf{e}_1, \mathbf{e}_2, \ldots, \mathbf{e}_\ell\} \quad \text{and} \quad \forall j \in \{1, 2, \ldots, l\} \colon \ell(e_j) = 0. \tag{2.41}$$

This and the fact for all $v \in V$, $l \in \mathbb{N}$, $e = (e_1, e_2, \ldots, e_l) \in P_G(v, v)$ with $\forall j \in \{1, 2, \ldots, l\} \colon \ell(e_j) = 0$ it holds that $\forall j \in \{1, 2, \ldots, l\} \colon (G(e_i, 0) \rightsquigarrow G(e_i, 1) \land G(e_i, 1) \rightsquigarrow G(e_i, 0))$ prove that for all $v \in V$, $\ell \in \mathbb{N}$, $\mathbf{e} = (\mathbf{e}_1, \mathbf{e}_2, \ldots, \mathbf{e}_\ell) \in P_{\mathscr{G}}([v], [v])$ it holds that

$$\forall j \in \{1, 2, \ldots, \ell\} \colon [G(\mathbf{e}_j, 0)] = [G(\mathbf{e}_j, 1)]. \tag{2.42}$$

Combining this and (2.39) demonstrates item (i). Observe that item (i), the assumption that $V$ is finite, and Corollary 2.11 establish that there exist sets $X_v$, $v \in \mathscr{V}$, which satisfy that for all $\mathbf{e} \in \mathscr{E}$ it holds that

$$X_{\mathscr{G}(\mathbf{e}, 0)} \text{ is in } \ell(\mathbf{e})\text{-relation to } X_{\mathscr{G}(\mathbf{e}, 1)}. \tag{2.43}$$

This proves item (ii). Next let $Y_v$, $v \in V$, satisfy for all $v \in V$ that

$$Y_v = X_{[v]}. \tag{2.44}$$

Note that (2.43) and (2.44) ensure that for all $\mathbf{e} \in \mathscr{E}$ it holds that

$$Y_{G(\mathbf{e}, 0)} = X_{[G(\mathbf{e}, 0)]} = X_{\mathscr{G}(\mathbf{e}, 0)} \text{ is in } \ell(\mathbf{e})\text{-relation to } X_{\mathscr{G}(\mathbf{e}, 1)} = X_{[G(\mathbf{e}, 1)]} = Y_{G(\mathbf{e}, 1)}. \tag{2.45}$$

In addition, observe that (2.44) implies that for all $e \in E$ with $[G(e, 0)] = [G(e, 1)]$ and $\ell(e) = 0$ it holds that

$$Y_{G(e, 0)} = X_{[G(e, 0)]} \subseteq X_{[G(e, 1)]} = Y_{G(e, 1)}. \tag{2.46}$$

Combining this, (2.45), and the fact that $E = \mathscr{E} \cup \{e \in E \colon [G(e, 0)] = [G(e, 1)] \land \ell(e) = 0\}$ demonstrates item (iii). The proof of Lemma 2.17 is thus complete. $\square$

**Theorem 2.18.** *Let $G$ be a graph with vertices $V$ and edges $E$, assume that $V$ is finite, and let $\ell \colon E \to \{0, 1, 2\}$ be a function (cf. Definition 2.2). Then the following two statements are equivalent:*

(i) *There exist $X_v$, $v \in V$, such that for all $e \in E$ it holds that $X_{G(e, 0)}$ is in $\ell(e)$-relation to $X_{G(e, 1)}$ (cf. Definition 1.1).*

(ii) *For all $v \in V$, $l \in \mathbb{N}$, $e = (e_1, e_2, \ldots, e_l) \in P_G(v, v)$ it holds that $\forall j \in \{1, 2, \ldots, l\} \colon \ell(e_j) = 0$ (cf. Definition 2.4).*

*Proof of Theorem 2.18.* Note that Lemma 2.17 shows that ((i) → (ii)). Moreover, observe that Lemma 2.15 establishes that ((ii) → (i)). The proof of Theorem 2.18 is thus complete. □

**Corollary 2.19.** *Let $n, m \in \mathbb{N}$, $i_1, i_2, \ldots, i_m, j_1, j_2, \ldots, j_m \in \{1, 2, \ldots, n\}$, $\ell_1, \ell_2, \ldots, \ell_m \in \{0, 1, 2\}$. Then the following two statements are equivalent:*

*(i) It holds that*

$$\exists\, X_1, X_2, \ldots, X_n \colon \left[\bigwedge\nolimits_{k \in \{1,2,\ldots,m\}} (X_{i_k} \text{ is in } \ell_k\text{-relation to } X_{j_k})\right] \quad (2.47)$$

*(cf. Definition 1.1).*

*(ii) For all $l \in \mathbb{N}$, $e_1, e_2, \ldots, e_l \in \{1, 2, \ldots, m\}$ with $j_{e_k} = i_{e_1}$ and $\forall\, k \in \mathbb{N} \cap (0, l) \colon j_{e_k} = i_{e_{k+1}}$ it holds that $\sum_{k=1}^{l} \ell_{e_k} = 0$.*

*Proof of Corollary 2.19.* Note that Theorem 2.18 proves that ((i) ↔ (ii)). The proof of Corollary 2.19 is thus complete. □

**Corollary 2.20.** *Let $n, m \in \mathbb{N}$, $i_1, i_2, \ldots, i_m, j_1, j_2, \ldots, j_m \in \{1, 2, \ldots, n\}$. Then the following two statements are equivalent:*

*(i) It does not hold that*

$$\exists\, X_1, X_2, \ldots, X_n \colon [X_{i_1} \in X_{j_1} \wedge X_{i_2} \in X_{j_2} \wedge \ldots \wedge X_{i_m} \in X_{j_m}]. \quad (2.48)$$

*(ii) There exist $l \in \mathbb{N}$, $e_1, e_2, \ldots, e_l \in \{1, 2, \ldots, m\}$ with $j_{e_k} = i_{e_1}$ and $\forall\, k \in \mathbb{N} \cap (0, l) \colon j_{e_k} = i_{e_{k+1}}$.*

*Proof of Corollary 2.20.* Observe that Corollary 2.19 ensures that ((i) ↔ (ii)). The proof of Corollary 2.20 is thus complete. □

## 3 Generating first-order logic sentences

In this section we describe how we generated the first-order logic sentences used in the evaluations described in Section 4. Roughly speaking, the test sets consist of tuples $(G, \ell, N)$ where, for some $n, m \in \mathbb{N}$, $G$ is a graph with vertices $\{1, 2, \ldots, n\}$ and edges $\{1, 2, \ldots, m\}$, $\ell \colon \{1, 2, \ldots, m\} \to \{0, 1, 2\}$ is a function, and $N \subseteq \{0, 1, \ldots, n-1\}$. For every $n, m \in \mathbb{N}$, every graph with vertices $\{1, 2, \ldots, n\}$ and edges $\{1, 2, \ldots, m\}$, and every $\ell \colon \{1, 2, \ldots, m\}$ we can generate a first-order logic sentence equivalent to the statement that

$$\exists\, X_1, X_2, \ldots, X_n \colon \left[\bigwedge\nolimits_{k \in \{1,2,\ldots,m\}} (X_{G(k,0)} \text{ is in } \ell(k)\text{-relation to } X_{G(k,1)})\right]. \quad (3.1)$$

Theorem 2.18 provides a condition in terms of $G$ and $\ell$ for deciding whether such a sentence is true or false. More complex statements can be derived from this by, roughly speaking, inserting a pair of negations before some quantifier and pushing one of the negations through to the right using the rules for quantifier negation and De Morgan's laws. We will describe our construction of first-order logic sentences in more detail in Section 3.1.

## 3.1 Constructing first-order logic sentences

For every $n, m \in \mathbb{N}$, every graph $G$ with vertices $\{1, 2, \ldots, n\}$ and edges $\{1, 2, \ldots, m\}$, every function $\ell\colon \{1, 2, \ldots, m\} \to \{0, 1, 2\}$, every set $N \subseteq \{0, 1, \ldots, n-1\}$, every $i \in \{0, 1, \ldots, n\}$, every $j \in \{1, 2, \ldots, m\}$, and every $p \in \{0, 1\}$ let $\Phi_{i,p}(G, \ell, N)$ and $\Xi_{j,p}(G, \ell, N)$ be the first-order logic formulae that satisfy

$$\Phi_{i,0}(G,\ell,N) = \begin{cases} \exists\, X_{i+1}\colon (\Phi_{i+1,0}(G,N)) & : i < n \wedge i \notin N \\ \neg(\forall\, X_{i+1}\colon (\Phi_{i+1,1}(G,N))) & : i < n \wedge i \in N \\ (\Xi_{1,0}(G,\ell,N)) \wedge (\Xi_{2,0}(G,\ell,N)) \wedge \\ \quad \ldots \wedge (\Xi_{m,0}(G,\ell,N)) & : i = n, \end{cases}$$

$$\Phi_{i,1}(G,\ell,N) = \begin{cases} \forall\, X_{i+1}\colon (\Phi_{i+1,1}(G,N)) & : i < n \wedge i \notin N \\ \neg(\exists\, X_{i+1}\colon (\Phi_{i+1,0}(G,N))) & : i < n \wedge i \in N \\ (\Xi_{1,1}(G,\ell,N)) \vee (\Xi_{2,1}(G,\ell,N)) \vee \\ \quad \ldots \vee (\Xi_{m,1}(G,\ell,N)) & : i = n, \end{cases} \quad (3.2)$$

$$\Xi_{j,0}(G,\ell,N) = \begin{cases} X_{G(j,0)} \subseteq X_{G(j,1)} & : \ell(j) = 0 \\ X_{G(j,0)} \subsetneq X_{G(j,1)} & : \ell(j) = 1 \\ X_{G(j,0)} \in X_{G(j,1)} & : \ell(j) = 2, \end{cases}$$

$$\text{and } \Xi_{j,1}(G,\ell,N) = \begin{cases} X_{G(j,0)} \nsubseteq X_{G(j,1)} & : \ell(j) = 0 \\ \neg(X_{G(j,0)} \subsetneq X_{G(j,1)}) & : \ell(j) = 1 \\ X_{G(j,0)} \notin X_{G(j,1)} & : \ell(j) = 2. \end{cases}$$

Observe that the fact that for every first-order logic formula $\Psi$ it holds that $\neg(\forall\, X\colon \Psi) \iff (\exists\, X\colon \neg\Psi)$, De Morgan's laws, and induction show that for every $n, m \in \mathbb{N}$, every graph $G$ with vertices $\{1, 2, \ldots, n\}$ and edges $\{1, 2, \ldots, m\}$, every function $\ell\colon \{1, 2, \ldots, m\} \to \{0, 1, 2\}$, every set $N \subseteq \{0, 1, \ldots, n-1\}$, and every $i \in \{0, 1, \ldots, n\}$ it holds that

$$\Phi_{i,0}(G,\ell,N) \iff \begin{cases} \exists\, X_{i+1}\colon \exists\, X_{i+2}\colon \ldots \exists\, X_n\colon (\Phi_{n,0}(G,\ell,N)) & : i < n \\ (\Xi_{1,0}(G,\ell,N)) \wedge (\Xi_{2,0}(G,\ell,N)) \wedge \\ \quad \ldots \wedge (\Xi_{m,0}(G,\ell,N)) & : i = n, \end{cases}$$

$$\text{and } \Phi_{i,1}(G,\ell,N) \iff \begin{cases} \neg(\exists\, X_{i+1}\colon \exists\, X_{i+2}\colon \ldots \exists\, X_n\colon (\Phi_{n,0}(G,\ell,N))) & : i < n \\ (\Xi_{1,1}(G,\ell,N)) \vee (\Xi_{2,1}(G,\ell,N)) \vee \\ \quad \ldots \vee (\Xi_{m,1}(G,\ell,N)) & : i = n. \end{cases} \quad (3.3)$$

In particular, it holds that

$$\Phi_{0,0}(G,\ell,N) \iff \exists\, X_1, X_2, \ldots, X_n\colon (\Xi_{1,1}(G,\ell,N)) \wedge (\Xi_{2,1}(G,\ell,N)) \wedge \\ \ldots \wedge (\Xi_{m,1}(G,\ell,N)). \quad (3.4)$$

Theorem 2.18 therefore shows that for every $n, m \in \mathbb{N}$, every graph $G$ with vertices $\{1, 2, \ldots, n\}$ and edges $\{1, 2, \ldots, m\}$, every function $\ell\colon \{1, 2, \ldots, m\} \to \{0, 1, 2\}$, and every $N \subseteq \{0, 1, \ldots, n-1\}$ it holds that $\Phi_{0,0}(G, \ell, N)$ is true (in Zermelo–Fraenkel set theory) if and only if it holds for all $v \in \{1, 2, \ldots, n\}$, $l \in \mathbb{N}$, $e = (e_1, e_2, \ldots, e_l) \in P_G(v, v)$, $j \in \{1, 2, \ldots, l\}$ that $\ell(e_j) = 0$, i.e., if the only cycles in the graph $G$ consist of edges labelled 0. We may therefore take the minimum number of edges in a cycle that does not consist only of edges labelled 0 as a measure for how hard it is to prove the corresponding statement wrong. For every $n, m \in \mathbb{N}$, every graph $G$ with vertices $\{1, 2, \ldots, n\}$ and edges $\{1, 2, \ldots, m\}$, and every function $\ell \colon \{1, 2, \ldots, m\} \to \{0, 1, 2\}$ we denote by $\beta(G, \ell) \in \mathbb{N} \cap \{\infty\}$ the number which satisfies

$$\beta(G, \ell) = \min\left(\left\{ l \in \mathbb{N} \colon \begin{bmatrix} \exists e_1, e_2, \ldots, e_l \in \{1, 2, \ldots, m\}, \\ j \in \{1, 2, \ldots, l\}, v \in \{1, 2, \ldots, n\} \colon \\ (e_1, e_2, \ldots, e_l) \in P_G(v, v) \land \ell(e_j) > 0 \end{bmatrix} \right\} \cup \{\infty\}\right). \quad (3.5)$$

## 3.2 The test sets

Using the construction described in Section 3.1 we can generate a first-order logic sentence from a labelled graph and a set of locations for negations. Our test sets thus consist of tuples $(G, \ell, N)$ where, for some $n, m \in \mathbb{N}$, $G$ is a graph with vertices $\{1, 2, \ldots, n\}$ and edges $\{1, 2, \ldots, m\}$, $\ell \colon \{1, 2, \ldots, m\} \to \{0, 1, 2\}$ is a function, and $N \subseteq \{0, 1, \ldots, n-1\}$, which are then translated into first-order logic sentences $\Phi_{0,0}(G, \ell, N)$. Intuitively, the complexity of the sentence $\Phi_{0,0}(G, \ell, N)$ derived from such a tuple $(G, \ell, N)$ increases with

- the number $n$ of vertices of the graph $G$ (i.e., the number of variables quantified over in $\Phi_{0,0}(G, \ell, N)$),

- the number $m$ of edges of the graph $G$ (i.e., the number of conjuncts or disjuncts in $\Phi_{0,0}(G, \ell, N)$),

- the set $\ell(\{1, 2, \ldots, m\})$ of labels of $G$ (i.e., the types of relations used in $\Phi_{0,0}(G, \ell, N)$),

- the number $|N|$ (i.e., the number of negations in $\Phi_{0,0}(G, \ell, N)$), and

- the number $\beta(G, \ell)$ in case $\beta(G, \ell) \in \mathbb{N}$ (i.e., in case $\Phi_{0,0}(G, \ell, N)$ is false, the minimum number of relations that need to be considered for a disproof of $\Phi_{0,0}(G, \ell, N)$).

We avoid graphs which have multiple edges between the same pair of vertices with the same label, as these result in sentences with repeated terms in the conjunction or disjunction.

We thus denote for every $n, m, k, b \in \mathbb{N}$, $L \subseteq \{0, 1, 2\}$ by $\Gamma_{L,k,b}^{n,m}$ the set which satisfies

$$\Gamma_{L,k,b}^{n,m} = \left\{ (G, \ell, N) \colon \begin{bmatrix} G \in \Gamma_{\{1,2,\ldots,n\},\{1,2,\ldots,m\}}, \ell \colon \{1, 2, \ldots, m\} \to L, \\ N \subseteq \{0, 1, \ldots, n-1\}, |N| = k, \beta(G, \ell) = b, \\ \{1, 2, \ldots, m\} \ni e \mapsto (G(e, 0), G(e, 1), \ell(e)) \\ \in \{1, 2, \ldots, n\}^2 \times \{0, 1, 2\} \text{ is injective} \end{bmatrix} \right\}. \quad (3.6)$$

Note that it holds for all $n, m, k \in \mathbb{N}$, $L \subseteq \{0, 1, 2\}$ that

$$\left\{ (G, \ell, N) \in \bigcup_{b \in \mathbb{N} \cup \{\infty\}} \Gamma_{k,L,b}^{n,m} \colon \Phi_{0,0}(G, \ell, N) \text{ is true in ZF} \right\} = \Gamma_{k,L,\infty}^{n,m}. \quad (3.7)$$

### 3.2.1 The membership grid test set

The membership grid test set is designed to enable a fine-grained evaluation of the influence of the number of variables and the number of conjuncts on the performance of LLMs in judging the truth of first-order logic sentences of the form

$$\exists\, X_1, X_2, \ldots, X_n \colon (X_{i_1} \in X_{j_1} \wedge X_{i_2} \in X_{j_2} \wedge \ldots \wedge X_{i_m} \in X_{j_m}) \tag{3.8}$$

for $n, m \in \mathbb{N}$, $i_1, i_2, \ldots, i_m, j_1, j_2, \ldots, j_m \in \{1, 2, \ldots, n\}$.

The membership grid test set consists of 5088 items. More precisely, it contains for every $n \in \{3, 4, \ldots, 15\}$, $m \in \mathbb{N} \cap [2, \min(10, \frac{n(n-1)}{2})]$, 24 items from $\Gamma^{n,m}_{\{2\},0,\infty}$ (chosen independently and uniformly at random),[3] 12 items from $\bigcup_{b=1}^{\infty} \Gamma^{n,m}_{\{2\},0,b}$ (chosen independently and uniformly at random), and 12 items from $\bigcup_{b=2}^{\infty} \Gamma^{n,m}_{\{2\},0,b}$ (chosen independently and uniformly at random).

This results in 5088 first-order logic sentences of the form

$$\exists\, X_1, X_2, \ldots, X_n \colon (X_{i_1} \in X_{j_1} \wedge X_{i_2} \in X_{j_2} \wedge \ldots \wedge X_{i_m} \in X_{j_m}) \tag{3.9}$$

for $n \in \{3, 4, \ldots, 15\}$, $m \in \mathbb{N} \cap [2, \min(10, \frac{n(n-1)}{2})]$, $i_1, i_2, \ldots, i_m, j_1, j_2, \ldots, j_m \in \{1, 2, \ldots, n\}$. For each $n \in \{3, 4, \ldots, 15\}$, $m \in \mathbb{N} \cap [2, \min(10, \frac{n(n-1)}{2})]$ there are 48 such sentences with $n$ variables and $m$ conjuncts, half of which are true in ZF and half of which are false in ZF. The sentences contain no negations and use only the set membership relation. Among the false sentences, half are chosen such that they do not contain a conjunct of the form $X \in X$ for any variable $X$.

### 3.2.2 The membership grid test set with negation

The membership grid test set with negations is constructed along the same lines as the membership grid test set (see Section 3.2.1), but with each resulting first-order logic sentence containing exactly one negation, enabling us to evaluate the influence of the presence of negations on the performance of LLMs.

More precisely, for every $n \in \{3, 4, \ldots, 15\}$, $m \in \mathbb{N} \cap [2, \min(10, \frac{n(n-1)}{2})]$, 24 items were chosen (independently and uniformly at random) from $\Gamma^{n,m}_{\{2\},0,\infty}$, 12 items were chosen (independently and uniformly at random) from $\bigcup_{b=1}^{\infty} \Gamma^{n,m}_{\{2\},0,b}$, and 12 items were chosen (independently and uniformly at random) from $\bigcup_{b=2}^{\infty} \Gamma^{n,m}_{\{2\},0,b}$.

This results in 5088 first-order logic sentences of the form

$$\begin{cases} \exists\, X_1, X_2, \ldots, X_k \colon \neg(\forall\, X_{k+1}, X_{k+2}, \ldots, X_n \colon (X_{i_1} \notin X_{j_1} \\ \qquad\qquad\qquad\qquad \vee X_{i_2} \notin X_{j_2} \vee \ldots \vee X_{i_m} \notin X_{j_m})) & : k > 0 \\ \neg(\forall\, X_1, X_2, \ldots, X_n \colon (X_{i_1} \notin X_{j_1} \vee X_{i_2} \notin X_{j_2} \vee \ldots \vee X_{i_m} \notin X_{j_m})) & : k = 0 \end{cases} \tag{3.10}$$

for $n \in \{3, 4, \ldots, 15\}$, $m \in \mathbb{N} \cap [2, \min(10, \frac{n(n-1)}{2})]$, $k \in \{0, 1, \ldots, n-1\}$, $i_1, i_2, \ldots, i_m, j_1, j_2, \ldots, j_m \in \{1, 2, \ldots, n\}$. For each $n \in \{3, 4, \ldots, 15\}$, $m \in \mathbb{N} \cap [2, \min(10, \frac{n(n-1)}{2})]$, there

---

[3] Note that for all $n, m \in \mathbb{N}$ with $m > \frac{n(n-1)}{2}$ it holds that $\Gamma^{n,m}_{\{2\},0,\infty} = \emptyset$ (an acyclic unlabelled directed graph with $n$ vertices and at most one edge from any vertex to any other can have at most $\frac{n(n-1)}{2}$ edges). Equivalently (by Corollary 2.20), any sentence of the form (3.8) with $m > \frac{n(n-1)}{2}$ pairwise different conjuncts is false in ZF.

are 48 such sentences, half of which are true in ZF and half of which are false in ZF. The sentences contain exactly one negation and use only the set membership relation. Among the false sentences, half are chosen such that they do not contain a disjunct of the form $Y \notin Y$ for any variable $Y$.

### 3.2.3 The all-relations grid test set

The all-relations grid test set is designed to enable a fine-grained evaluation of the influence of the number of variables and the number of conjuncts on the performance of LLMs in judging the truth of first-order logic sentences of the form

$$\exists\, X_1, X_2, \ldots, X_n \colon (X_{i_1} R_1 X_{j_1} \wedge X_{i_2} R_2 X_{j_2} \wedge \ldots \wedge X_{i_m} R_m X_{j_m}) \tag{3.11}$$

for some $n, m \in \mathbb{N}$, $i_1, i_2, \ldots, i_m, j_1, j_2, \ldots, j_m \in \{1, 2, \ldots, n\}$, and relations $R_1, R_2, \ldots, R_m$ chosen among set membership, subset, and proper subset. Comparison with the membership grid test set (see Section 3.2.1) furthermore allows us to evaluate the influence of the presence of different types of relations on the performance of LLMs.

The all-relations grid test set consists of 5568 items. For every $n \in \{3, 4, \ldots, 15\}$, $m \in \mathbb{N} \cap [2, \min(10, \frac{3n(n-1)}{2})]$, 24 items were chosen (independently and uniformly at random) from $\Gamma^{n,m}_{\{0,1,2\},0,\infty}$,[4] 12 items were chosen (independently and uniformly at random) from $\bigcup_{b=1}^{\infty} \Gamma^{n,m}_{\{0,1,2\},0,b}$, and 12 items were chosen (independently and uniformly at random) from $\bigcup_{b=2}^{\infty} \Gamma^{n,m}_{\{0,1,2\},0,b}$.

This results in 5568 first-order logic sentences of the form

$$\exists\, X_1, X_2, \ldots, X_n \colon (X_{i_1} R_1 X_{j_1} \wedge X_{i_2} R_2 X_{j_2} \wedge \ldots \wedge X_{i_m} R_m X_{j_m}) \tag{3.12}$$

for some $n \in \{3, 4, \ldots, 15\}$, $m \in \mathbb{N} \cap [2, \min(10, 3\frac{n(n-1)}{2})]$, $i_1, i_2, \ldots, i_m, j_1, j_2, \ldots, j_m \in \{1, 2, \ldots, n\}$, and relations $R_1, R_2, \ldots, R_m$ chosen among set membership, subset, or proper subset. For each $n \in \{3, 4, \ldots, 15\}$, $m \in \mathbb{N} \cap [2, \min(10, \frac{3n(n-1)}{2})]$, there are 48 such sentences with $n$ variables and $m$ conjuncts, half of which are true in ZF and half of which are false in ZF. The sentences contain no negations. Among the false sentences, half are chosen such that they do not contain a conjunct of the form $Y \in Y$ or $Y \subsetneq Y$ for any variable $Y$.

### 3.2.4 The mixed test set

The mixed test set contains a wide range of items, giving a single benchmark number for the performance of LLMs on a diverse set of first-order logic sentences of the kind described in Section 3.1. We use this to evaluate the influence of other factors, such as the encoding of the first-order logic sentences (e.g., in Unicode symbols, LaTeX notation, or words; see Section 3.3.1) or the prompt design (e.g., a simple yes/no prompt vs. a *chain of thought* (CoT) prompt or 0-shot vs. fewshot prompting; see Section 3.3.2) on the performance of LLMs.

The mixed test set consists of 756 elements, of which

---

[4]Note that for all $n, m \in \mathbb{N}$ with $m > \frac{3n(n-1)}{2}$ it holds that $\Gamma^{n,m}_{\{0,1,2\},0,\infty} = \emptyset$ (an acyclic directed graph with $n$ vertices and $l$ possible edge labels where no two edges with the same start vertices and end vertices, respectively, have identical labels, can have at most $\frac{ln(n-1)}{2}$ edges)

(i) for every $n \in \{3, 4, ..., 8\}$, $m \in \{2, 3, ..., \min(8, \frac{n(n-1)}{2})\}$, $L \in \{\{1\}, \{2\}\}$ two items are chosen from $\Gamma_{L,0,\infty}^{n,m}$ (uniformly at random), and two from $\bigcup_{b \in \mathbb{N}} \Gamma_{L,0,b}^{n,m}$ (with lower weight on items from $\Gamma_{L,0,b}^{n,m}$),

(ii) for every $n \in \{3, 4, ..., 8\}$, $m \in \{2, 3, ..., \min(8, \frac{n(n-1)}{2})\}$, $L \in \{\{1\}, \{2\}\}$ one item is chosen from $\Gamma_{L,1,\infty}^{n,m} \cup \Gamma_{L,2,\infty}^{n,m}$ (uniformly at random), and one from $\bigcup_{b \in \mathbb{N}} (\Gamma_{L,1,b}^{n,m} \cup \Gamma_{L,2,b}^{n,m})$ (with lower weight on items from $\Gamma_{L,1,1}^{n,m} \cup \Gamma_{L,2,1}^{n,m}$),

(iii) for every $n \in \{3, 4, ..., 8\}$, $m \in \{2, 3, ..., \min(8, \frac{n(n-1)}{2})\}$ three items are chosen from $\Gamma_{\{0,1,2\},0,\infty}^{n,m}$ (uniformly at random), and three from $\bigcup_{b \in \mathbb{N}} \Gamma_{L,0,b}^{n,m}$ (with lower weight on items from $\Gamma_{\{0,1,2\},0,b}^{n,m}$), and

(iv) for every $n \in \{3, 4, ..., 8\}$, $m \in \{2, 3, ..., \min(8, \frac{n(n-1)}{2})\}$ one item is chosen from $\Gamma_{\{0,1,2\},1,\infty}^{n,m} \cup \Gamma_{\{0,1,2\},2,\infty}^{n,m}$ (uniformly at random), and one from $\bigcup_{b \in \mathbb{N}} (\Gamma_{\{0,1,2\},1,b}^{n,m} \cup \Gamma_{\{0,1,2\},2,b}^{n,m})$ (with lower weight on items from $\Gamma_{\{0,1,2\},1,1}^{n,m} \cup \Gamma_{\{0,1,2\},2,1}^{n,m}$).

This results in 756 first-order logic sentences of the kind defined in Section 3.1, half of them true in ZF and half of them false in ZF, with 3 to 8 variables and 2 to 8 operands in the conjunction/disjunction. Of these 756 sentences

(i) 140 use only the set membership relation and contain no negations,

(ii) 140 use only the proper subset relation and contain no negations,

(iii) 70 use only the set membership relation and contain either one or two negations,

(iv) 70 use only the proper subset relation and contain either one or two negations,

(v) 252 may use any of the relations set membership, subset, and proper subset and contain no negations,

(vi) 84 may use any of the relations set membership, subset, and proper subset and contain either one or two negations.

### 3.2.5 The hard test set

The hard test set, like the mixed test set contains a wide range of items, but is designed to be more challenging for LLMs in that the resulting first-order logic sentences contain more variables, more operands in the conjunction/disjunction, and more negations, are not restricted to partciular types of relations, and have a higher minimum number of relations that need to be considered for a disproof in case the sentence is false.

The hard test set consists of 200 elements, of which 100 are chosen from

$$\bigcup_{n=6}^{12} \bigcup_{m=n}^{14} \bigcup_{k=0}^{4} \Gamma_{\{0,1,2\},k,\infty}^{n,m} \tag{3.13}$$

and 100 are chosen from

$$\bigcup_{n=6}^{12} \bigcup_{m=n}^{14} \bigcup_{k=0}^{4} \bigcup_{b=3}^{\infty} \Gamma_{\{0,1,2\},k,b}^{n,m}. \tag{3.14}$$

This results in 200 first-order logic sentences of the kind defined in Section 3.1, half of them true in ZF and half of them false in ZF, with 6 to 12 variables, 6 to 14 operands in the conjunction/disjunction, and containing 0 to 4 negations. All of the sentences may use any of the relations set membership, subset, and proper subset. For the false sentences, the minimum number of relations that need to be considered for a disproof is at least 3 (weighted with a bias towards higher values).

### 3.3 Testing conditions

While the test sets described in Section 3.2 above provide the first-order logic sentences to be evaluated, several choices have to be made regarding the prompting of the LLMs.

#### 3.3.1 Encodings for first-order logic formulae

In our tests, we employed four different ways of encoding first-order logic formulae for the prompt. As an example, consider the following formula (which is one of the sentences in the mixed test set):

$$\neg(\forall A\colon (\neg(\exists B, C\colon ((A \in C) \land (B \subsetneq A) \land (C \subseteq B) \land (C \subseteq C))))) \tag{3.15}$$

The following encodings were employed:

**Unicode** The formula is given in first-order logic notation with mathematical symbols encoded in Unicode. The formula in (3.15) is encoded as follows:

¬ (∀ A: (¬ (∃ B, C: ((A ∈ C) ∧ (B ⊊ A) ∧ (C ⊆ B) ∧ (C ⊆ C)))))

**Unicode (fewer parentheses)** The formula is given in first-order logic notation with mathematical symbols encoded in Unicode as in the Unicode encoding, but with the minimum number of parentheses necessary under the standard rules of operator precedence. The formula in (3.15) is encoded as follows:

¬ ∀ A: ¬ ∃ B, C: (A ∈ C ∧ B ⊊ A ∧ C ⊆ B ∧ C ⊆ C)

**LaTeX** The formula is given in first-order logic notation with mathematical symbols encoded in LaTeX. The formula in (3.15) is encoded as follows:

\neg (\forall A: (\neg (\exists B, C: ((A \in C) \land (B \subsetneq A) \land (C \subseteq B) \land (C \subseteq C)))))

**words** The formula is written out in natural language. The formula in (3.15) is encoded as follows:

It does not hold that for every set A it holds that there do not exist sets B and C such that A is an element of C, B is a proper subset of A, C is a subset of B, and C is a subset of C

### 3.3.2 Prompting strategies

In our tests, we employed different prompting strategies. These differ, first, in the formulation of the system message and user instructions supplied to the LLM and, second, in whether the LLM was given only the instructions along with the first-order logic sentence in question or was also given examples of correct and incorrect answers to similar questions.

**3.3.2.1 Instructions** We employed three different sets of system messages and instructions for the LLMs, simple yes/no prompting, a simple CoT prompt (cf., for instance, [32, 74]), and a more elaborate CoT prompt:

**yes/no** In this strategy, the system message was kept as simple as possible and the prompt demanded a simple yes/no answer:

> System: `You are a helpful mathematical assistant.`
>
> User: `Is the following statement true in Zermelo-Fraenkel set theory?`
>    `[statement]`
>    `Answer only 'yes' or 'no'!`

**CoT** In this condition, the system message was kept as simple as possible and the prompt demanded step-by-step reasoning:

> System: `You are a helpful mathematical assistant.`
>
> User: `Is the following statement true in Zermelo-Fraenkel set theory?`
>    `[statement]`
>    `Proceed step-by-step and reason methodically.`

**CoT (complex)** In this condition, a more complex system message was used, employing common prompting techniques such as having the LLM assume the persona of a skilled expert, asking the LLM to lay out its thought process step-by-step, and encouraging it to consider both a proof and disproof of the statement in question:

> System: `You are a professor of mathematics specializing in`
>    `mathematical logic. Your audience are graduate math students,`
>    `who understand mathematical notation. When asked a question, lay`
>    `out your thought process step-by-step in a systematic but`
>    `succinct way, employing your deep knowledge of logic and set`
>    `theory and your considerable experience and skill in applying`
>    `common proof techniques. Consider carefully how to prove or`
>    `disprove the statement. Do not draw a conclusion until you are`
>    `satisfied you have thought through the problem sufficiently, but`
>    `do continue until you are satisfied and draw a conclusion at the`
>    `end of your answer, indicating clearly whether you believe the`
>    `statement to be true or not.`

> User: `Is the following statement true in Zermelo-Fraenkel set theory? [statement]`

The CoT prompts were deliberately chosen in such a way that they largely leave it to the LLM to decide how to structure the answer. We extracted a final yes/no answer from the LLM's output in these conditions, by feeding it to another LLM. We used GPT-4o mini for this purpose. Using a small model for this task instead of the same model that generated the original answer has the advantage of being faster and cheaper. The task of extracting a final answer from the LLM's output is a text summarization/classification task that even small LLMs typically excel at. One more complication in the CoT and CoT (complex) conditions is that the LLMs' responses cannot always be summarized as claiming the given logical statement to be true or false. Instead, we categorized responses into four classes:

**true** The LLM's response comes to the conclusion that the statement is true or likely true.

**false** The LLM's response comes to the conclusion that the statement is false or likely false.

**vague** The LLM's response does not come to any clear conclusion. This includes cases where the LLM's response states that it was unable to decide whether the statement is true or false and more analysis is needed as well as cases where the LLM's response starts looping or cannot be interpreted as either claiming that the statement is true or claiming that it is false.

**undecidable** The LLM's response states that it is impossible to decide whether the statement is true or false. Most commonly, this takes the form of the LLM claiming that the information given is insufficient to decide the truth or falsity of the statement or that the truth of the statement "depends on the values of the variables". There were also cases where an LLM claims something to the effect that the statement is independent of the Zermelo–Fraenkel axioms (e.g., that neither the statement nor its negation can be proven from the Zermelo–Fraenkel axioms or that there exist models of ZF where the statement is true as well as models where the statement is false). In rare cases, an LLM claimed that the statement is not syntactically well-formed.

What separates these answers from the 'vague' category is that they are potentially valid responses in the sense that it is indeed impossible to decide the truth or falsity of a syntactically malformed statement or of a statement that has free variables or of a statement that is independent of the axioms of Zermelo–Fraenkel set theory. Still, since no such statements were included in the test set, such answers are inevitably incorrect.

The exact prompt used for the classification by GPT-4o mini was as follows:

> System: `You will be given a mathematical question that asks whether a statement is true or false and an answer. Your job is to summarize the answer. If the answer concludes that the statement is definitely`

```
        true (in Zermelo-Fraenkel set theory/ZF/ZFC), respond TRUE, if the
        answer concludes that the statement is definitely false (in
        Zermelo-Fraenkel set theory/ZF/ZFC), respond FALSE. If the answer
        claims that it is mathematically impossible to decide whether the
        statement is true, respond UNDECIDABLE. If the answer comes to no
        definite conclusion, respond VAGUE. The answer may be wrong or
        contain mistakes, however, do NOT attempt to answer the question
        yourself! Just summarize the provided answer! Respond only with one
        of the words TRUE, FALSE, UNDECIDABLE, or VAGUE.
 User:  QUESTION:
        Is the following statement true?
        [statement]
        ===========================================
        ANSWER:
        [answer as provided by the LLM under test]
        ===========================================
        Summarize the answer!
```

After the automatic classification by GPT-4O MINI, the classification for all answers that were classified as vague or undecidable, was double-checked manually. All classifications where GPT-4O MINI was "uncertain", as measured by the log probability of the classification being below a threshold of $-10^{-5}$, were also double-checked manually.

**3.3.2.2  0-shot vs. few-shot prompting**  We employed three different strategies for providing the LLM with examples of correct and incorrect answers to similar questions:

**0-shot**  In this condition, no additional context was provided beyond the system message and user instructions as described in Section 3.3.2.1.

**3-shot (with yes/no prompting)**  In this condition, the LLM was asked to determine the truth of the logical statement in ZF with three examples as additional context. The logical statements for the examples were generated at random as part of the test set along with the statements to be evaluated. For each prompt, one of the examples contains a true statement, one contains a false statement, and the third statement is chosen to be true or false with equal probabilities. The example statements are chosen to be similar to the statement to be evaluated. More precisely, if the statement to be evaluated contained negations, the examples contained negations as well, and if the statement to be evaluated contained only one type of relations (i.e., only set membership relations or only proper subset relations), the examples contained only that type of relations as well (and conversely, if the statement to be evaluated had no restrictions on the type of relations used, neither did the examples). This type of few-shot prompting was only used in conjunction with simple yes/no prompts (cf. Section 3.3.2.1).

The prompt was of thus of following form:

```
System: You are a helpful mathematical assistant.
User: Is the following statement true in Zermelo-Fraenkel set theory?
    [true statement]
    Answer only 'yes' or 'no'!
Assistant: Yes
User: Is the following statement true in Zermelo-Fraenkel set theory?
    [false statement]
    Answer only 'yes' or 'no'!
Assistant: No
User: Is the following statement true in Zermelo-Fraenkel set theory?
    [statement]
    Answer only 'yes' or 'no'!
Assistant: [Yes/No]
User: Is the following statement true in Zermelo-Fraenkel set theory?
    [statement]
    Answer only 'yes' or 'no'!
```

**2-shot (with CoT prompting)** In this condition, the LLM was asked to determine the truth of the logical statement in ZF with two examples as additional context. This type of fewshot prompting was only employed in conjunction with CoT instructions (cf. Section 3.3.2.1). The examples were drawn from a pool of 12 questions with answers that either prove or disprove the statement, written by hand. For each prompt, one of the examples contained a true statement and the other a false statement. The examples were chosen such that they were similar to the statement to be evaluated. More precisely, if the statement to be evaluated contained negations, the examples contained negations as well, and if the statement to be evaluated contained only one type of relation (i.e., only set membership relations or only proper subset relations), the examples contained only that type of relation as well (and conversely, if the statement to be evaluated had no restrictions on the type of relations used, neither did the examples). In each case, if the statement contains negations, the answer proceeds by noting that an equivalent statement without negations can be obtained by pushing the negations through the quantifiers to the right. This results in a statement that contains only existential quantification, which is then either proved by providing a concrete example or disproved by showing that the statement either implies the existence of a cycle in the membership relation (thus contradicting the Axiom of Regularity) or the existence of a cycle in the proper subset relation (thus contradicting the Axiom of Extensionality). In some of the cases where a concrete example was given to prove the statement, its construction was described in a systematic fashion suggestive of a general method.

One example of a prompt in this condition is the following:

```
System: You are a helpful mathematical assistant.
User: Is the following statement true in Zermelo-Fraenkel set theory?
```

> ∃ A: (¬ (∀ B, C: (¬ (∃ D, E: ((D ∈ B) ∧ (B ⊆ E) ∧ (E ⊆ A) ∧ (A ⊆ B) ∧ (D ⊊ E) ∧ (A ⊊ C) ∧ (E ∈ C))))))
> Proceed step-by-step and reason methodically.
>
> Assistant: We first simplify the statement by pushing the negations through to the right. This yields the equivalent statement ∃ A, B, C, D, E: ((D ∈ B) ∧ (B ⊆ E) ∧ (E ⊆ A) ∧ (A ⊆ B) ∧ (D ⊊ E) ∧ (A ⊊ C) ∧ (E ∈ C)). Now let D be the empty set. Having defined D, let B, E, and A be the sets such that B = E = A = D ∪ {D}. Then D ∈ B and D ⊆ E. Since D ∉ D, we have that D ⊊ E. Having defined A and E, let C = A ∪ {A, E}. Then E ∈ C and A ⊆ C. Since A ∈ C but A ∉ A, we obtain that A ⊊ C. Hence, the sets we have defined satisfy that ((D ∈ B) ∧ (B ⊆ E) ∧ (E ⊆ A) ∧ (A ⊆ B) ∧ (D ⊊ E) ∧ (A ⊊ C) ∧ (E ∈ C)). Therefore the simplified statement is true, and since the simplified statement is equivalent to the original statement, the original statement is also true.
>
> User: Is the following statement true in Zermelo-Fraenkel set theory?
> ¬ (∀ A, B, C, D, E, F, G, H: ((¬ (B ⊊ E)) ∨ (E ∉ D) ∨ (D ⊈ E) ∨ (¬ (E ⊊ C)) ∨ (B ⊈ E)))
> Proceed step-by-step and reason methodically.
>
> Assistant: We first simplify the statement by pushing the negation through to the right. This yields the equivalent statement ∃ A, B, C, D, E, F, G, H: ((B ⊊ E) ∧ (E ∈ D) ∧ (D ⊆ E) ∧ (E ⊊ C) ∧ (B ⊆ E)). Assume for the sake of contradiction that the original statement and hence this simplified statement is true. Let A, B, C, D, E, F, G, H be sets such that B ⊊ E, E ∈ D, D ⊆ E, E ⊊ C, and B ⊆ E. Since E ∈ D and D ⊆ E, we obtain that E ∈ E, which is a contradiction to the axiom of regularity. This contradiction establishes that the statement is false.
>
> User: Is the following statement true in Zermelo-Fraenkel set theory?
> ¬ (∀ A, B, C: ((A ⊈ A) ∨ (A ∉ B) ∨ (C ⊈ C) ∨ (A ⊈ B) ∨ (C ∉ A) ∨ (¬ (A ⊊ B)) ∨ (¬ (B ⊊ C)) ∨ (C ⊈ A)))
> Proceed step-by-step and reason methodically.

# 4 Evaluating large language models (LLMs) on first-order logic questions

In this section we present the results of our evaluations of the performance of the LLMs on the test sets described in Section 3.2 under the conditions described in Section 3.3. The following LLMs were evaluated:

**Claude.** From Anthropic's Claude family of LLMs (cf. [4]) we tested the following models:

- CLAUDE 3 HAIKU in the version from 07/03/2024 as supplied on AWS Bedrock under model ID `anthropic.claude-3-haiku-20240307-v1:0` (identical to the version supplied by Anthropic itself).

- CLAUDE 3.5 HAIKU in the version from 22/10/2024 as supplied on AWS Bedrock under model ID `anthropic.claude-3-5-haiku-20241022-v1:0` (identical to the version supplied by Anthropic itself).

- CLAUDE 3.5 SONNET in the versions from 20/06/2024 and 22/10/2024 as supplied on AWS Bedrock under model IDs `anthropic.claude-3-5-sonnet-20240620-v1:0` and `anthropic.claude-3-5-sonnet-20241022-v2:0`, respectively (identical to the versions supplied by Anthropic itself).

**DeepSeek.** We tested the following models by DeepSeek:

- DEEPSEEK-V3 (cf. [10]) as supplied on DeepSeek's API under model ID `deepseek-chat`.

- DEEPSEEK-R1 (cf. [11]) as supplied on DeepSeek's API under model ID `deepseek-reasoner`.

**Gemini.** From Google DeepMind's Gemini family of LLMs (cf. [19]) we tested the following models:

- GEMINI 1.5 FLASH 8B in the version from October 2024 as supplied on Google's Gemini API under model ID `gemini-1.5-flash-8b-001`.

- GEMINI 1.5 FLASH in the versions from May 2024 and September 2024 as supplied on Google's Gemini API under model IDs `gemini-1.5-flash-001` and `gemini-1.5-flash-002`, respectively.

- GEMINI 2.0 FLASH in the version from January 2025 as supplied on Google's Gemini API under model ID `gemini-2.0-flash-001`.

- GEMINI 1.5 PRO in the versions from February 2024 and September 2024 as supplied on Google's Gemini API under model IDs `gemini-1.5-pro-001` and `gemini-1.5-pro-002`, respectively.

- GEMINI 2.0 PRO EXPERIMENTAL in the version from 05/02/2025 as supplied on Google's Gemini API under model ID `gemini-2.0-pro-exp-02-05`.

- GEMINI 2.0 FLASH THINKING EXPERIMENTAL in the version from 21/01/2025 as supplied on Google's Gemini API under model ID `gemini-2.0-flash-thinking-exp-01-21`.

**GPT/o1/o3.** From OpenAI's GPT (cf. [53, 54]), o1 (cf. [55]), and o3 (cf. [52]) families of LLMs we tested the following models:

- GPT-3.5 TURBO INSTRUCT in the version from September 2023 as supplied on OpenAI's API under model ID `gpt-3.5-turbo-instruct`.

- GPT-4O MINI in the version from 18/07/2024 as supplied on OpenAI's API under model ID `gpt-4o-mini-2024-07-18`.

- GPT-4O in the versions from 13/05/2024, 06/08/2024, and 20/11/2024 as supplied on OpenAI's API under model IDs `gpt-4o-2024-05-13`, `gpt-4o-2024-08-06`, and `gpt-4o-2024-11-20`, respectively.

- O1-MINI in the version from 12/09/2024 as supplied on OpenAI's API under model ID `o1-mini-2024-09-12`.

- O1 in the preview version from 12/09/2024 as well as the final version from 17/12/2024 as supplied on OpenAI's API under model IDs `o1-preview-2024-09-12` and `o1-2024-12-17`, respectively.

- O3-MINI in the version from 31/01/2025 as supplied on OpenAI's API under model ID `o3-mini-2025-01-31`.

**Llama.** From Meta's Llama 3 family of LLMs (cf. [23]) we tested the following models:

- LLAMA 3.1 8B in the version from July 2024 as supplied on AWS Bedrock under model ID `meta.llama3-1-8b-instruct-v1:0`.

- LLAMA 3.1 70B in the version from July 2024 as supplied on AWS Bedrock under model ID `meta.llama3-1-70b-instruct-v1:0`.

- LLAMA 3.1 405B in the version from July 2024 as supplied on AWS Bedrock under model ID `meta.llama3-1-405b-instruct-v1:0`.

- LLAMA 3.3 70B in the version from December 2024 as supplied on AWS Bedrock under model ID `meta.llama3-3-70b-instruct-v1:0`.

**Mistral.** We tested the following models by Mistral AI:

- MISTRAL NEMO in the version from July 2024 as supplied on Mistral AI's API under model ID `open-mistral-nemo-2407`.

- MISTRAL SMALL in the versions from September 2024 and January 2025 as supplied on Mistral AI's API under model IDs `mistral-small-2409` and `mistral-small-2501`, respectively.

- MISTRAL LARGE in the versions from July 2024 and November 2024 as supplied on AWS Bedrock under model ID `mistral.mistral-large-2407-v1:0` (identical to the one supplied on Mistral AI's API) and on Mistral AI's API under model id `mistral-large-2411`, respectively.

In addition, we ran some tests with early versions of some of the test sets on OpenAI's now deprecated GPT-3.5 class model TEXT-DAVINCI-003. These results are not entirely comparable to the results of the other models due to slight differences in the employed test sets and the phrasing of the prompts.[5] For the sake of reproducibility, we used a temperature of 0.0 for all models.

Note that we did not test all models on all datasets in all testing conditions (mostly due to cost constraints).

---

[5]The earlier versions of the test sets were generated according to the same principles as the final versions, but due to the randomness involved, are not identical. The prompts we used for TEXT-DAVINCI-003 merely asked for the truth or falsity of the statement without specifying "in Zermelo–Fraenkel set theory". Due to the deprecation of the TEXT-DAVINCI-003 model in January 2024, we could not re-run the tests with the final versions of the test sets and prompts. Later tests on other models usually showed only a minor impact on the results from these discrepancies. Still, for the above reasons, the results for TEXT-DAVINCI-003 should not be taken as directly comparable to the results for the other models.

## 4.1 Evaluating LLM performance by formula complexity for statements involving set membership relations

For this series of evaluations, we used the membership grid test set (see Section 3.2.2) to evaluate the influence of the number of variables and the number of conjuncts on the performance of LLMs in judging the truth of first-order logic sentences of the form

$$\exists X_1, X_2, \ldots, X_n \colon ((X_{i_1} \in X_{j_1}) \wedge (X_{i_2} \in X_{j_2}) \wedge \ldots \wedge (X_{i_m} \in X_{j_m})). \qquad (4.1)$$

for $n, m \in \mathbb{N}$, $i_1, i_2, \ldots, i_m, j_1, j_2, \ldots, j_m \in \{1, 2, \ldots, n\}$. For prompting, we used the Unicode encoding for the first-order logic sentences (cf. Section 3.3.1), we used simple yes/no instructions (cf. Section 3.3.2.1) and a 0-shot prompt (cf. Section 3.3.2.2). One example of a prompt as used in this series of evaluations is the following:

```
System: You are a helpful mathematical assistant.

User: Is the following statement true in Zermelo-Fraenkel set theory?
      ∃ A, B, C, D, E, F: ((B ∈ F) ∧ (D ∈ E) ∧ (B ∈ D) ∧ (A ∈ C) ∧ (F ∈
      B))
      Answer only 'yes' or 'no'!
```

Table 4.1 shows the results of our evaluations for all models we tested, listing the percentage of correct answers for each model by the number of conjuncts appearing in the evaluated statements. Figure 4.1 plots these percentages for all current models against the number of conjuncts and Figure 4.2 shows the same data but split up into state-of-the-art frontier models and smaller models, respectively, for easier comparability. The plots in Figure 4.6 plot the percentages of correct answers by number of conjuncts, grouped by model family. Finally, Figures 4.3, 4.4, and 4.5 show the results for all current models, by number of variables and number of conjuncts appearing in the evaluated statements. Each cell in the grids in these figures shows the percentage of correct answers for the 48 statements with the corresponding number of variables and conjuncts.

|  | \multicolumn{9}{c}{Number of conjuncts} |
| Model | 2 | 3 | 4 | 5 | 6 | 7 | 8 | 9 | 10 |
| --- | --- | --- | --- | --- | --- | --- | --- | --- | --- |
| Claude 3 Haiku (2024-03-07) | 54.5 | 53.2 | 54.5 | 53.5 | 50.7 | 51.5 | 54.5 | 52.1 | 54.9 |
| Claude 3.5 Haiku (2024-10-22) | 97.9 | 95.2 | 91.8 | 85.9 | 74.1 | 76.1 | 62.7 | 57.4 | 53.0 |
| Claude 3.5 Sonnet (2024-06-20) | **100.0** | 97.9 | 95.1 | 89.4 | **90.6** | 85.0 | **84.5** | **81.1** | 73.9 |
| Claude 3.5 Sonnet (2024-10-22) | **100.0** | 97.0 | 87.8 | 81.6 | 84.0 | 80.7 | 82.2 | 76.9 | 75.0 |
| DeepSeek-V3 | 99.4 | 98.1 | 89.6 | 85.1 | 81.8 | 73.1 | 72.5 | 71.0 | 68.9 |
| Gemini 1.5 Flash 8b (001) | 72.8 | 60.7 | 61.6 | 54.5 | 54.9 | 52.3 | 52.5 | 52.1 | 51.1 |
| Gemini 1.5 Flash (001) | **100.0** | 95.0 | 82.3 | 70.7 | 70.1 | 66.7 | 63.4 | 61.7 | 62.3 |
| Gemini 1.5 Flash (002) | **100.0** | 98.2 | 92.9 | 88.7 | 87.3 | 82.2 | 78.8 | 78.2 | 78.6 |
| Gemini 2.0 Flash (001) | 66.5 | 58.0 | 54.0 | 51.0 | 51.4 | 51.1 | 52.5 | 51.1 | 51.9 |
| Gemini 1.5 Pro (001) | **100.0** | 99.2 | 96.0 | 85.9 | 84.9 | 79.7 | 72.9 | 71.8 | 70.8 |
| Gemini 1.5 Pro (002) | **100.0** | **99.8** | **97.6** | **93.6** | 89.1 | **87.9** | 83.7 | 77.5 | **78.8** |
| GPT-3.5 text-davinci-003 | 83.7 | 82.9 | 81.1 | 71.0 | 72.2 | 71.0 | 65.3 | 63.1 | 63.8 |
| GPT-3.5 Turbo Instruct | 65.7 | 54.8 | 50.3 | 50.0 | 50.0 | 50.0 | 50.0 | 50.0 | 50.0 |
| GPT-4o mini (2024-07-18) | 80.3 | 62.5 | 57.3 | 58.0 | 59.7 | 56.6 | 58.1 | 57.8 | 59.1 |
| GPT-4o (2024-05-13) | 99.0 | 97.1 | 93.2 | 86.6 | 81.9 | 75.6 | 72.0 | 70.5 | 71.4 |
| GPT-4o (2024-08-06) | 99.8 | 97.4 | 90.8 | 80.7 | 75.9 | 71.8 | 68.4 | 68.6 | 66.9 |
| GPT-4o (2024-11-20) | 99.5 | 97.3 | 91.5 | 84.7 | 80.6 | 77.3 | 73.7 | 72.2 | 70.8 |
| Llama 3.1 8b | 50.5 | 50.5 | 50.2 | 50.0 | 50.0 | 50.0 | 50.0 | 50.0 | 50.0 |
| Llama 3.1 70b | **100.0** | 97.1 | 88.9 | 80.7 | 77.1 | 72.3 | 65.2 | 65.9 | 68.4 |
| Llama 3.1 405b | **100.0** | 98.9 | 97.0 | **93.6** | **90.6** | **88.1** | 81.4 | 77.8 | 72.7 |
| Llama 3.3 70b | **100.0** | 93.9 | 71.0 | 67.9 | 65.8 | 61.6 | 58.3 | 61.6 | 59.7 |
| Mistral Nemo | 53.0 | 53.0 | 52.8 | 57.5 | 54.2 | 54.0 | 50.8 | 55.3 | 53.4 |
| Mistral Small (2024-09) | 73.7 | 56.6 | 55.7 | 55.0 | 55.7 | 53.6 | 52.8 | 53.4 | 51.9 |
| Mistral Small (2025-01) | 93.3 | 54.2 | 53.0 | 51.2 | 50.3 | 50.2 | 50.2 | 50.0 | 50.0 |
| Mistral Large (2024-07) | **100.0** | 98.2 | 83.3 | 69.3 | 64.6 | 61.0 | 59.5 | 61.9 | 58.9 |
| Mistral Large (2024-11) | 99.4 | 63.5 | 53.5 | 53.5 | 52.4 | 50.9 | 51.3 | 51.1 | 50.4 |

Table 4.1: Percentage of correct answers in the set membership test set by number of conjuncts achieved by all tested models

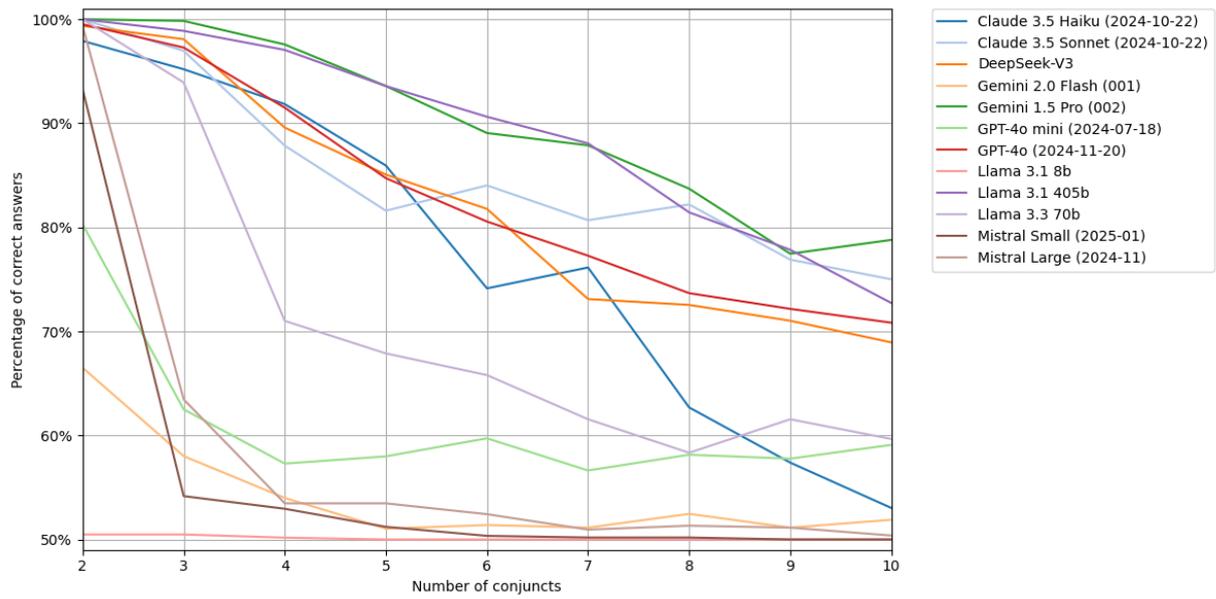

Figure 4.1: Percentage of correct answers in the set membership test set by number of conjuncts achieved by current models

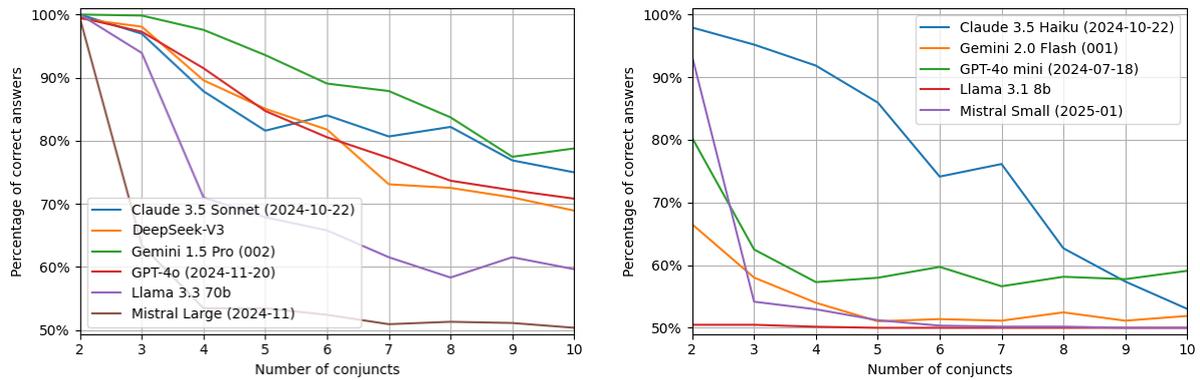

Figure 4.2: Percentage of correct answers in the set membership test set by number of conjuncts achieved by state-of-the-art frontier models (left) and smaller models (right), respectively

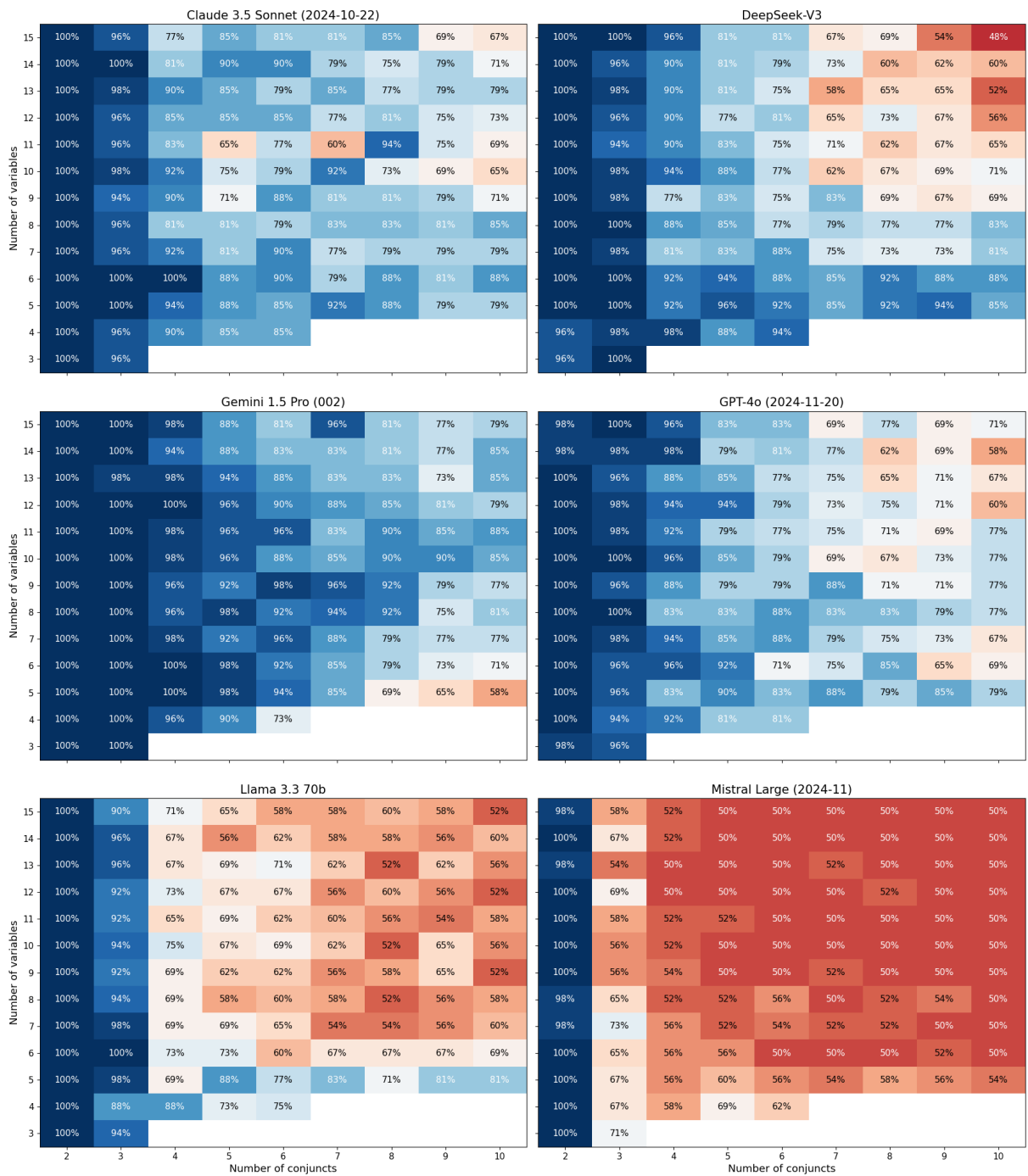

Figure 4.3: Percentage of correct answers in the set membership test set by number of variables and conjuncts achieved by state-of-the-art frontier models

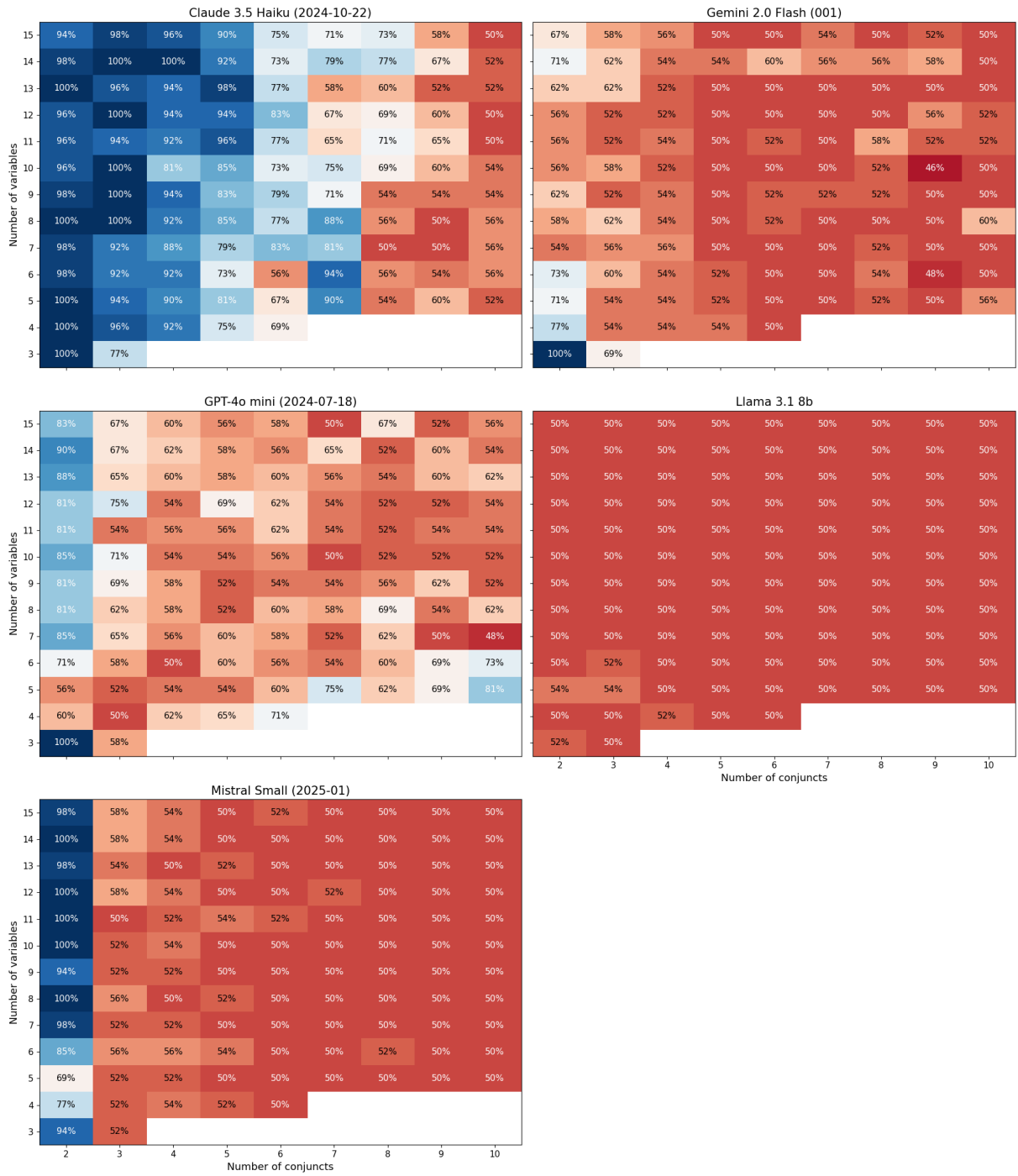

Figure 4.4: Percentage of correct answers in the set membership test set by number of variables and conjuncts achieved by state-of-the-art smaller models

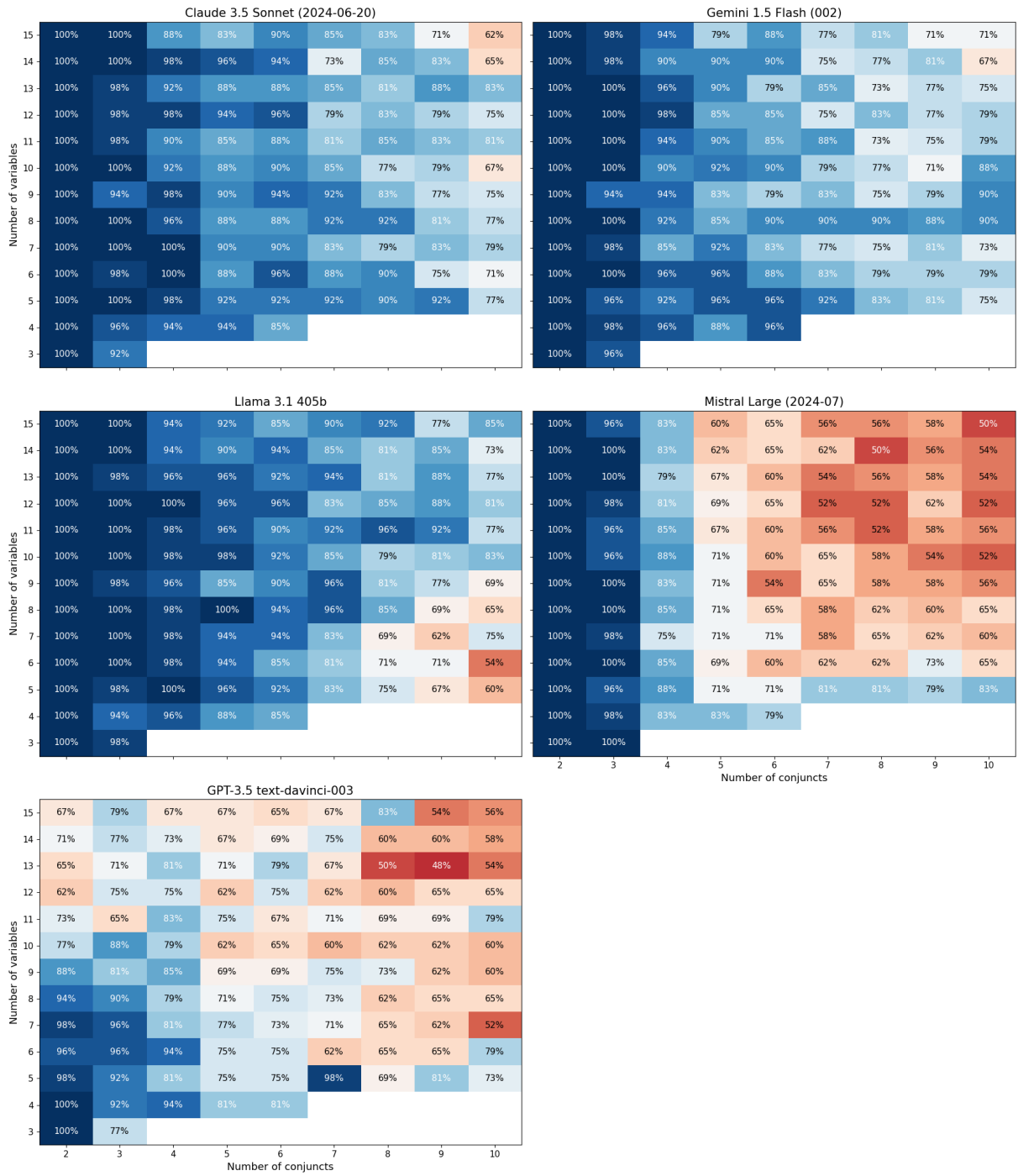

Figure 4.5: Percentage of correct answers in the set membership test set by number of variables and conjuncts achieved by select older models

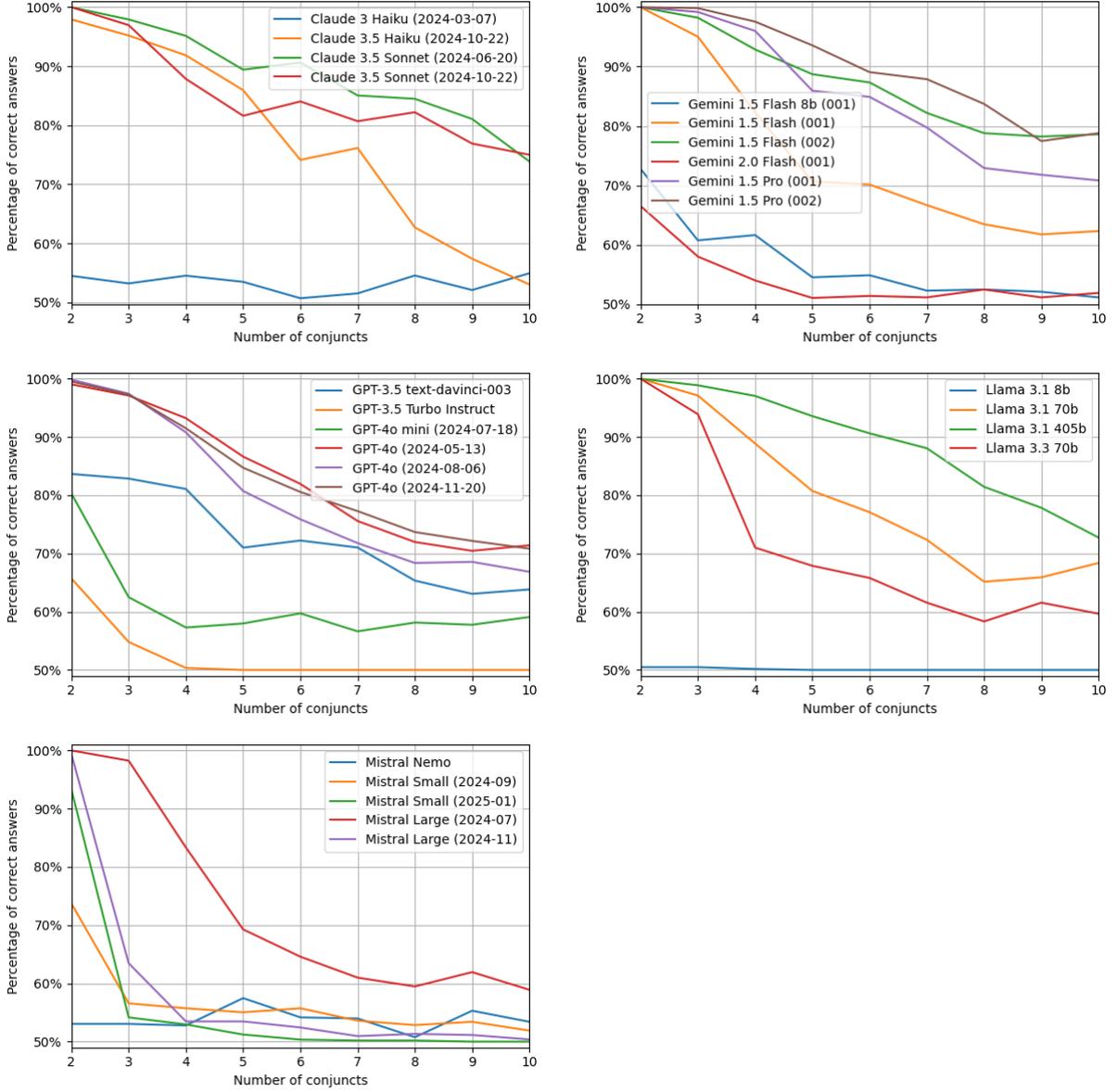

Figure 4.6: Comparison of the performance of models from the same family on the set membership test set by number of conjuncts

## 4.2 Evaluating LLM performance by formula complexity for statements involving set membership relations and negations

For this series of evaluations, we used the membership grid test set with negation (see Section 3.2.2) to evaluate the influence of the number of variables and the number of conjuncts on the performance of LLMs in judging the truth of first-order logic sentences of the form

$$\begin{cases} \exists X_1, X_2, \ldots, X_k \colon \neg(\forall X_{k+1}, X_{k+2}, \ldots, X_n \colon (X_{i_1} \notin X_{j_1} \\ \qquad\qquad\qquad \land X_{i_2} \notin X_{j_2} \land \ldots \land X_{i_m} \notin X_{j_m})) & : k > 0 \\ \neg(\forall X_1, X_2, \ldots, X_n \colon (X_{i_1} \notin X_{j_1} \land X_{i_2} \notin X_{j_2} \land \ldots \land X_{i_m} \notin X_{j_m})) & : k = 0 \end{cases} \quad (4.2)$$

for $n, m \in \mathbb{N}$, $k \in \{0, 1, ..., n-1\}$, $i_1, i_2, ..., i_m, j_1, j_2, ..., j_m \in \{1, 2, ..., n\}$. As in Section 4.1, for prompting, we used the Unicode encoding for the first-order logic sentences (cf. Section 3.3.1), we used simple yes/no instructions (cf. Section 3.3.2.1) and a 0-shot prompt (cf. Section 3.3.2.2). One example of a prompt as used in this series of evaluations is the following:

> System: `You are a helpful mathematical assistant.`
>
> User: `Is the following statement true in Zermelo-Fraenkel set theory?`
> `    ∃ A: (¬ (∀ B, C: ((B ∉ A) ∨ (A ∉ C))))`
> `    Answer only 'yes' or 'no'!`

Table 4.2 shows the results of our evaluations on the set membership grid test set with negation for all models we tested, listing the percentage of correct answers for each model by the number of conjuncts appearing in the evaluated statements. Figure 4.7 plots these percentages for all current models against the number of conjuncts and Figure 4.8 shows the same data but split up into state-of-the-art frontier models and smaller models, respectively, for easier comparability. Finally, Figures 4.9, 4.10, and 4.11 show the results for all current models, by number of variables and number of conjuncts appearing in the evaluated statements. Each cell in the grids in these figures shows the percentage of correct answers for the 48 statements with the corresponding number of variables and conjuncts.

|  | \multicolumn{9}{c}{Number of conjuncts} |
| --- | --- | --- | --- | --- | --- | --- | --- | --- | --- |
| Model | 2 | 3 | 4 | 5 | 6 | 7 | 8 | 9 | 10 |
| Claude 3 Haiku (2024-03-07) | 53.2 | 50.3 | 50.2 | 50.0 | 50.0 | 50.4 | 50.0 | 50.0 | 50.2 |
| Claude 3.5 Haiku (2024-10-22) | 50.2 | 50.0 | 50.0 | 50.0 | 50.0 | 50.0 | 50.0 | 50.0 | 50.0 |
| Claude 3.5 Sonnet (2024-06-20) | 50.6 | 52.1 | 51.2 | 49.8 | 51.9 | 49.2 | 51.5 | 49.2 | 50.9 |
| Claude 3.5 Sonnet (2024-10-22) | 60.3 | 58.5 | 55.0 | 53.8 | **58.2** | **56.2** | **57.0** | 52.1 | **55.5** |
| DeepSeek-V3 | 62.2 | 54.8 | **56.1** | 54.7 | 52.3 | 51.5 | 54.4 | 48.1 | 49.4 |
| Gemini 1.5 Flash 8b (001) | 50.0 | 50.0 | 50.0 | 50.0 | 50.0 | 50.0 | 50.0 | 50.0 | 50.0 |
| Gemini 1.5 Flash (001) | 50.6 | 49.8 | 50.5 | 50.0 | 52.6 | 49.6 | 51.5 | 49.4 | 51.9 |
| Gemini 1.5 Flash (002) | 54.5 | 60.1 | 55.7 | 56.4 | 53.8 | 54.2 | 51.7 | 51.5 | 53.6 |
| Gemini 2.0 Flash (001) | 52.6 | 50.6 | 50.0 | 50.7 | 51.2 | 48.3 | 50.9 | 49.1 | 52.1 |
| Gemini 1.5 Pro (001) | 50.6 | 51.6 | 49.0 | 50.7 | 52.3 | 48.9 | 51.7 | 50.0 | 51.9 |
| Gemini 1.5 Pro (002) | 50.3 | 50.3 | 50.0 | 50.3 | 49.7 | 50.0 | 50.0 | 50.0 | 50.0 |
| GPT-3.5 text-davinci-003 | 50.0 | 51.3 | 51.0 | 50.7 | 50.7 | 51.3 | 49.1 | 50.6 | 50.2 |
| GPT-3.5 Turbo Instruct | 50.0 | 50.0 | 50.0 | 50.0 | 50.0 | 50.0 | 50.0 | 49.8 | 50.0 |
| GPT-4o mini (2024-07-18) | 51.1 | 51.3 | 49.8 | 50.0 | 52.1 | 50.4 | 50.8 | 49.4 | 50.9 |
| GPT-4o (2024-05-13) | 44.2 | 51.1 | 52.8 | **58.2** | 51.0 | 55.9 | 55.1 | **55.3** | 54.9 |
| GPT-4o (2024-08-06) | 45.8 | 50.0 | 50.0 | 51.9 | 51.6 | 52.5 | 51.5 | 53.6 | 50.9 |
| GPT-4o (2024-11-20) | 51.4 | 58.3 | 53.3 | **58.2** | 54.3 | 55.3 | 54.7 | 51.3 | 53.2 |
| Llama 3.1 8b | 50.0 | 50.0 | 50.0 | 50.0 | 50.0 | 50.0 | 50.0 | 50.0 | 50.0 |
| Llama 3.1 70b | **62.5** | **62.2** | 52.3 | 53.6 | 53.5 | 53.6 | 50.4 | 50.0 | 49.1 |
| Llama 3.1 405b | 51.3 | 58.2 | 54.9 | 53.8 | 53.1 | 51.5 | 49.4 | 52.3 | 50.9 |
| Llama 3.3 70b | 50.6 | 49.5 | 50.2 | 49.1 | 50.9 | 50.0 | 50.8 | 50.2 | 50.4 |
| Mistral Nemo | 49.8 | 50.0 | 50.5 | 49.7 | 50.9 | 49.2 | 50.8 | 49.2 | 49.1 |
| Mistral Small (2024-09) | 50.3 | 49.8 | 50.0 | 50.0 | 50.0 | 50.0 | 50.0 | 50.0 | 49.6 |
| Mistral Small (2025-01) | 47.0 | 50.5 | 50.9 | 48.4 | 52.4 | 49.6 | 50.0 | 51.5 | 51.3 |
| Mistral Large (2024-07) | 53.0 | 53.4 | 51.2 | 50.7 | 50.7 | 51.5 | 53.2 | 50.4 | 51.7 |
| Mistral Large (2024-11) | 50.3 | 50.5 | 49.7 | 49.1 | 49.7 | 49.6 | 50.4 | 49.4 | 50.0 |

Table 4.2: Percentage of correct answers in the set membership test set with a single negation by number of conjuncts achieved by all models

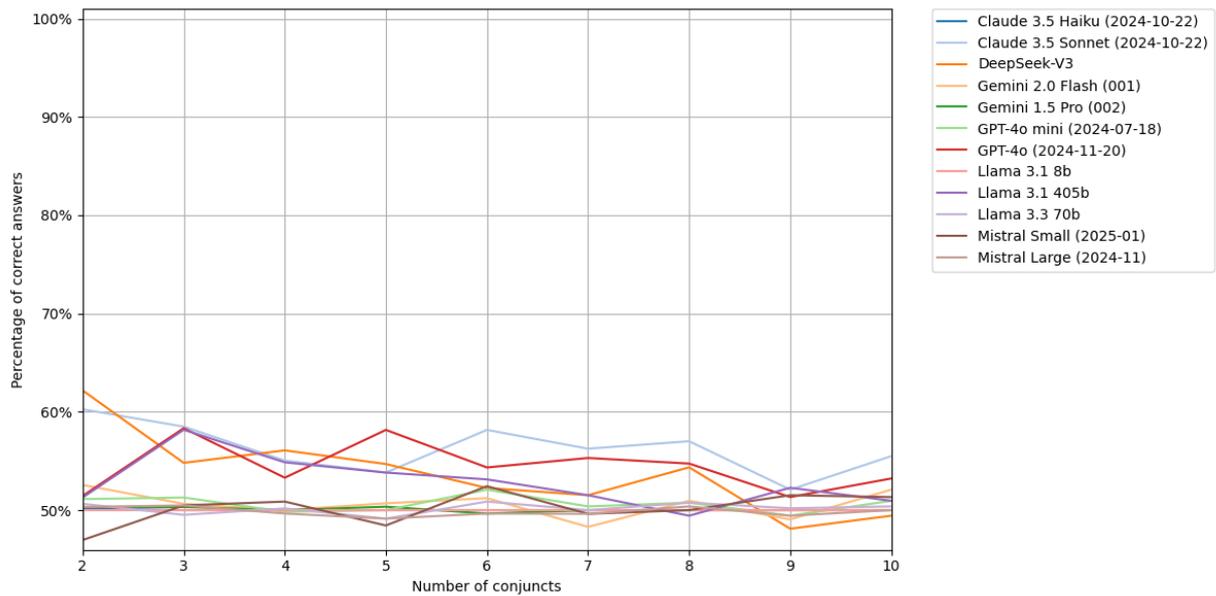

Figure 4.7: Percentage of correct answers in the set membership test set with negation by number of conjuncts achieved by current models

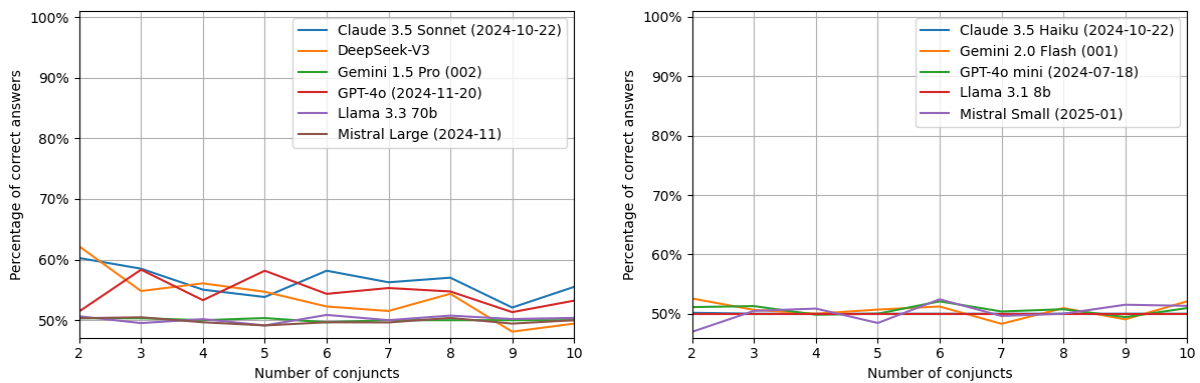

Figure 4.8: Percentage of correct answers in the set membership test set with negation by number of conjuncts achieved by state-of-the-art frontier models (left) and smaller models (right), respectively

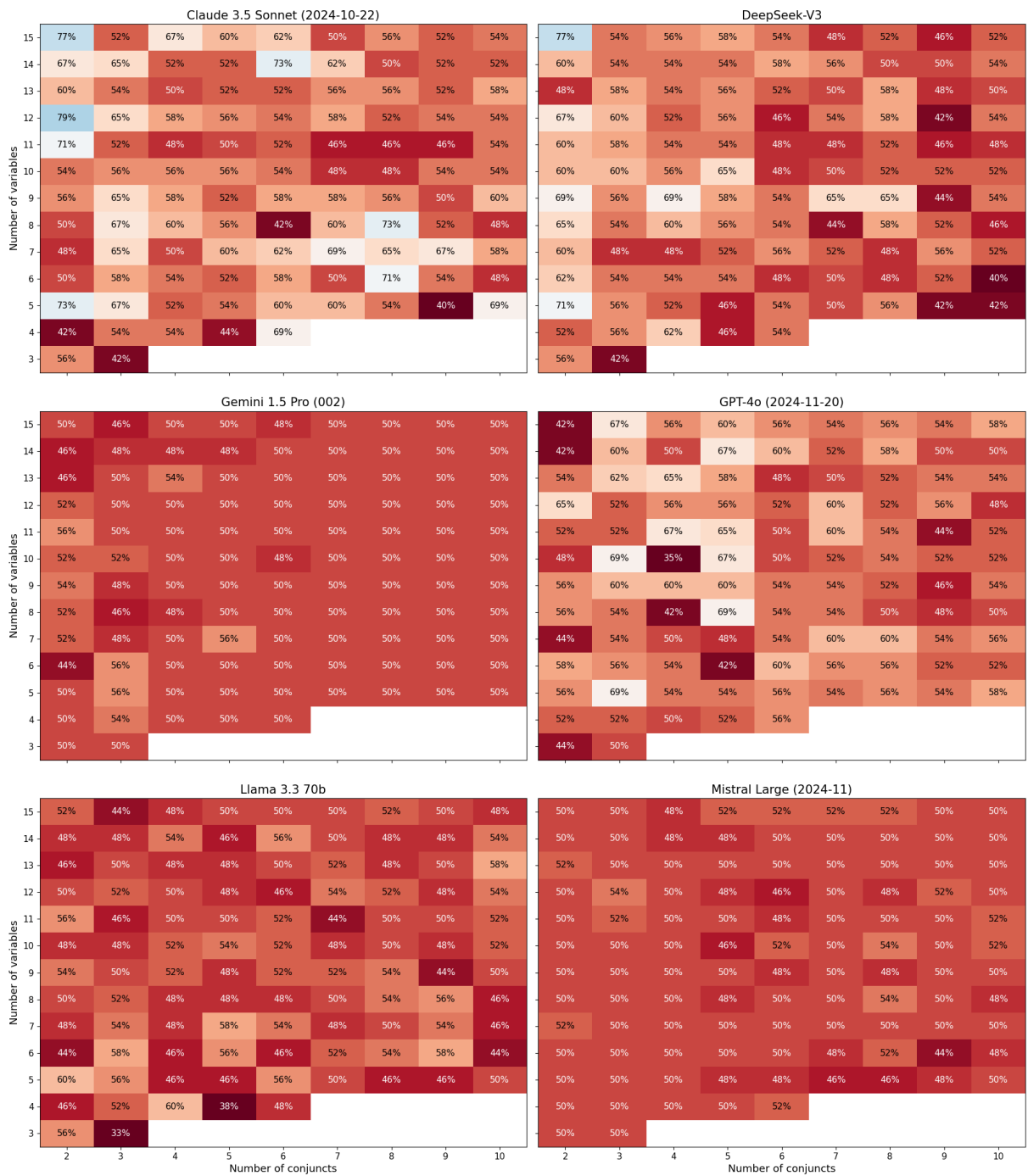

Figure 4.9: Percentage of correct answers in the set membership test set with negation by number of variables and conjuncts achieved by state-of-the-art frontier models

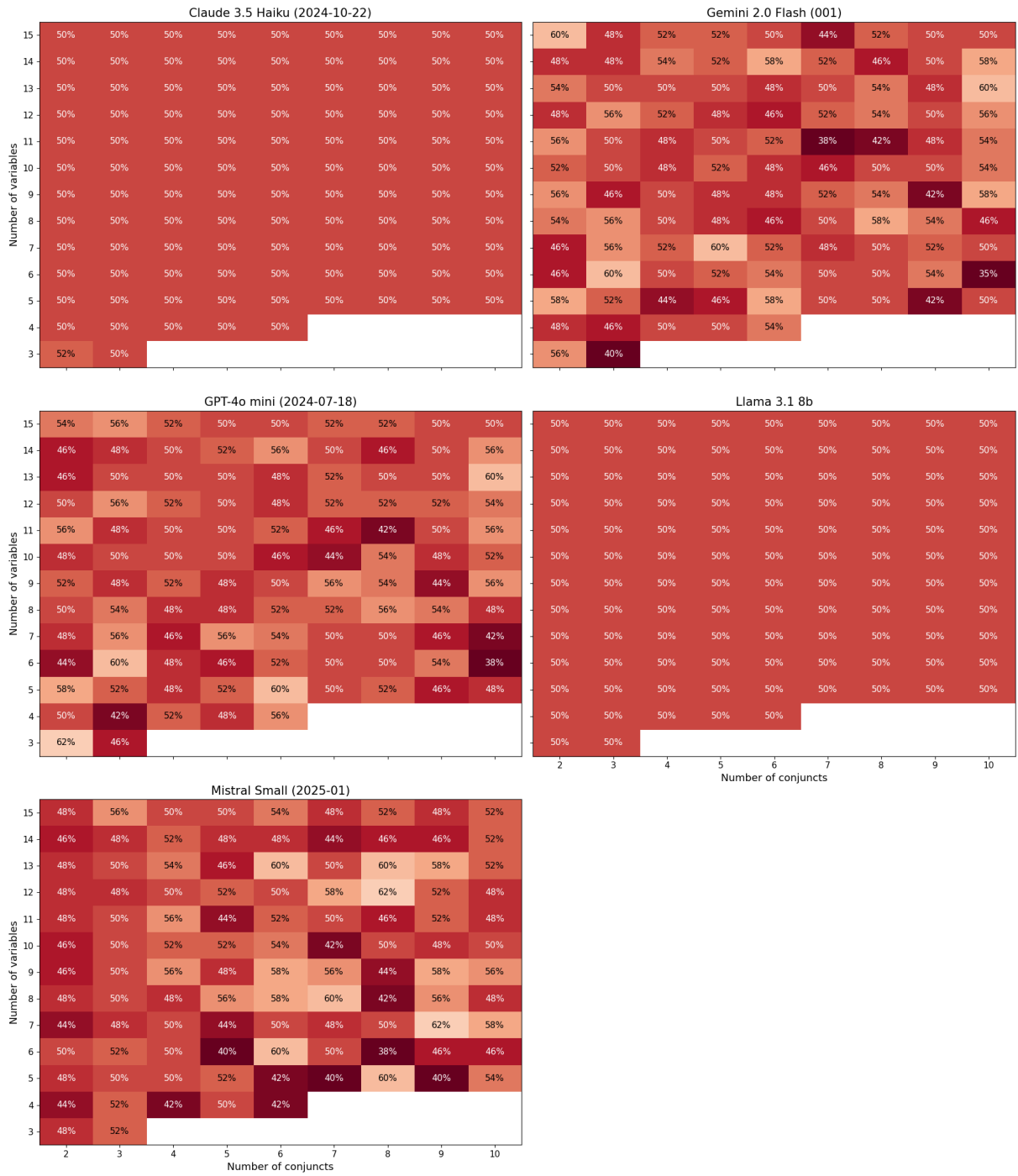

Figure 4.10: Percentage of correct answers in the set membership test set with negation by number of variables and conjuncts achieved by state-of-the-art smaller models

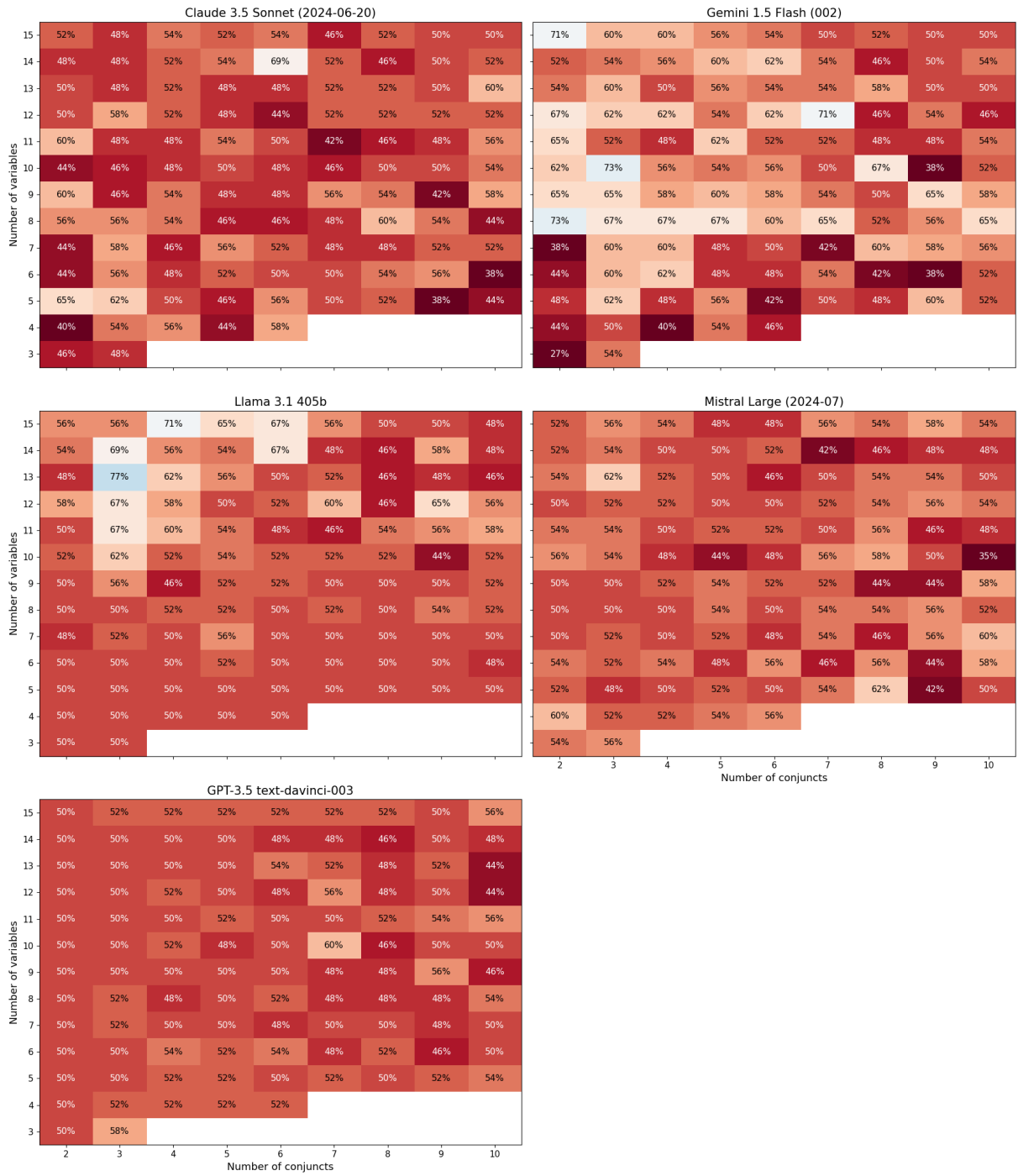

Figure 4.11: Percentage of correct answers in the set membership test set with negation by number of variables and conjuncts achieved by select older models

## 4.3 Evaluating LLM performance by formula complexity for statements involving set membership, subset, and proper subset relations

For this series of evaluations, we used the all-relations grid test set (see Section 3.2.3) to evaluate the influence of the number of variables and the number of conjuncts on the performance of LLMs in judging the truth of first-order logic sentences of the form

$$\exists X_1, X_2, \ldots, X_n : (X_{i_1} R_1 X_{j_1} \wedge X_{i_2} R_2 X_{j_2} \wedge \ldots \wedge X_{i_m} R_m X_{j_m}) \quad (4.3)$$

for some $n, m \in \mathbb{N}$, $i_1, i_2, \ldots, i_m, j_1, j_2, \ldots, j_m \in \{1, 2, \ldots, n\}$, and relations $R_1, R_2, \ldots, R_m$ chosen among set membership, subset, or proper subset. For prompting, we used simple yes/no instructions (cf. Section 3.3.2.1) and a 0-shot prompt (cf. Section 3.3.2.2). We ran all tests with two different encodings for the first-order logic formulae, the Unicode and the words encoding (cf. Section 3.3.1). One example of a prompt as used in this series of evaluations in the Unicode encoding is the following:

```
System: You are a helpful mathematical assistant.

User: Is the following statement true in Zermelo-Fraenkel set theory?
    ∃ A, B, C: ((A ⊆ C) ∧ (B ∈ C) ∧ (A ⊊ C) ∧ (B ∈ A) ∧ (C ⊆ B) ∧ (A ⊆
    B))
    Answer only 'yes' or 'no'!
```

The same prompt with the words encoding looks as follows:

```
System: You are a helpful mathematical assistant.

User: Is the following statement true in Zermelo-Fraenkel set theory?
    There exist sets A, B, and C such that A is a subset of C, B is an
    element of C, A is a proper subset of C, B is an element of A, C is
    a subset of B, and A is a subset of B
    Answer only 'yes' or 'no'!
```

Table 4.3 shows the results of our evaluations on the all-relations grid test set for all models we tested, both for the Unicode and the words encoding, listing the percentage of correct answers for each model by the number of conjuncts appearing in the evaluated statements. Figure 4.12 plots these percentages for all current models against the number of conjuncts and Figure 4.13 shows the same data but split up into state-of-the-art frontier models and smaller models, respectively, for easier comparability. The plots in Figure 4.14 plot the percentages of correct answers by number of conjuncts, grouped by model family. Finally, Figures 4.15, 4.16, and 4.17 show the results for all current models, by number of variables and number of conjuncts appearing in the evaluated statements in the Unicode encoding, and Figures 4.18, 4.19, and 4.20 show the same for the words encoding. Each cell in the grids in these figures shows the percentage of correct answers for the 48 statements with the corresponding number of variables and conjuncts.

| Model | Enc. | \multicolumn{9}{c}{Number of conjuncts} |
| | | 2 | 3 | 4 | 5 | 6 | 7 | 8 | 9 | 10 |
| --- | --- | --- | --- | --- | --- | --- | --- | --- | --- | --- |
| Claude 3 Haiku (2024-03-07) | U | 58.2 | 59.1 | 57.1 | 55.4 | 57.7 | 62.3 | 60.4 | 62.0 | 60.1 |
|  | W | 94.2 | 85.4 | 74.7 | 69.9 | 61.9 | 60.7 | 57.7 | 56.1 | 55.9 |
| Claude 3.5 Haiku (2024-10-22) | U | 97.3 | 92.5 | 86.5 | 79.2 | 69.2 | 63.3 | 56.1 | 54.8 | 52.4 |
|  | W | 92.6 | 86.1 | 80.0 | 74.5 | 68.1 | 65.1 | 61.2 | 57.2 | 60.1 |
| Claude 3.5 Sonnet (2024-06-20) | U | 99.5 | 98.2 | 94.2 | 92.6 | 91.2 | **91.3** | 85.3 | 82.2 | 82.8 |
|  | W | **100.0** | 96.6 | 92.1 | 90.2 | 90.4 | 88.5 | 86.4 | 84.3 | 83.5 |
| Claude 3.5 Sonnet (2024-10-22) | U | 98.6 | 97.9 | 93.3 | 90.1 | 91.0 | 88.1 | 88.9 | **86.4** | 84.9 |
|  | W | 98.9 | 90.9 | 85.3 | 85.4 | 84.0 | 87.5 | 84.0 | **85.3** | 84.9 |
| DeepSeek-V3 | U | **100.0** | 98.4 | 90.1 | 86.1 | 79.3 | 76.1 | 73.2 | 72.8 | 71.9 |
|  | W | 99.7 | 98.9 | 92.8 | 86.9 | 83.2 | 79.5 | 77.6 | 78.2 | 76.9 |
| Gemini 1.5 Flash 8b (001) | U | 92.1 | 78.7 | 73.9 | 68.3 | 63.9 | 64.9 | 65.2 | 61.7 | 62.3 |
|  | W | 98.6 | 88.1 | 80.4 | 74.0 | 70.2 | 71.6 | 67.1 | 66.8 | 69.1 |
| Gemini 1.5 Flash (001) | U | 97.1 | 96.3 | 86.4 | 79.8 | 78.5 | 75.8 | 75.6 | 72.4 | 70.8 |
|  | W | 99.5 | 96.0 | 92.3 | 88.8 | 83.3 | 81.1 | 79.2 | 74.7 | 74.5 |
| Gemini 1.5 Flash (002) | U | 96.8 | 96.8 | 91.3 | 88.5 | 84.5 | 81.7 | 81.7 | 79.2 | 79.2 |
|  | W | **100.0** | 98.1 | 95.8 | 92.6 | 89.1 | 88.5 | 87.0 | 83.2 | **85.1** |
| Gemini 2.0 Flash (001) | U | 71.6 | 65.4 | 59.1 | 58.3 | 57.2 | 57.7 | 57.7 | 59.6 | 56.4 |
|  | W | 81.9 | 73.1 | 67.1 | 63.9 | 62.5 | 64.3 | 62.7 | 60.9 | 63.9 |
| Gemini 1.5 Pro (001) | U | 98.6 | 99.0 | 94.9 | 92.5 | 86.9 | 83.2 | 80.3 | 76.0 | 73.3 |
|  | W | 99.5 | 98.6 | 93.9 | 92.5 | 88.6 | 86.7 | 84.3 | 77.9 | 79.2 |
| Gemini 1.5 Pro (002) | U | 98.1 | 98.1 | **96.5** | **95.0** | **92.9** | 90.4 | **90.1** | 85.7 | **85.6** |
|  | W | **100.0** | **99.4** | **96.2** | 92.9 | **92.1** | 89.9 | 87.3 | 83.7 | 82.3 |
| GPT-3.5 text-davinci-003 | U | 87.0 | 82.2 | 71.2 | 71.2 | 62.5 | 67.8 | 65.4 | 61.9 | 60.2 |
|  | W | 92.5 | 88.8 | 77.4 | 73.2 | 64.3 | 60.9 | 60.6 | 59.1 | 61.5 |
| GPT-3.5 Turbo Instruct | U | 67.9 | 51.8 | 50.0 | 49.8 | 50.3 | 50.5 | 50.5 | 50.8 | 50.2 |
|  | W | 92.6 | 70.4 | 55.0 | 52.9 | 50.8 | 51.4 | 51.4 | 51.0 | 50.3 |
| GPT-4o mini (2024-07-18) | U | 86.1 | 68.3 | 63.8 | 64.4 | 63.9 | 65.9 | 66.2 | 63.0 | 58.9 |
|  | W | 91.7 | 76.9 | 69.1 | 67.1 | 62.5 | 63.5 | 64.3 | 64.1 | 64.8 |
| GPT-4o (2024-05-13) | U | 98.6 | 96.6 | 91.8 | 85.9 | 83.8 | 80.1 | 80.0 | 74.5 | 74.0 |
|  | W | 99.0 | 97.0 | 94.6 | 91.7 | 90.7 | 84.6 | 83.3 | 78.4 | 78.6 |
| GPT-4o (2024-08-06) | U | 97.6 | 97.0 | 90.9 | 83.2 | 83.3 | 78.5 | 82.2 | 77.1 | 74.5 |
|  | W | 97.0 | 88.9 | 87.5 | 84.8 | 87.2 | 83.7 | 82.2 | 80.3 | 79.5 |
| GPT-4o (2024-11-20) | U | 98.4 | 97.3 | 90.5 | 84.1 | 85.6 | 79.3 | 81.6 | 77.2 | 73.1 |
|  | W | 97.6 | 94.7 | 90.2 | 86.2 | 84.3 | 80.4 | 79.5 | 78.0 | 75.7 |
| Llama 3.1 8b | U | 52.7 | 52.7 | 54.0 | 52.2 | 51.4 | 50.8 | 50.5 | 50.3 | 50.0 |
|  | W | 72.9 | 70.4 | 70.8 | 67.9 | 62.3 | 57.5 | 56.1 | 53.2 | 53.0 |
| Llama 3.1 70b | U | 99.2 | 97.4 | 87.8 | 78.8 | 77.2 | 73.6 | 71.3 | 69.7 | 70.3 |
|  | W | 98.4 | 96.5 | 91.7 | 86.9 | 83.0 | 81.6 | 76.0 | 74.8 | 76.9 |
| Llama 3.1 405b | U | 98.9 | **99.4** | 93.4 | 88.6 | 86.4 | 85.9 | 84.6 | 82.1 | 81.4 |
|  | W | 98.9 | 98.1 | 92.6 | 91.2 | 89.4 | 88.6 | 85.4 | 81.7 | 80.4 |
| Llama 3.3 70b | U | 99.5 | 94.7 | 83.7 | 78.5 | 80.0 | 78.2 | 77.4 | 74.4 | 74.7 |
|  | W | 99.0 | 95.5 | 90.4 | 86.7 | 86.5 | 83.0 | 80.3 | 76.4 | 78.5 |
| Mistral Nemo | U | 57.1 | 59.5 | 58.8 | 57.4 | 56.4 | 56.2 | 57.7 | 56.7 | 55.9 |
|  | W | 89.1 | 83.7 | 79.8 | 76.9 | 71.2 | 67.8 | 67.8 | 63.6 | 62.7 |
| Mistral Small (2024-09) | U | 82.4 | 72.9 | 60.7 | 60.1 | 57.1 | 56.7 | 57.7 | 56.1 | 53.0 |
|  | W | 96.5 | 86.7 | 74.4 | 71.5 | 74.4 | 72.3 | 76.6 | 71.8 | 73.3 |
| Mistral Small (2025-01) | U | 91.2 | 78.4 | 70.5 | 63.9 | 61.2 | 58.3 | 57.5 | 56.7 | 55.0 |
|  | W | 72.9 | 66.5 | 63.3 | 57.7 | 58.8 | 59.3 | 58.3 | 58.8 | 56.2 |
| Mistral Large (2024-07) | U | 99.2 | 98.9 | 91.8 | 86.7 | 80.8 | 78.8 | 77.4 | 74.4 | 76.2 |
|  | W | 98.2 | 97.6 | 94.2 | 89.4 | 85.4 | 84.6 | 82.1 | 77.2 | 80.0 |
| Mistral Large (2024-11) | U | 93.9 | 76.3 | 66.7 | 62.2 | 62.7 | 62.7 | 64.7 | 64.7 | 61.1 |
|  | W | 98.4 | 96.3 | 91.0 | 82.9 | 80.6 | 78.4 | 74.4 | 73.7 | 70.5 |

Table 4.3: Percentage of correct answers in the all-relations grid test set by number of conjuncts and encoding (U = Unicode encoding, W = words encoding) achieved by all models

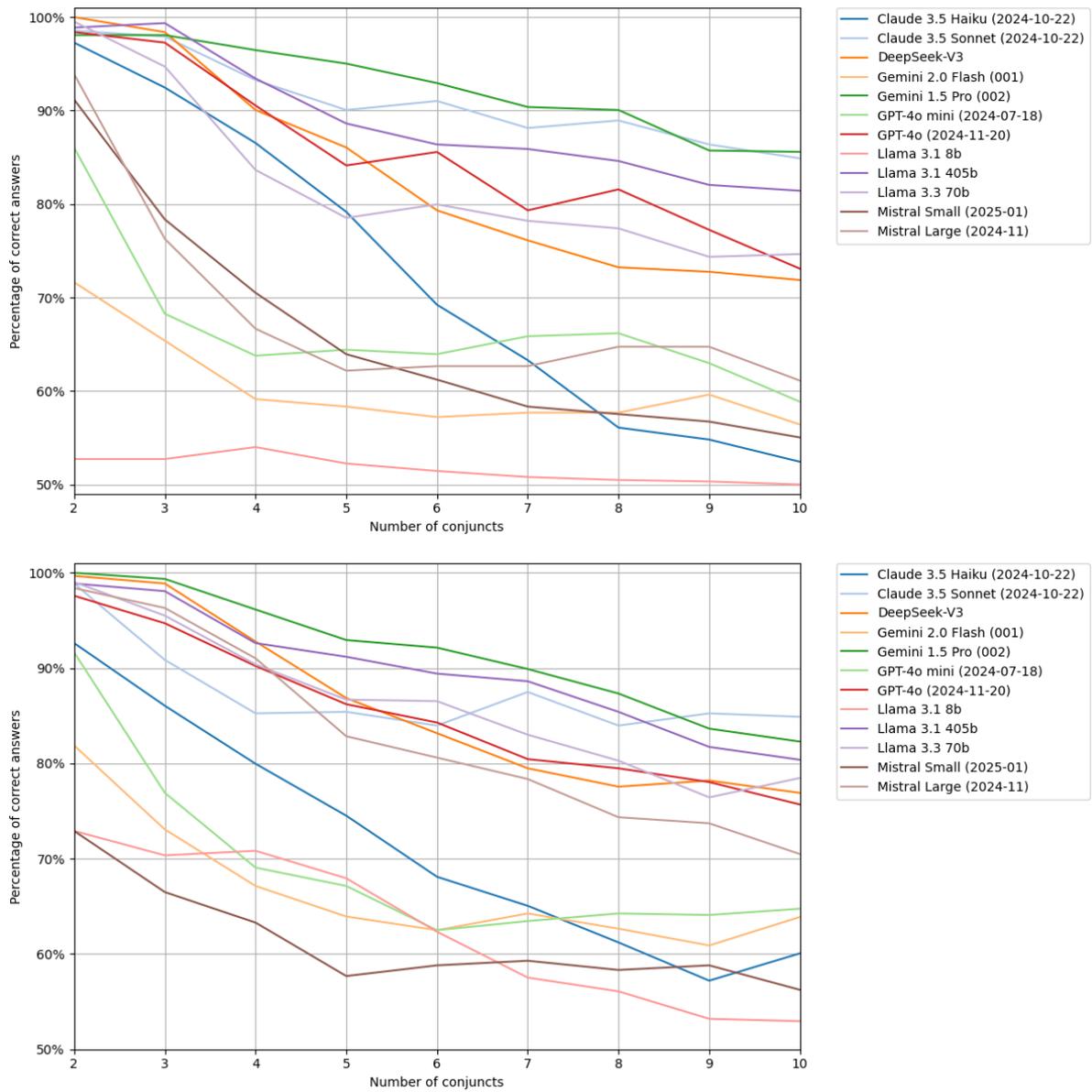

Figure 4.12: Percentage of correct answers in the all-relations grid test by number of conjuncts achieved by current models, in the Unicode encoding (top) and in the words encoding (bottom), respectively

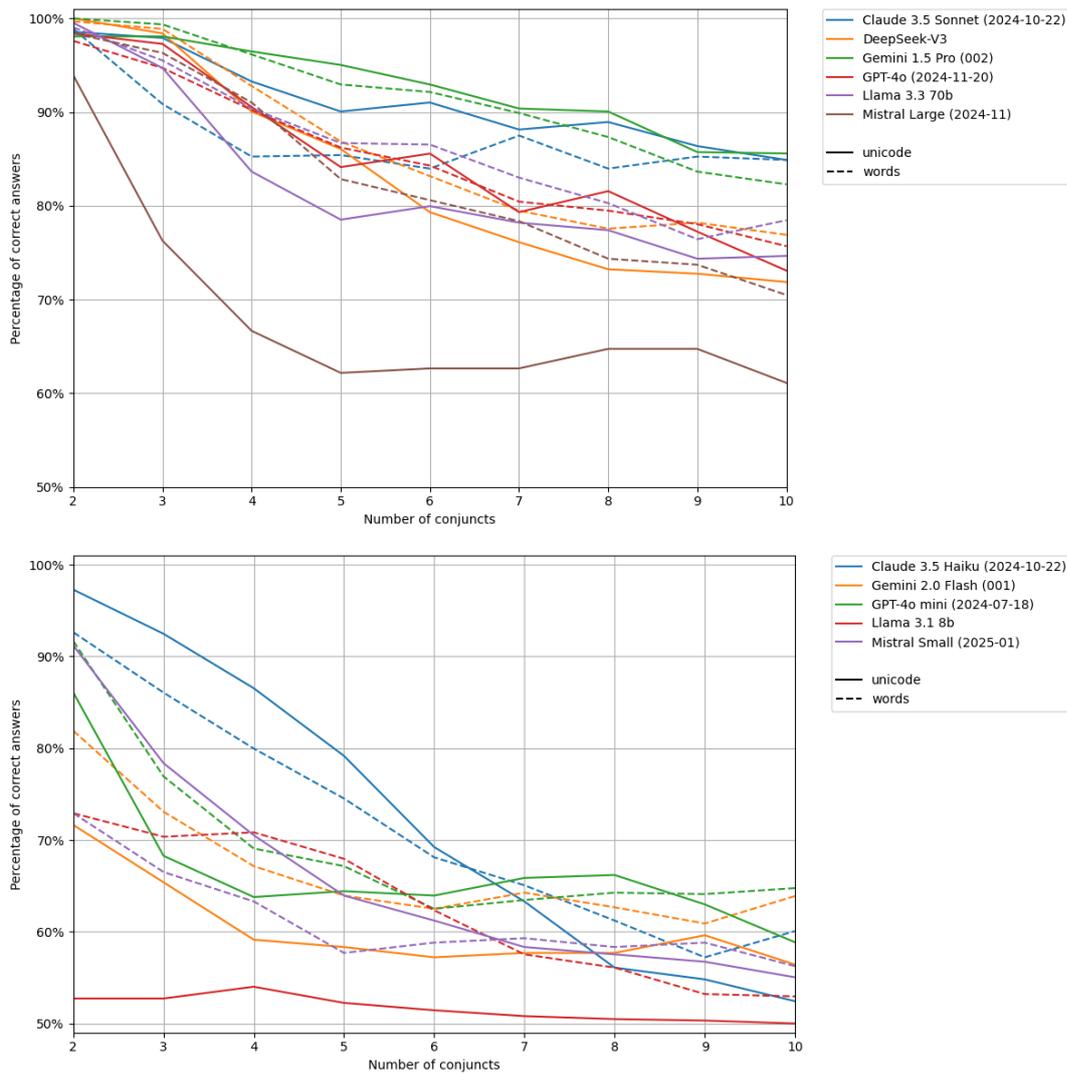

Figure 4.13: Percentage of correct answers in the all-relations grid test by number of conjuncts achieved by state-of-the-art frontier models (top) and state-of-the-art smaller models (bottom), respectively, comparing the use of the Unicode and words encoding in the prompt

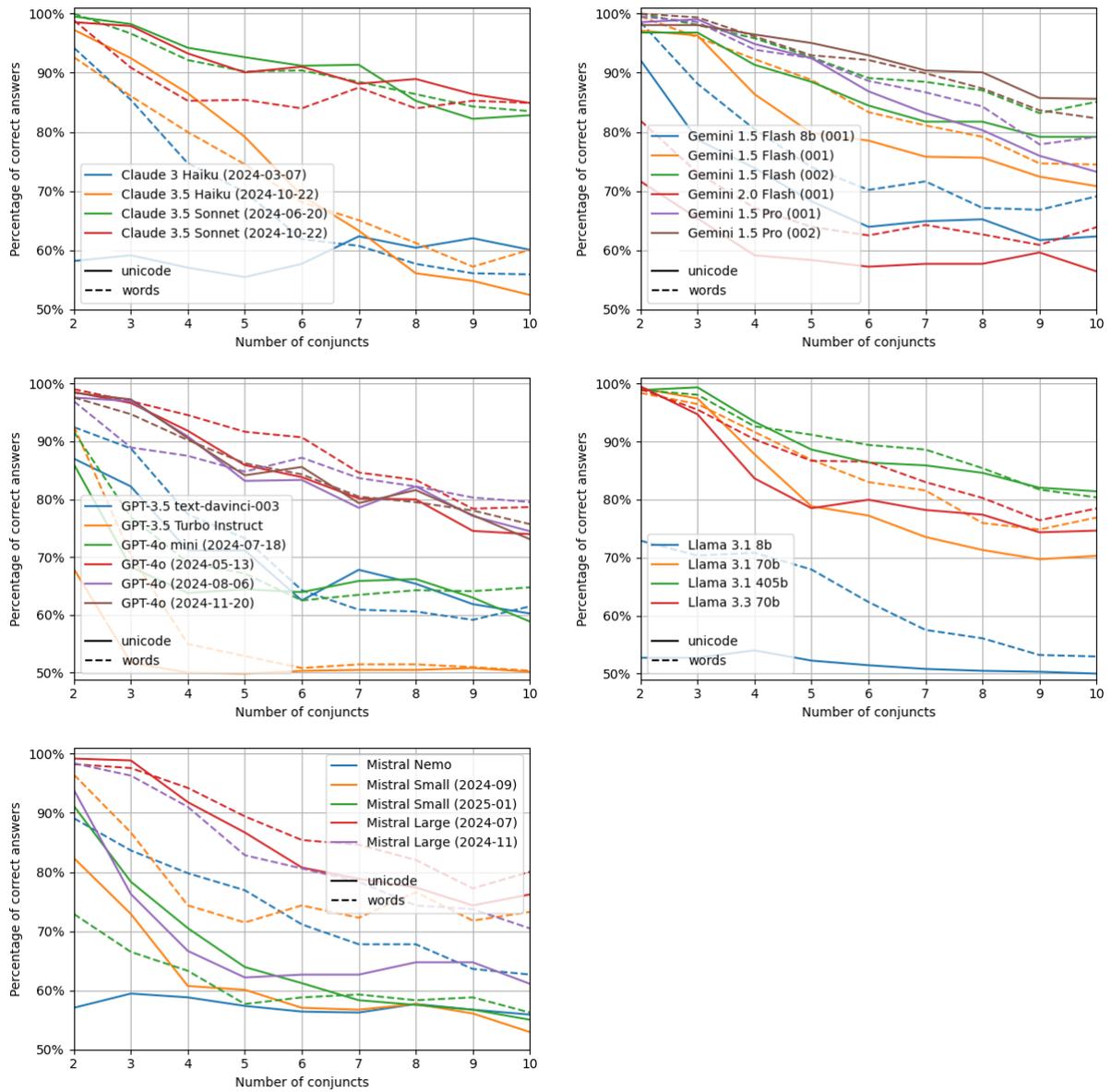

Figure 4.14: Comparison of the performance of models from the same family on the all-relations grid test by number of conjuncts, comparing the use of the Unicode and words encoding in the prompt

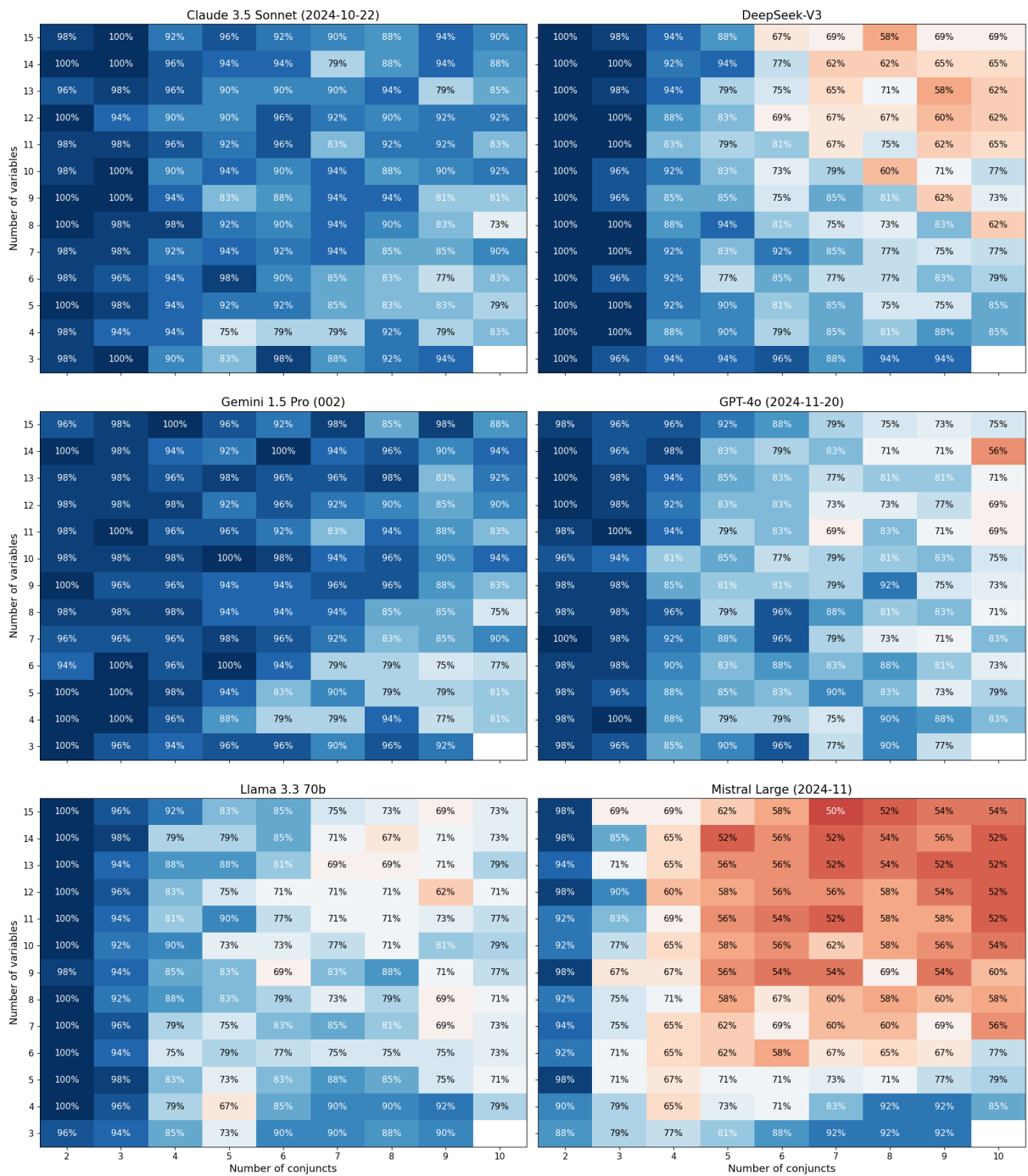

Figure 4.15: Percentage of correct answers in the all-relations grid test in the Unicode encoding by number of variables and conjuncts achieved by state-of-the-art frontier models

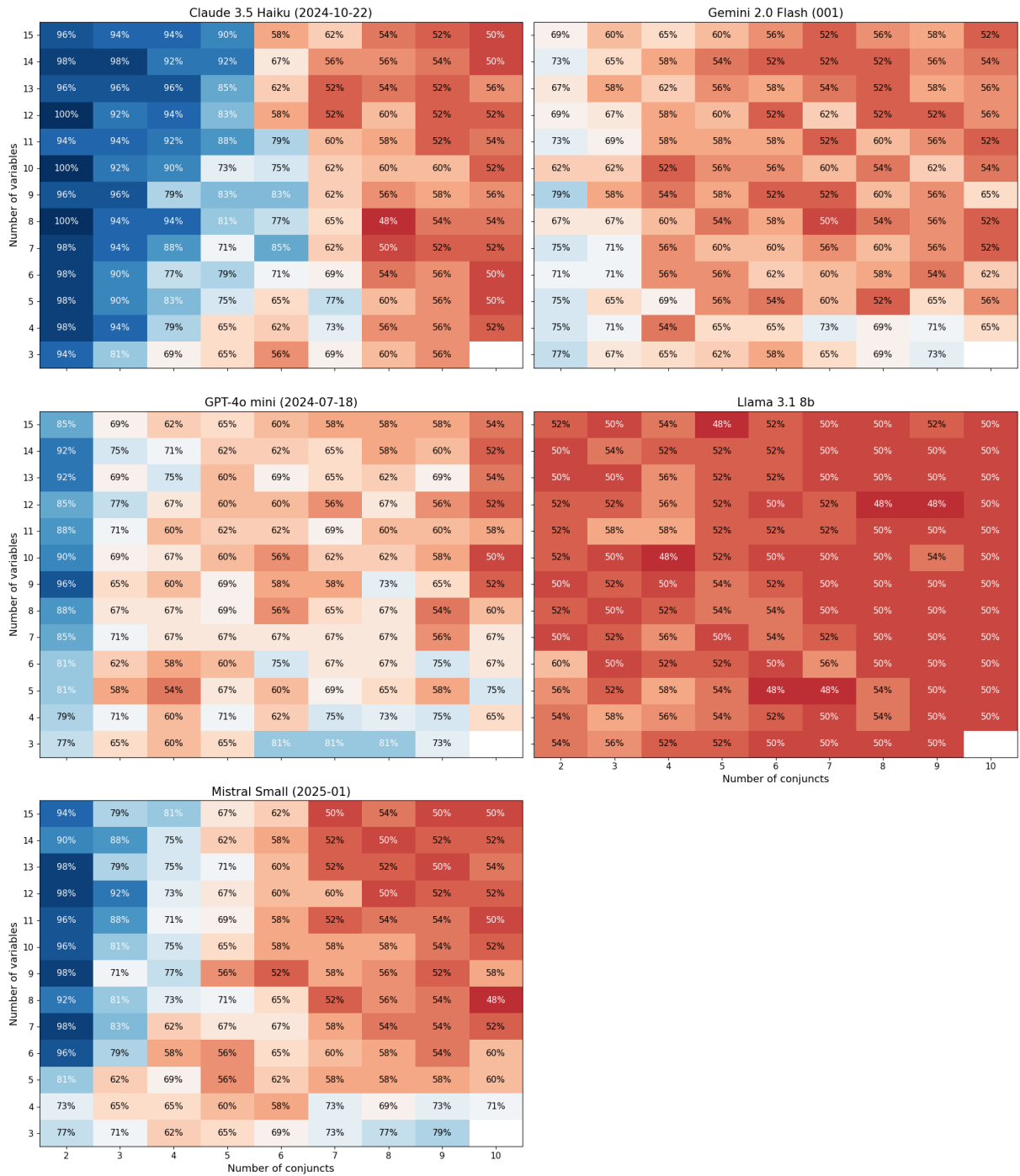

Figure 4.16: Percentage of correct answers in the all-relations grid test in the Unicode encoding by number of variables and conjuncts achieved by state-of-the-art smaller models

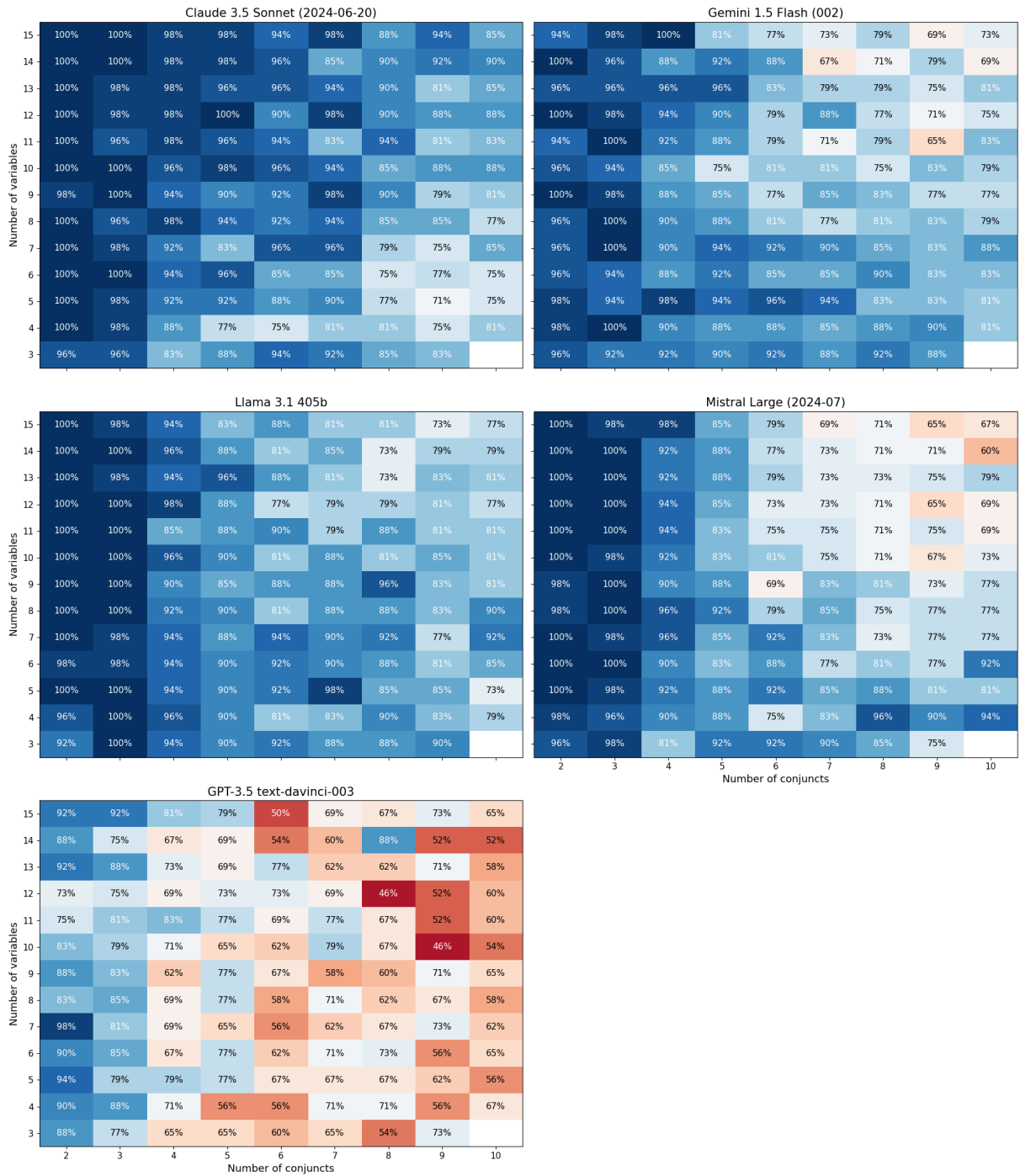

Figure 4.17: Percentage of correct answers in the all-relations grid test in the Unicode encoding by number of variables and conjuncts achieved by select older models

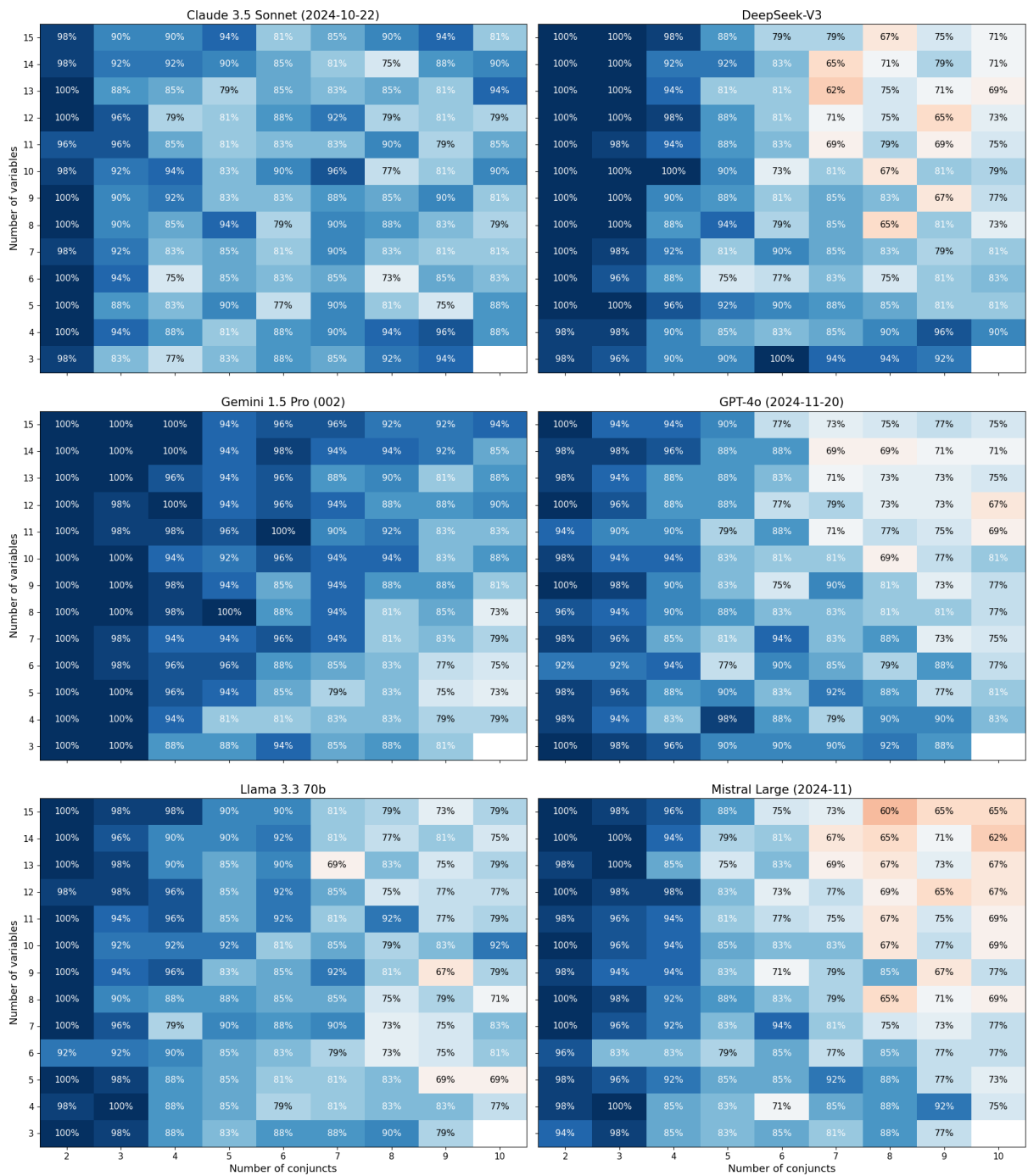

Figure 4.18: Percentage of correct answers in the all-relations grid test in the words encoding by number of variables and conjuncts achieved by state-of-the-art frontier models

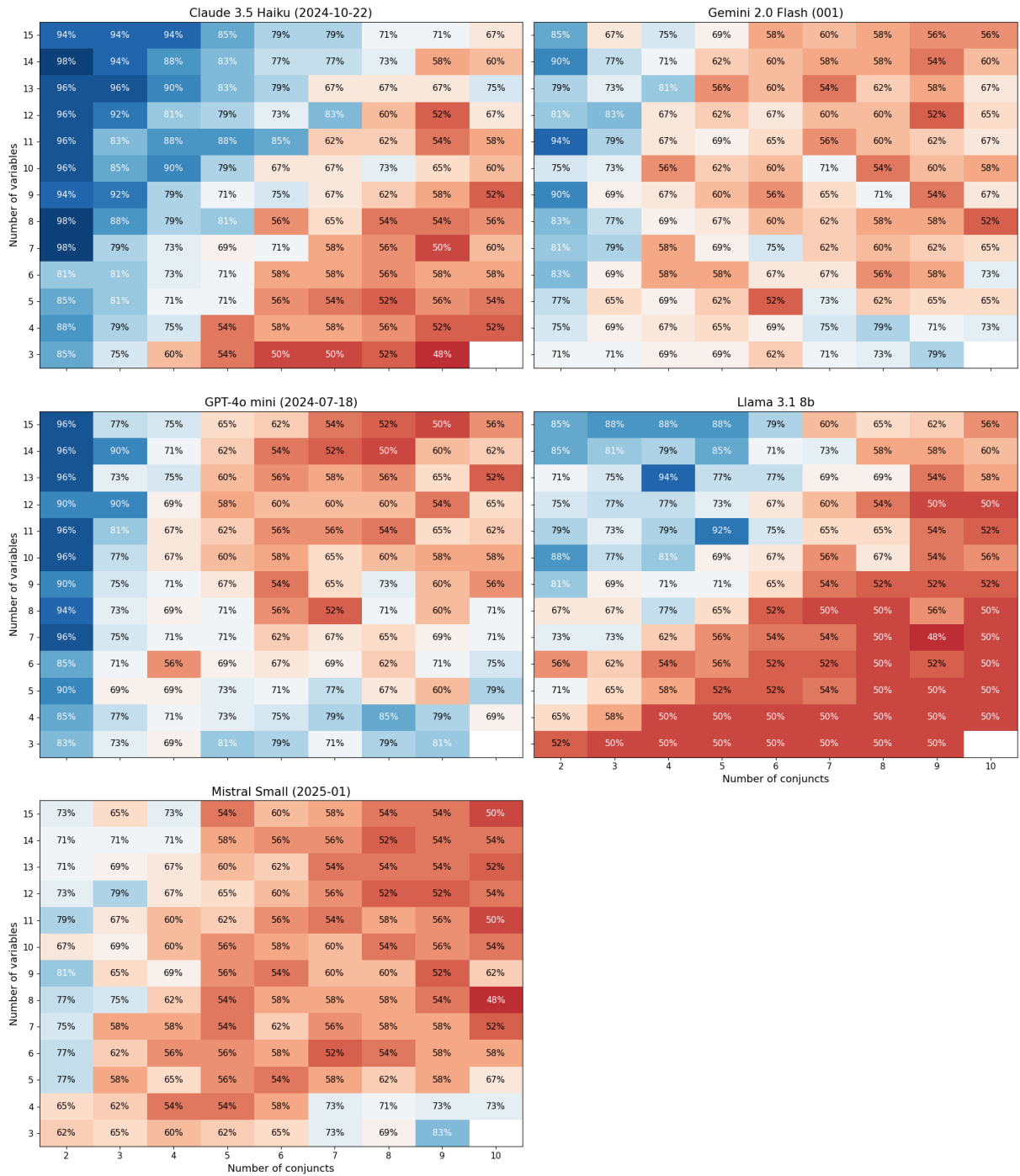

Figure 4.19: Percentage of correct answers in the all-relations grid test in the words encoding by number of variables and conjuncts achieved by state-of-the-art smaller models

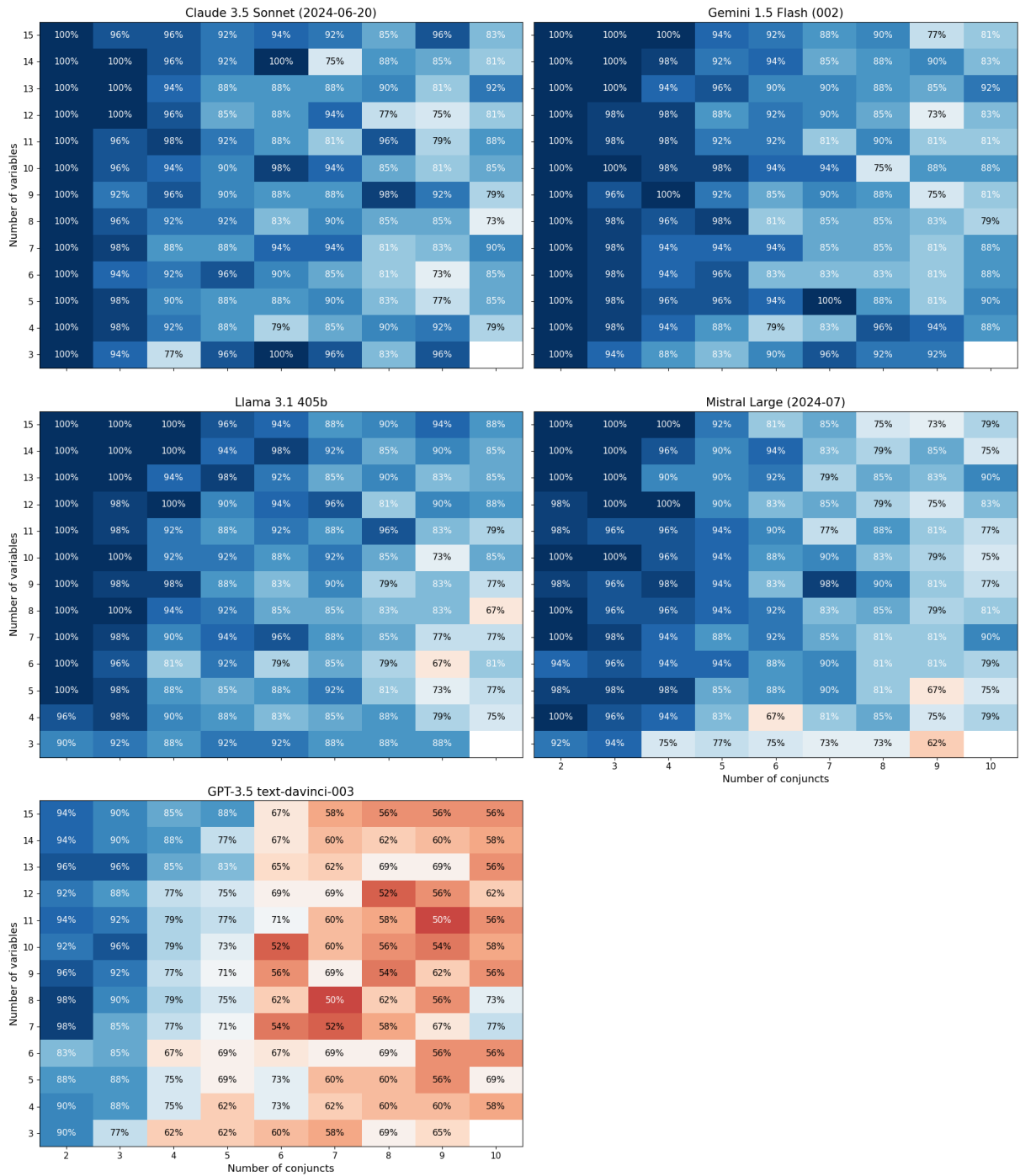

Figure 4.20: Percentage of correct answers in the all-relations grid test in the words encoding by number of variables and conjuncts achieved by select older models

## 4.4 Evaluating the influence of different prompting strategies on LLM performance

### 4.4.1 Comparing different encodings

For this series of evaluations, we used the mixed test set (cf. Section 3.2.4) to evaluate how different ways of encoding logical statements affect the performance of the LLMs. In all cases, we used a 0-shot yes/no prompt (cf. Section 3.3.2.1), so the prompt was of the following form:

```
System: You are a helpful mathematical assistant.

User: Is the following statement true in Zermelo-Fraenkel set theory?
      [statement in the chosen encoding]
      Answer only 'yes' or 'no'!
```

For the encoding of the logical statements, we used the four different encodings described in Section 3.3.1 above, i.e., Unicode, Unicode (fewer parentheses), LaTeX, and words. The results of these tests are shown in Table 4.4 and Figure 4.21. Table 4.4 shows the percentage of correct answers for each model and for each encoding as well as the percentage of prompts where the model answered 'yes'. Note that the construction of the mixed test set (cf. Section 3.2.4) ensures that 50% of the statements are true and 50% are false. In Figure 4.21, the opacity in each bar indicates whether an answer was correct or incorrect, while the hatching indicates whether the model answered 'yes' or 'no'.

|  | Condition | | | |
|---|---|---|---|---|
| Model | Unicode | Unicode (fewer parentheses) | LaTeX | Words |
| Claude 3 Haiku (2024-03-07) | 62.2 (18.8) | 58.2 (20.6) | 64.3 (19.0) | 73.5 (41.0) |
| Claude 3.5 Haiku (2024-10-22) | 71.3 (24.5) | 71.2 (23.8) | 74.2 (34.8) | 71.2 (24.3) |
| Claude 3.5 Sonnet (2024-06-20) | 77.9 (61.8) | 79.9 (48.3) | 78.8 (61.1) | 77.8 (51.1) |
| Claude 3.5 Sonnet (2024-10-22) | 82.0 (53.7) | **83.2** (56.0) | **84.3** (52.0) | 80.0 (59.7) |
| DeepSeek-V3 | 78.7 (54.6) | 82.5 (61.1) | 80.7 (60.6) | 81.0 (54.2) |
| Gemini 1.5 Flash 8b (001) | 66.5 (80.6) | 67.1 (81.1) | 64.2 (84.8) | 71.7 (73.8) |
| Gemini 1.5 Flash (001) | 74.6 (63.8) | 74.7 (56.7) | 75.1 (61.4) | 73.4 (43.0) |
| Gemini 1.5 Flash (002) | 78.7 (60.2) | 81.3 (60.2) | 80.6 (58.1) | 80.0 (63.1) |
| Gemini 2.0 Flash (001) | 59.7 (82.7) | 62.2 (81.5) | 62.6 (79.5) | 66.3 (79.2) |
| Gemini 1.5 Pro (001) | 79.2 (55.4) | 79.5 (60.4) | 80.7 (58.2) | 75.7 (50.0) |
| Gemini 1.5 Pro (002) | 80.4 (61.6) | 80.7 (61.4) | 80.8 (62.3) | 79.4 (60.8) |
| GPT-3.5 Turbo Instruct | 51.7 (97.2) | 50.5 (99.5) | 53.7 (95.5) | 63.0 (79.6) |
| GPT-4o mini (2024-07-18) | 68.3 (73.5) | 67.9 (73.4) | 68.5 (73.8) | 70.1 (75.4) |
| GPT-4o (2024-05-13) | 78.3 (48.7) | 77.2 (53.7) | 79.9 (53.4) | 79.6 (53.2) |
| GPT-4o (2024-08-06) | 79.2 (62.6) | 77.6 (61.2) | 79.4 (63.0) | 77.0 (63.5) |
| GPT-4o (2024-11-20) | 80.6 (59.1) | 78.8 (60.1) | 77.9 (65.7) | 80.8 (59.9) |
| Llama 3.1 8b | 50.8 (0.8) | 50.4 (0.7) | 52.9 (2.9) | 53.4 (3.4) |
| Llama 3.1 70b | 78.0 (41.0) | 81.0 (43.1) | 79.0 (47.5) | 77.9 (36.4) |
| Llama 3.1 405b | **83.1** (46.8) | 80.2 (35.7) | 81.9 (40.6) | **81.3** (52.2) |
| Llama 3.3 70b | 72.6 (68.7) | 72.5 (67.7) | 74.6 (69.3) | 76.7 (62.7) |
| Mistral Nemo | 59.8 (25.7) | 58.5 (34.7) | 58.9 (18.4) | 72.6 (34.3) |
| Mistral Small (2024-09) | 66.8 (53.6) | 69.0 (51.6) | 70.9 (48.9) | 76.3 (50.1) |
| Mistral Small (2025-01) | 70.8 (49.9) | 74.3 (57.7) | 68.5 (52.4) | 70.1 (75.7) |
| Mistral Large (2024-07) | 80.2 (42.3) | **83.2** (52.8) | 80.3 (39.8) | **81.3** (41.9) |
| Mistral Large (2024-11) | 72.4 (76.9) | 71.3 (77.6) | 72.2 (73.0) | 79.4 (64.6) |

Table 4.4: Percentages of correct answers achieved by LLMs on the mixed test set for different kinds of encodings of the logical statements (with the percentage of 'yes' answers and the percentage of refusals to answer in parentheses)

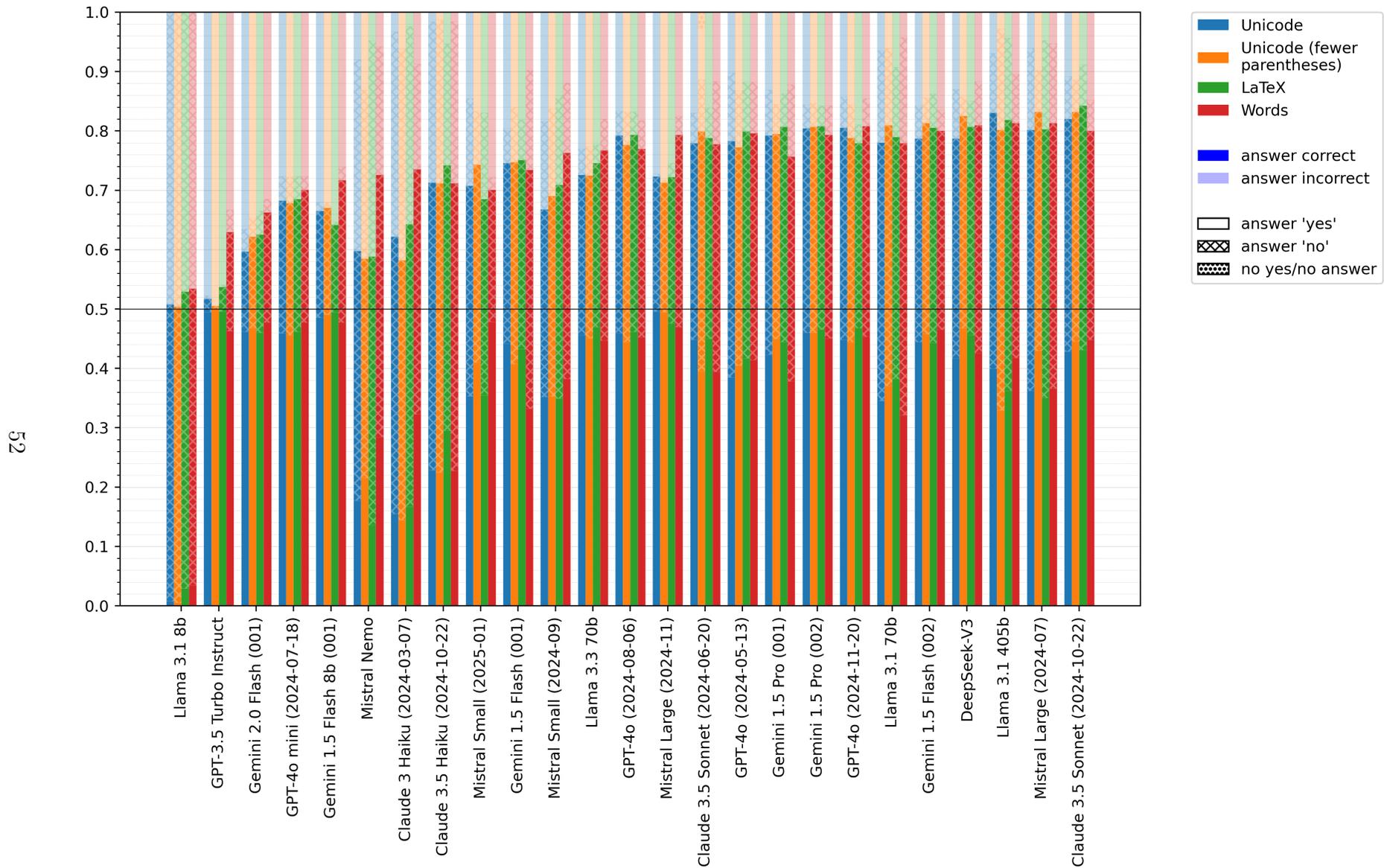

Figure 4.21: Percentages of correct answers achieved by LLMs on the mixed test set for different kinds of encodings of the logical statements

### 4.4.2 Comparing different prompting strategies

For this series of evaluations, we used the mixed test set (cf. Section 3.2.4) to evaluate how different instructions affect the performance of the LLMs. In all cases, we used the Unicode encoding for the first-order logic formulae (cf. Section 3.3.1) and a 0-shot prompt (cf. Section 3.3.2.2). For prompting, we compared the three sets of instructions described in Section 3.3.2.1, i.e., yes/no, CoT, and CoT (complex).

The results of these tests are shown in Table 4.5 and Figure 4.22. Table 4.5 shows the percentage of correct answers for each model and for each prompting strategy as well as the percentage of prompts where the model answered 'yes' and the percentage of prompts where the model did not arrive at a yes/no answer (i.e., where the model's answer fell into either the 'vague' or the 'undecidable' category, cf. Section 3.3.2.1). Note that the construction of the mixed test set (cf. Section 3.2.4) ensures that 50% of the statements are true and 50% are false. In Figure 4.21, the opacity in each bar indicates whether an answer was correct or incorrect, while the hatching indicates whether the model's answer amounted to 'yes', 'no', or something else.

|  | Condition | | |
| --- | --- | --- | --- |
| Model | 0-shot yes/no | 0-shot CoT | 0-shot CoT (long) |
| Claude 3 Haiku (2024-03-07) | 62.2 (18.8, 0.0) | 64.6 (72.1, 6.0) | 70.6 (69.2, 0.8) |
| Claude 3.5 Haiku (2024-10-22) | 71.3 (24.5, 0.0) | 77.9 (51.6, 0.9) | 76.9 (56.3, 0.4) |
| Claude 3.5 Sonnet (2024-06-20) | 77.9 (61.8, 0.0) | 91.5 (48.7, 0.0) | 89.9 (51.9, 0.0) |
| Claude 3.5 Sonnet (2024-10-22) | 82.0 (53.7, 0.0) | 90.5 (42.1, 0.0) | 92.3 (45.2, 0.0) |
| DeepSeek-V3 | 78.7 (54.6, 0.0) | **96.3** (51.9, 0.0) | **95.0** (50.3, 0.0) |
| Gemini 1.5 Flash 8b (001) | 66.5 (80.6, 0.0) | 76.7 (51.6, 1.5) | 77.0 (52.9, 0.9) |
| Gemini 1.5 Flash (001) | 74.6 (63.8, 0.0) | 75.0 (40.1, 0.4) | 70.2 (38.2, 0.0) |
| Gemini 1.5 Flash (002) | 78.7 (60.2, 0.0) | 88.6 (45.9, 0.8) | 86.8 (46.7, 0.1) |
| Gemini 2.0 Flash (001) | 59.7 (82.7, 0.0) | 86.1 (55.6, 5.4) | 86.6 (54.8, 2.8) |
| Gemini 1.5 Pro (001) | 79.2 (55.4, 0.0) | 76.1 (39.7, 0.7) | 77.2 (42.1, 0.0) |
| Gemini 1.5 Pro (002) | 80.4 (61.6, 0.0) | 92.3 (48.9, 0.3) | 93.3 (48.8, 0.0) |
| GPT-3.5 Turbo Instruct | 51.7 (97.2, 0.0) | 50.4 (98.3, 0.7) | 55.7 (35.6, 9.4) |
| GPT-4o mini (2024-07-18) | 68.3 (73.5, 0.0) | 76.1 (68.8, 0.1) | 80.8 (64.0, 0.1) |
| GPT-4o (2024-05-13) | 78.3 (48.7, 0.0) | 85.4 (56.6, 1.6) | 86.5 (53.0, 0.8) |
| GPT-4o (2024-08-06) | 79.2 (62.6, 0.0) | 89.0 (54.2, 0.8) | 89.9 (52.6, 0.3) |
| GPT-4o (2024-11-20) | 80.6 (59.1, 0.0) | 88.8 (52.8, 0.3) | 90.9 (53.4, 0.1) |
| Llama 3.1 8b | 50.8 (0.8, 0.0) | 64.6 (39.6, 5.2) | 61.5 (39.8, 5.4) |
| Llama 3.1 70b | 78.0 (41.0, 0.0) | 78.2 (56.7, 4.8) | 79.1 (57.4, 1.6) |
| Llama 3.1 405b | **83.1** (46.8, 0.0) | 83.2 (47.9, 3.2) | 82.8 (50.3, 1.7) |
| Llama 3.3 70b | 72.6 (68.7, 0.0) | 83.3 (49.5, 9.7) | 87.0 (53.6, 2.5) |
| Mistral Nemo | 59.8 (25.7, 0.0) | 65.9 (62.4, 2.9) | 63.4 (39.3, 0.3) |
| Mistral Small (2024-09) | 66.8 (53.6, 0.0) | 72.5 (72.2, 0.8) | 72.5 (70.1, 0.3) |
| Mistral Small (2025-01) | 70.8 (49.9, 0.0) | 74.1 (62.3, 2.8) | 73.8 (63.1, 2.8) |
| Mistral Large (2024-07) | 80.2 (42.3, 0.0) | 84.1 (58.5, 2.1) | 84.8 (58.9, 0.5) |
| Mistral Large (2024-11) | 72.4 (76.9, 0.0) | 87.4 (57.4, 0.8) | 87.3 (54.2, 0.1) |

Table 4.5: Percentages of correct answers achieved by LLMs on the mixed test set for plain yes/no prompt and different CoT prompts (with the percentage of 'yes' answers and the percentage of refusals to answer in parentheses)

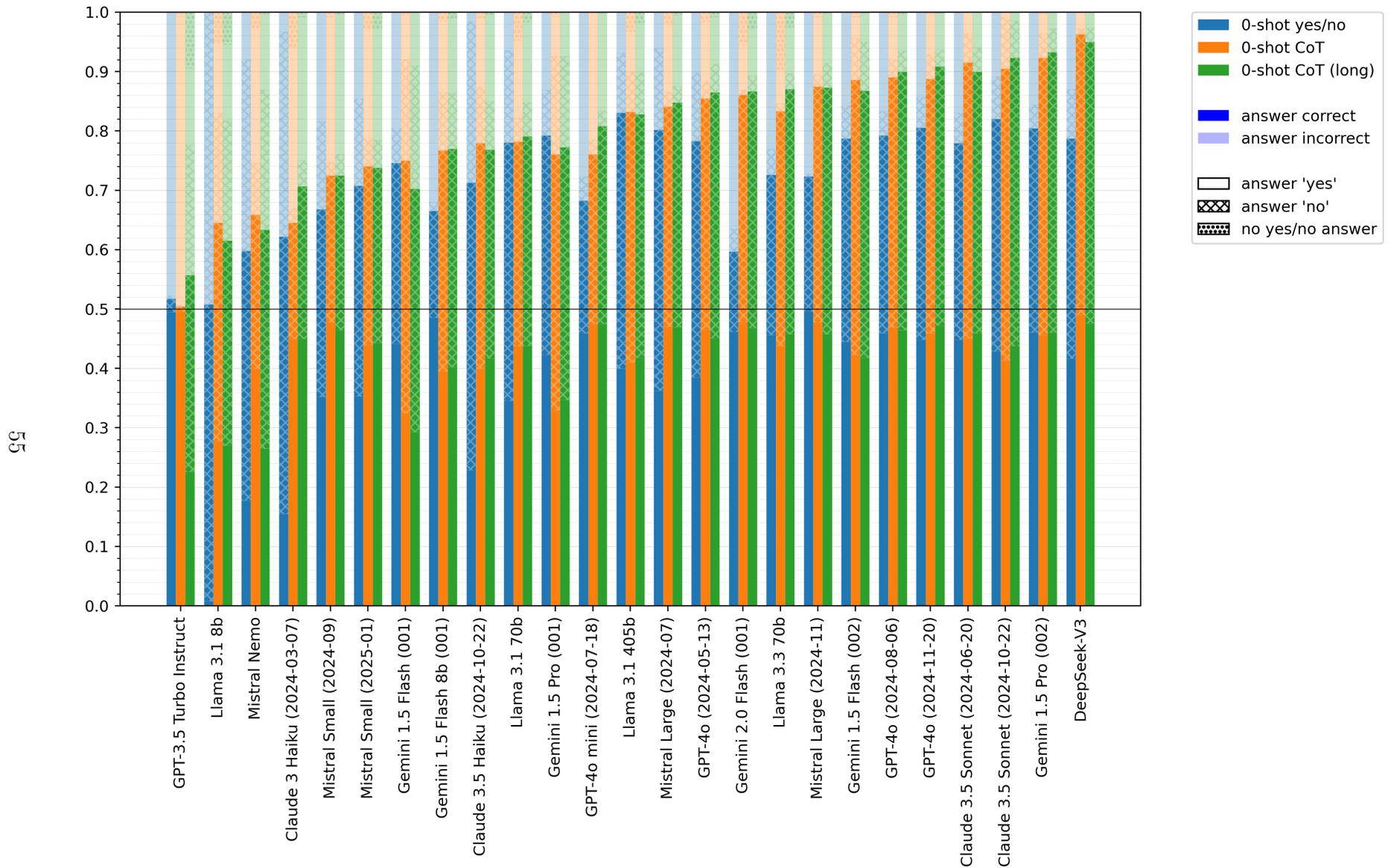

Figure 4.22: Percentages of correct answers achieved by LLMs on the mixed test set for a plain yes/no prompt and different CoT prompts

### 4.4.3 Comparing 0-shot and few-shot prompting

For this series of evaluations, we used the mixed test set (cf. Section 3.2.4) to evaluate how fewshot prompting affects the performance of the LLMs. In all cases, we used the Unicode encoding for the first-order logic formulae (cf. Section 3.3.1). We compared three different prompting strategies: 0-shot yes/no prompting, 3-shot yes/no prompting, and 2-shot CoT prompting (cf. Section 3.3.2.2). The results of these tests are shown in Table 4.6 and Figure 4.23. Table 4.6 shows the percentage of correct answers for each model and for each prompting strategy as well as the percentage of prompts where the model answered 'yes' and the percentage of prompts where the model did not arrive at a yes/no answer (i.e., where the model's answer fell into either the 'vague' or the 'undecidable' category, cf. Section 3.3.2.1). Note that the construction of the mixed test set (cf. Section 3.2.4) ensures that 50% of the statements are true and 50% are false. In Figure 4.23, the opacity in each bar indicates whether an answer was correct or incorrect, while the hatching indicates whether the model's answer amounted to 'yes', 'no', or something else.

|  | Condition | | |
| Model | 0-shot yes/no | 3-shot yes/no | 2-shot CoT |
| --- | --- | --- | --- |
| Claude 3 Haiku (2024-03-07) | 62.2 (18.8, 0.0) | 71.3 (75.0, 0.0) | 71.2 (60.4, 0.4) |
| Claude 3.5 Haiku (2024-10-22) | 71.3 (24.5, 0.0) | 74.3 (50.5, 0.0) | 80.3 (52.5, 0.0) |
| Claude 3.5 Sonnet (2024-06-20) | 77.9 (61.8, 0.0) | 83.7 (56.2, 0.0) | 93.8 (51.5, 0.0) |
| Claude 3.5 Sonnet (2024-10-22) | 82.0 (53.7, 0.0) | **88.6** (51.6, 0.0) | **94.4** (44.4, 0.0) |
| DeepSeek-V3 | 78.7 (54.6, 0.0) | 83.2 (64.7, 0.0) | 82.7 (60.7, 0.0) |
| Gemini 1.5 Flash 8b (001) | 66.5 (80.6, 0.0) | 68.3 (75.9, 0.0) | 74.9 (42.7, 2.8) |
| Gemini 1.5 Flash (001) | 74.6 (63.8, 0.0) | 75.4 (53.7, 0.0) | 79.0 (46.0, 0.4) |
| Gemini 1.5 Flash (002) | 78.7 (60.2, 0.0) | 77.0 (52.9, 0.0) | 89.9 (45.2, 0.1) |
| Gemini 2.0 Flash (001) | 59.7 (82.7, 0.0) | 72.0 (63.5, 0.0) | 80.4 (58.7, 7.5) |
| Gemini 1.5 Pro (001) | 79.2 (55.4, 0.0) | 79.0 (59.1, 0.0) | 78.8 (41.0, 0.5) |
| Gemini 1.5 Pro (002) | 80.4 (61.6, 0.0) | 84.9 (57.9, 0.0) | 93.8 (48.8, 0.0) |
| GPT-3.5 Turbo Instruct | 51.7 (97.2, 0.0) | 55.0 (48.9, 0.1) | 59.4 (73.8, 0.1) |
| GPT-4o mini (2024-07-18) | 68.3 (73.5, 0.0) | 72.9 (63.6, 0.0) | 76.7 (68.8, 0.4) |
| GPT-4o (2024-05-13) | 78.3 (48.7, 0.0) | 77.5 (62.7, 0.0) | 86.2 (56.7, 1.7) |
| GPT-4o (2024-08-06) | 79.2 (62.6, 0.0) | 76.1 (68.1, 0.0) | 89.4 (54.2, 1.1) |
| GPT-4o (2024-11-20) | 80.6 (59.1, 0.0) | 79.1 (52.4, 0.0) | 88.6 (46.4, 0.3) |
| Llama 3.1 8b | 50.8 (0.8, 0.0) | 69.3 (52.9, 0.0) | 62.4 (43.7, 1.3) |
| Llama 3.1 70b | 78.0 (41.0, 0.0) | 77.8 (63.8, 0.0) | 80.4 (64.3, 1.3) |
| Llama 3.1 405b | **83.1** (46.8, 0.0) | 82.4 (56.7, 0.0) | 89.6 (50.7, 0.0) |
| Llama 3.3 70b | 72.6 (68.7, 0.0) | 79.1 (61.6, 0.0) | 86.5 (59.3, 1.5) |
| Mistral Nemo | 59.8 (25.7, 0.0) | 60.2 (55.7, 0.0) | 63.5 (57.7, 0.1) |
| Mistral Small (2024-09) | 66.8 (53.6, 0.0) | 69.2 (68.1, 0.0) | 73.0 (66.8, 0.4) |
| Mistral Small (2025-01) | 70.8 (49.9, 0.0) | 71.3 (62.0, 0.0) | 76.2 (59.0, 2.5) |
| Mistral Large (2024-07) | 80.2 (42.3, 0.0) | 80.6 (57.5, 0.0) | 83.6 (61.4, 0.3) |
| Mistral Large (2024-11) | 72.4 (76.9, 0.0) | 77.2 (57.1, 0.0) | 86.4 (58.1, 0.3) |

Table 4.6: Percentages of correct answers achieved by LLMs on the mixed test set for a plain 0-shot yes/no prompt, a 3-shot prompt, and a 3-shot CoT prompt (with the percentage of 'yes' answers and the percentage of refusals to answer in parentheses)

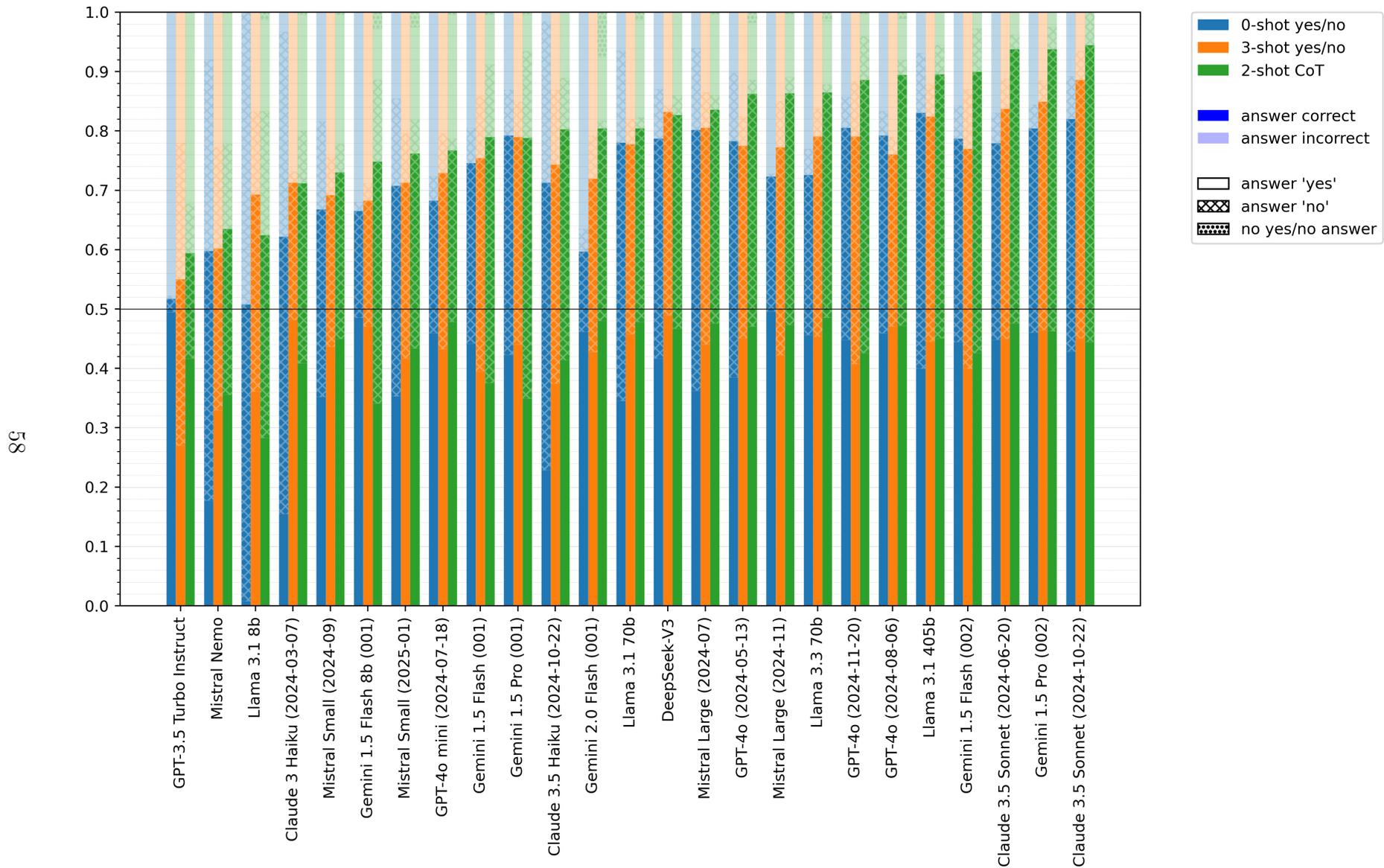

Figure 4.23: Percentages of correct answers achieved by LLMs on the mixed test set for a plain 0-shot yes/no prompt, a 3-shot prompt, and a 2-shot CoT prompt

## 4.5 Evaluating LLM performance on judging more complex logical statements

For this series of evaluations we used the hard test set (cf. Section 3.2.5) to evaluate the performance of LLMs on a set of logical statements that are on average much more complex then the statements in the mixed test set. In all cases, we used the Unicode encoding for the first-order logic formulae (cf. Section 3.3.1) and a 0-shot prompt (cf. Section 3.3.2.2). For the "reasoning" models (DeepSeek-R1, Gemini 2.0 Flash Thinking Experimental, o1-mini, o1-preview, o1, and o3-mini), we employed simple yes/no instructions (as explicit CoT instructions are discouraged for these models). For all other models, we used CoT instructions as described in the CoT condition in Section 3.3.2.1.

Since the reasoning models from OpenAI's o1 and o3 family do not support the temperature parameter, we did not set the temperature to 0.0 for these models. In addition, the experimental Gemini 2.0 Flash Thinking Experimental model seems to support neither a temperature parameter nor a system message, so these parameters were not set in our evaluations. We evaluated the o1 and o3-mini models for each of the three available values of the `reasoning_effort` parameter (low, medium, and high).

| Model | |
|---|---|
| Claude 3.5 Sonnet (2024-10-22) | 74.0 (46.0, 0.0) |
| DeepSeek-V3 | 74.0 (57.5, 1.0) |
| DeepSeek-R1 | 95.5 (51.5, 0.0) |
| Gemini 1.5 Flash (002) | 60.0 (42.5, 10.0) |
| Gemini 2.0 Flash (001) | 52.5 (76.5, 12.5) |
| Gemini 1.5 Pro (002) | 64.5 (63.5, 0.0) |
| Gemini 2.0 Pro Exp (2025-02-05) | 72.5 (67.5, 0.0) |
| Gemini 2.0 Flash Thinking Exp (2025-01-21) | 87.0 (54.0, 0.0) |
| GPT-4o mini (2024-07-18) | 51.0 (91.5, 1.0) |
| GPT-4o (2024-11-20) | 53.5 (72.0, 9.5) |
| o1-preview (2024-09-12), medium | 78.5 (56.5, 0.5) |
| o1-mini (2024-09-12), medium | 85.0 (56.0, 0.0) |
| o3-mini (2025-01-31), low | 56.5 (33.5, 0.0) |
| o3-mini (2025-01-31), medium | 90.0 (59.0, 0.0) |
| o3-mini (2025-01-31), high | 96.5 (52.5, 0.5) |
| o1 (2024-12-17), low | 86.5 (47.5, 0.0) |
| o1 (2024-12-17), medium | 95.5 (52.5, 0.0) |
| o1 (2024-12-17), high | **98.5** (50.5, 0.0) |
| Llama 3.1 405b | 58.0 (59.5, 7.5) |
| Llama 3.3 70b | 42.5 (63.5, 25.5) |
| Mistral Small (2025-01) | 46.5 (66.0, 8.0) |
| Mistral Large (2024-11) | 50.0 (50.0, 23.0) |

Table 4.7: Percentages of correct answers achieved by LLMs on the hard test set (with the percentage of 'yes' answers and the percentage of refusals to answer in parentheses)

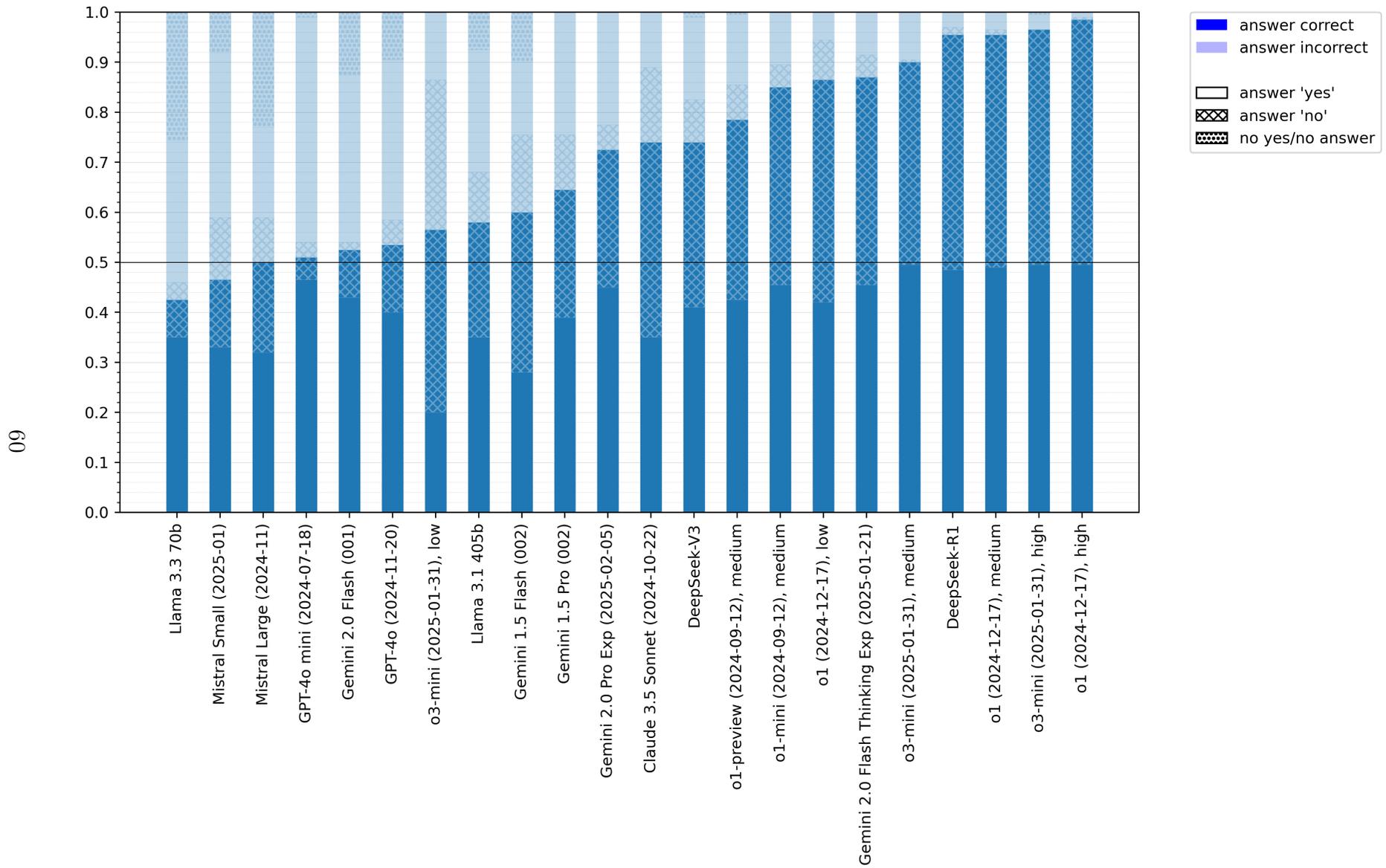

Figure 4.24: Percentages of correct answers achieved by LLMs on the hard test set

**Acknowledgements**

This project has been partially funded by the Deutsche Forschungsgemeinschaft (DFG, German Research Foundation) in the frame of the priority programme SPP 2298 'Theoretical Foundations of Deep Learning' – Project no. 464123384. Moreover, this work has been supported by the Ministry of Culture and Science NRW as part of the Lamarr Fellow Network.